\documentclass{article}

\usepackage[final,nonatbib]{neurips_2024}
\usepackage[square,numbers]{natbib}
\usepackage[utf8]{inputenc} %
\usepackage[T1]{fontenc}    %
\usepackage[normalem]{ulem}

\usepackage{url}            %
\usepackage{booktabs}       %
\usepackage{amsfonts}       %
\usepackage{nicefrac}       %
\usepackage{microtype}      %
\usepackage{xcolor, soul}
\usepackage{authblk}
\usepackage{pythonhighlight}
\usepackage{wrapfig}
\usepackage{bm}
\usepackage[most]{tcolorbox}
\usepackage[textsize=tiny]{todonotes}

\usepackage{xcolor, soul}

\usepackage[toc,page,header]{appendix}
\usepackage{minitoc}

\usepackage{microtype}
\usepackage{graphicx}
\usepackage{caption}
\usepackage{subcaption}
\usepackage{dsfont}
\usepackage{lipsum}
\usepackage{multirow}
\usepackage{float}

\newcommand{\bignumber}[1]{{\Large #1}}
\newcommand{\mednumber}[1]{{\large #1}}
\definecolor{ForestGreen}{RGB}{34,139,34}
\usepackage{marvosym}
\usepackage{fontawesome}
\newcommand{\yesmark}{\textcolor{ForestGreen}{\large \faSmileO}}%
\newcommand{\nomark}{\textcolor{red}{\large \faFrownO{}}}%
\newcommand{\sosomark}{\textcolor{orange}{\large \faMehO{}}}%

\usepackage{amsmath}
\usepackage{amssymb}
\usepackage{mathtools}
\usepackage{amsthm}

\usepackage[colorlinks]{hyperref}
\usepackage[capitalize,noabbrev]{cleveref}

\title{Multilinear Mixture of Experts: \\Scalable Expert Specialization through Factorization}

\author{
\textbf{
James Oldfield}$^1$\thanks{Corresponding author: \texttt{j.a.oldfield@qmul.ac.uk}}\quad
\textbf{Markos Georgopoulos}\quad
\textbf{Grigorios G. Chrysos}$^{2}$\quad
\textbf{Christos Tzelepis}$^{3}$\quad
\textbf{Yannis Panagakis}$^{4,5}$
\textbf{Mihalis A. Nicolaou}$^6$\quad
\textbf{Jiankang Deng}$^7$\quad
\textbf{Ioannis Patras}$^1$
\vspace{-0.75em}
}
\affil{
$^{1}$Queen Mary University of London\quad
$^{2}$University of Wisconsin-Madison\quad
$^{3}$City University of London\quad
$^{4}$National and Kapodistrian University of Athens\quad
$^{5}$Archimedes AI, Athena RC\quad
$^{6}$The Cyprus Institute\quad
$^{7}$Imperial College London
}

\begin{document}
\doparttoc %
\faketableofcontents %

\maketitle

\begin{abstract}
The Mixture of Experts (MoE) paradigm provides a powerful way to decompose dense layers into smaller, modular computations often more amenable to human interpretation, debugging, and editability. However, a major challenge lies in the computational cost of scaling the number of experts high enough to achieve fine-grained specialization. In this paper, we propose the \textbf{Mu}ltilinear \textbf{M}ixture \textbf{o}f \textbf{E}xperts ($\bm{\mu}$\textbf{MoE}) layer to address this, focusing on vision models. $\mu$MoE layers enable scalable expert specialization by performing an implicit computation on prohibitively large weight tensors \textit{entirely in factorized form}. Consequently, $\mu$MoEs (1) avoid the restrictively high inference-time costs of dense MoEs, yet (2) do not inherit the training issues of the popular sparse MoEs' discrete (non-differentiable) expert routing. We present both qualitative and quantitative evidence that scaling $\mu$MoE layers when fine-tuning foundation models for vision tasks leads to more specialized experts at the class-level, further enabling manual bias correction in CelebA attribute classification. Finally, we show qualitative results demonstrating the expert specialism achieved when pre-training large GPT2 and MLP-Mixer models with parameter-matched $\mu$MoE blocks at every layer, maintaining comparable accuracy. Our code is available at: \url{https://github.com/james-oldfield/muMoE}.
\end{abstract}

\section{Introduction}
\label{sec:introduction}

The Mixture of Experts (MoE) architecture \cite{jacobs1991adaptive} has reemerged as a powerful class of conditional computation,
playing the pivotal role in scaling up recent large language \cite{jiang2024mixtral,lepikhin2021gshard,fedus2022switch,gale2023megablocks}, vision \cite{riquelme2021scaling}, and multi-modal models \cite{mustafa2022multimodal}.
MoEs apply different subsets of layers (referred to as `experts') for each input, in contrast to the traditional approach of using the same single layer for all inputs.
This provides a form of input-conditional computation \cite{ha2017hypernetworks,vaswani2017attention,han2021dynamic,chen2020dynamic} that is expressive yet efficient.
However, through their substantial performance gains, an important emergent property of MoEs is frequently underutilized: the innate tendency of experts to specialize in distinct subtasks. %
Indeed, the foundational work of \citet{jacobs1991task} on MoEs describes this property, highlighting how implementing a particular function with modular building blocks (experts) often leads to subcomputations that are easier to understand individually than their dense layer counterparts--with larger expert counts allowing for more fine-grained specialization.

Independent of model performance, a successful decomposition of the layer's functionality into human-comprehensible subtasks offers many significant benefits. Firstly, the mechanisms through which a network produces an output are more \textit{interpretable}: the output is a sum of modular components, each contributing individual functionality.
Yet, the value of interpretable computation extends beyond just transparency \cite{lipton2018mythos} and explainability \cite{ribeiro2016should}.
An important corollary of successful task decomposition amongst experts is that layers are easier to debug and edit. Biased or unsafe behaviors can be better localized to specific experts' subcomputation, facilitating manual correction or surgery in a way that minimally affects the other functionality of the network.
Addressing such behaviors is particularly crucial in the context of foundation models; being often fine-tuned as black boxes pre-trained on unknown, potentially imbalanced
data distributions. Furthermore, there is evidence that traditional fairness techniques are less effective in large-scale models \cite{mao2023lastlayer,cherepanova2021technical}.
However, to achieve fine-grained expert specialism at the class level (or more granular still), one needs the ability to significantly scale up the number of experts.
When using only a small expert count, each expert is forced to process and generalize across \textit{multiple} distinct semantic concepts, hindering specialization.
Conversely, a large expert count means each can specialize to a more specific set of semantically similar inputs.
Alas, the dominating `sparse' MoE paradigm of selecting only the top-$K$ experts \cite{shazeer2017} is not only parameter-inefficient for large expert counts, but also has several well-known issues due to its discrete expert routing--often leading to training instability and difficulties in scaling the total expert count, amongst other challenges \cite{mohammed2022models,puigcerver2024sparse}.
\begin{wrapfigure}[9]{r}{0.425\textwidth}
    \centering
    \captionof{table}{Benefits of the proposed $\mu$MoEs' model form over existing MoEs.}
    \resizebox{0.40\textwidth}{!}{%
    \begin{tabular}{@{}lccc@{}}
        \toprule
         & & \textbf{Parameter-} & \textbf{FLOPs-} \\
         & \textbf{Differentiable} & \textbf{efficient} & \textbf{efficient} \\ \midrule
        Dense MoE \cite{jacobs1991adaptive} & \yesmark & \nomark & \nomark \\
        Sparse MoE \cite{shazeer2017} & \nomark & \nomark & \yesmark \\
        \midrule
        \textbf{$\bm{\mu}$MoE (ours)} & \yesmark & \yesmark & \yesmark \\
        \bottomrule
    \end{tabular}%
    }
    \label{tab:moe-compare}
\end{wrapfigure}

In this paper, we propose the \textit{Multilinear Mixture of Experts} ($\mu$MoE) layer to address these issues. %
$\mu$MoEs are designed to scale gracefully to dense operations involving \textit{tens of thousands} of experts at once through implicit computations on a factorized form of the experts' weights.
Furthermore, in contrast to the dominant sparse MoEs' \cite{shazeer2017} non-differentiable nature, $\mu$MoEs are differentiable by design, and thus do not inherit the associated training issues.
We summarize the benefits of $\mu$MoEs' model form over existing MoEs in \cref{tab:moe-compare}.
Crucially, we show evidence that scaling up the number of $\mu$MoE experts leads to increased expert specialism when fine-tuning foundation models for vision tasks. %
Our evidence is provided in three forms: (1) firstly, through the usual qualitative evaluation of inspecting inputs by their expert coefficients. Secondly (2), we further explore the \textit{causal} role of each expert through counterfactual interventions \cite{elazar2021amnesic}.
Lastly, (3) we show how final-layer $\mu$MoE expert specialism facilitates the practical task of model editing--how subcomputation in specific combinations of experts biased towards demographic subpopulations can be manually corrected through straightforward guided edits.

Building on these findings, we demonstrate that $\mu$MoEs offer a compelling alternative to MLPs for pre-training both vision and language models with up to $100$M parameters--enabling large numbers of specialized experts while maintaining comparable performance and parameter counts to the original networks' \textit{single} dense MLPs.

Our contributions and core claims can be summarized as follows:
\begin{itemize}
    \item We introduce $\mu$MoE layers--a mechanism for computing vast numbers of subcomputations and efficiently fusing them conditionally on the input.
    \item We show both qualitatively (through visualization) and quantitatively (through counterfactual intervention) that \textit{increasing the number of $\mu$MoE experts increases task modularity}--learning to specialize in processing just specific input classes when fine-tuning large foundation models for vision tasks. Further, we show manual editing of $\mu$MoE expert combinations can straightforwardly mitigate demographic bias in CelebA attribute classification.
    \item We pre-train both language (GPT2) and vision (MLP-mixer) $\mu$MoE networks, establishing experimentally that models with parameter-matched $\mu$MoE blocks are competitive with existing MLP blocks whilst facilitating expert specialism (qualitatively) throughout.
\end{itemize}

\section{Related Work}
\label{sec:related-work}

\paragraph{Mixture of Experts}
Recent years have seen a resurgence of interest in the Mixture of Experts (MoE) architecture for input-conditional computation \cite{shazeer2017,jacobs1991task,bengio2015conditional,jiang2024mixtral}.
One primary motivation for MoEs is their increased model capacity through large parameter count \cite{shazeer2017,fedus2022switch,jiang2024mixtral}.
In contrast to a single dense layer, the outputs of multiple experts performing separate computations are combined (sometimes with multiple levels of hierarchy \cite{jordan1993hierarchical,Eigen2013LearningFR}).
A simple approach to fusing the outputs is by taking either a convex \cite{Eigen2013LearningFR} or linear \cite{yang2019condconv} combination of the output of each expert.
The seminal work of \citet{shazeer2017} however proposes to take a \textit{sparse} combination of only the top-$K$ most relevant experts, greatly reducing the computational costs of evaluating them all.
More recent works employ a similar sparse gating function to apply just a subset of experts \cite{jiang2024mixtral,du2022glam}, scaling to billions \cite{lepikhin2021gshard} and trillions of parameters \cite{fedus2022switch}.
The discrete expert selection choice of sparse MoEs is not without its problems, however--often leading to several issues including training stability and expert under-utilization \cite{mohammed2022models,puigcerver2024sparse}.

Particularly relevant to this paper are works focusing on designing MoE models to give rise to more interpretable subcomputation \cite{gupta2022multitasklearners,Gururangan2022domainlm,ismail2023interpretablemoe}--hearkening back to one of the original works of \citet{jacobs1991task}, where experts learned subtasks of discriminating between different lower/uppercase vowels.
Indeed a common observation is that MoE experts appear to specialize in processing inputs with similar high-level features.
Researchers have observed MoE experts specializing in processing specific syntax \cite{shazeer2017} and parts-of-speech \cite{Lewis2021BASELS} for language models,
and foreground/background \cite{wu2022residual} and image categories (e.g. `wheeled vehicles') \cite{yang2019condconv} in vision.
Evidence of shared vision-language specialism is even found in the multi-modal MoEs of \citet{mustafa2022multimodal}.

Several works instead target how to make conditional computation more efficient: by sharing expert parameters across layers \cite{xue2022go}, factorizing gating network parameters \cite{davis2013low}, or dynamic convolution operations \cite{li2021revisiting}.
Relatedly, \citet{gao2022ttparameter} jointly parameterize the experts' weight matrices with a Tensor-Train decomposition \cite{Oseledets2011TensorTrainD}.
However, such approach still suffers from the Sparse MoE's instability and expert under-utilization issues, and stochastic masking of gradients must be performed to lead to balanced experts.
Furthermore, whilst \citet{gao2022ttparameter} share parameters across expert matrices, efficient implicit computation of thousands of experts simultaneously is not facilitated, in contrast to the $\mu$MoE layer.

\paragraph{Factorized layers} in the context of deep neural networks provide several important benefits. Replacing traditional operations with low-rank counterparts allows efficient fine-tuning \cite{hu2021lora} / training \cite{novikov2015tensorizing,garipov2016ultimate}, and modeling of higher-order interactions \cite{alex2016exponential,georgopoulos2021mitigating,babiloni2020tesa,georgopoulos2020multilinear,cheng2024multilinear}, and convolutions \cite{kossaifi2020factconv}.
In addition to reducing computational costs, tensor factorization has also proven beneficial in the context of multi-task/domain learning \cite{bulat2020incremental,yang2017multitask} through the sharing of parameters/low-rank factors across tasks.
Furthermore, parameter efficiency through weight factorization often facilitates the design and efficient implementation of novel architectures such as polynomial networks \cite{chrysos2020p,chrysos2021pami,babiloni2021poly} or tensor contraction layers \cite{kossaifi2017tensor}.
The recent DFC layer in \citet{babiloni2023factorised} also performs dynamic computation using the CP decomposition \cite{Hitchcock1927TheEO} like $\mu$MoEs.
Nevertheless, the two works have very different goals and model properties due to how the weight matrices are generated.
$\mu$MoEs take a sparse, convex combination of $N$ explicit experts' latent factors.
This consequently leads to specialized subcomputations in a way that facilitates the interpretability and editability presented in this paper.
DFCs can be seen to apply an MLP to input vectors at this step in analogy, which does not provide the necessary model properties of interest here.

\section{Methodology}

We first formulate the proposed $\mu$MoE layer in \cref{sec:mmoe}, introducing 2 unique resource-efficient models and forward passes in \cref{sec:fast-mmoe}. %
Finally, we show in \cref{sec:bilinear-moe} how $\mu$MoEs recover linear MoEs as a special case.

\paragraph{Notation} \label{sec:notation}
We denote scalars $x\in\mathbb{R}$ with lower-case letters, and vectors $\mathbf{x}\in\mathbb{R}^{I_1}$ and matrices $\mathbf{X}\in\mathbb{R}^{I_1\times I_2}$ in lower- and upper-case boldface latin letters respectively.
Tensors $\mathcal{X}\in\mathbb{R}^{I_1\times I_2 \times \ldots \times I_d}$ of order $d$ are denoted with calligraphic letters.
We refer to the $(i_1,i_2,\ldots,i_d)$-th element of this tensor with both $\mathcal{X}(i_1,i_2,\ldots,i_d)\in\mathbb{R}$ and $x_{i_1i_2\ldots i_d}\in\mathbb{R}$.
Finally, we use a colon to index into all elements along a particular mode: given $\mathcal{X}\in\mathbb{R}^{I_1 \times I_2 \times I_3}$ for example, $\mathbf{X}_{::i_3}\in\mathbb{R}^{I_1\times I_2}$ or equivalently $\mathcal{X}(:,:,i_3)\in\mathbb{R}^{I_1\times I_2}$ is the matrix at index $i_3$ of the final mode of the tensor.
We use $\mathcal{X}\times_n\mathbf{u}$ to denote the \textbf{mode-$n$ (vector) product} \cite{kolda2009tensorapplications}
of a tensor $\mathcal{X}\in\mathbb{R}^{I_1\times I_2 \times\ldots \times I_N}$ and vector $\mathbf{u}\in\mathbb{R}^{I_n}$ whose resulting elements are given by $(\mathcal{X}\times_n\mathbf{u})_{i_1\ldots i_{n-1} i_{n+1}\ldots i_N}= \sum_{i_n=1}^{I_n}x_{i_1i_2\ldots i_N} u_{i_n}$.

\subsection{The $\boldsymbol\mu$MoE layer}
\label{sec:mmoe}

$\mu$MoEs provide a scalable way to execute and fuse large numbers of operations on an input vector by formalizing conditional computation through resource-efficient multilinear operations. A $\mu$MoE layer comprised of $N$ many experts (and a single level of expert hierarchy) is parameterized by weight tensor $\mathcal{W}\in\mathbb{R}^{N\times I \times O}$ and expert gating parameter $\mathbf{G}\in\mathbb{R}^{I\times N}$.
Given an input vector $\mathbf{z}\in\mathbb{R}^I$ (denoting the hidden representation of an individual token, for example), its forward pass can be expressed through the series of tensor contractions:
    \begin{wrapfigure}[16]{r}{0.5\textwidth}
        \centering
        \includegraphics[width=0.40\textwidth]{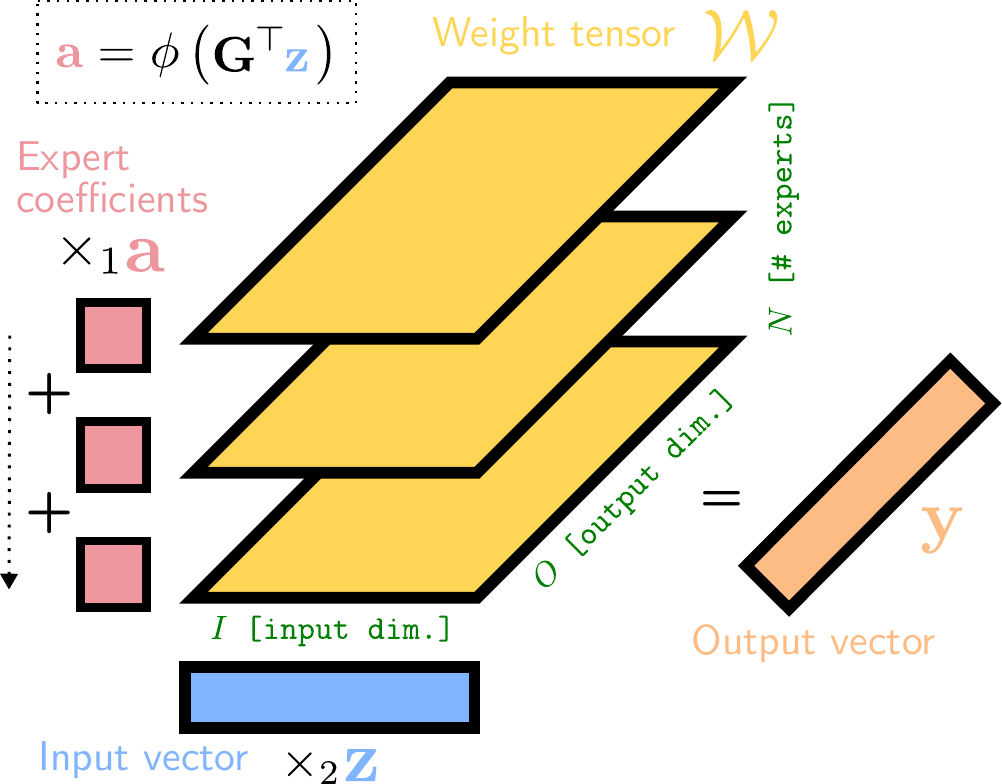}
        \caption{The forward pass of an (unfactorized) $\mu$MoE layer as a series of tensor contractions: the experts' weight matrices (yellow $2$D slices) are matrix-multiplied with the input vector and summed (weighted by the red expert coefficients).}
        \label{fig:overview}
    \end{wrapfigure}
\begin{align}
    \label{eq:mmoe-moden}
    \mathbf{a} &=\phi(\mathbf{G}^\top\mathbf{z})\in\mathbb{R}^{N},\nonumber\\
    \mathbf{y} &= \mathcal{W}\times_1 \mathbf{a} \times_{2} \mathbf{z} \nonumber\\
        &= \sum_{n=1}^N\sum_{i=1}^I \mathbf{w}_{ni:} z_i a_n \in \mathbb{R}^O,
\end{align}
where $\mathbf{a}$ is the vector of expert coefficients and $\phi$ is the entmax activation \cite{peters2019entmax,correia2019entmax}.
The $\mu$MoE layer can be understood as taking a sparse, convex combination of $N$ many affine transformations\footnote{Incrementing the dimension of the second `input' mode of the weight tensor $\mathcal{W}\in\mathbb{R}^{N\times (I+1) \times O}$ and appending a $1$ to the input vector $\mathbf{z}\in\mathbb{R}^{I+1}$ folds a per-expert bias term into the computation.} of input vector $\mathbf{z}$, weighted by the coefficients in $\mathbf{a}$.
The first tensor contraction in the forward pass
($\sum_i\mathbf{W}_{:i:}z_i\in\mathbb{R}^{N\times O}$)
matrix-multiplies the input vector with \textit{every} expert's weight matrix.
The following tensor contraction with expert coefficients $\mathbf{a}$ takes a linear combination of the results, yielding the output vector.
The forward pass can be visualized intuitively as multiplying and summing over the modes in a 3D tensor, which we illustrate in \cref{fig:overview}.
Furthermore, $\mu$MoEs readily generalize to hierarchical conditional computations by introducing additional modes to the weight tensor and corresponding vectors of expert coefficients (see \cref{sec:app:full-model}).

\subsubsection{Computation in factorized form}
\label{sec:fast-mmoe}

Our key insight
is that the dense $\mu$MoE forward pass over all $N$ experts simultaneously can be \textbf{computed entirely in factorized form, needing never materialize prohibitively large weight tensors}.
This allows $\mu$MoEs' computations to scale gracefully to many thousands of experts simultaneously, without the problematic top-$K$ gating \cite{shazeer2017}.
To achieve this, we (1) first parameterize the experts' weights $\mathcal{W}\in\mathbb{R}^{N \times I \times O}$ with a tensor factorization and (2) re-derive fast forward passes of \cref{eq:mmoe-moden} to operate solely in factorized form.

In the context of a $\mu$MoE layer, the various choices of tensor factorizations make different trade-offs regarding parameter/FLOP counts and rank constraints.
We derive two unique resource-efficient $\mu$MoE variants to suit different computational budgets and choices of expert counts.
We now present the derivations of the forward passes of the factorized $\mu$MoE models (with \texttt{einsum} pseudocode implementations in \cref{sec:app:implementations}):

\paragraph{CP$\boldsymbol\mu$MoE}
Imposing CP structure \cite{Hitchcock1927TheEO,Carroll1970AnalysisOI} of rank $R$ on the weight tensor, we can write
$\mathcal{W}=\sum_{r=1}^R \mathbf{u}^{(1)}_{r} \circ \mathbf{u}^{(2)}_{r} \circ \mathbf{u}^{(3)}_{r}\in\mathbb{R}^{N\times I \times O}$ as a sum of $R$ outer products, with factor matrices $\mathbf{U}^{(1)}\in\mathbb{R}^{R \times N}, \mathbf{U}^{(2)}\in\mathbb{R}^{R \times I}, \mathbf{U}^{(3)}\in\mathbb{R}^{R \times O}$.
This reduces the parameter count from $NIO$ (such as with sparse/dense MoEs and regular $\mu$MoEs) to just $R(N+I+O)$. 
Crucially, we can further rewrite the CP$\mu$MoE layer's forward pass entirely in factorized form without ever materializing the full tensor (plugging the CP-composed tensor into \cref{eq:mmoe-moden}) as:
\begin{align}
    \mathbf{y}
        &= \sum_{n=1}^N\sum_{i=1}^I \bigg(
        \sum_{r=1}^R \mathbf{u}^{(1)}_{r} \circ \mathbf{u}^{(2)}_{r} \circ \mathbf{u}^{(3)}_{r} \bigg)_{ni:}
        z_i
        a_n
        = \sum_{r=1}^R
            \big({\mathbf{U}^{(2)}}\mathbf{z}\big)_r
            \big({\mathbf{U}^{(1)}}\mathbf{a}\big)_r
            \mathbf{u}^{(3)}_{r}
            \in\mathbb{R}^O,\label{eq:fast-CP}
\end{align}
with \cref{eq:fast-CP} being analogous to the fast computation in \citet{babiloni2023factorised}, only here the operations of combining the weights and producing the outputs can be expressed in a single step.
Whilst the original naive CP$\mu$MoE forward pass has a FLOP count\footnote{We adopt the convention of counting fused multiply-adds as one operation \cite{fvcore}. Note that the small additional expert coefficients cost is constant across models and thus ignored in comparisons.} of $NIO$, the fast computation above has just $R(N+I+O)$ (the same number of factorized layer parameters).
With moderate values of both $R$ and $N$, the layer becomes significantly more resource-efficient than vanilla $\mu$MoEs.%

\paragraph{TR$\boldsymbol\mu$MoE}
We propose a second $\mu$MoE variant based on the Tensor Ring \cite{Zhao2016TensorRD} (TR) factorization that can offer even better efficiency for large values of $N$.
In TR format, $\mathcal{W}\in\mathbb{R}^{N\times I \times O}$ has three factor tensors:
$\mathcal{U}^{(1)}\in\mathbb{R}^{R_1 \times N \times R_2}$,
$\mathcal{U}^{(2)}\in\mathbb{R}^{R_2 \times I \times R_3}$,
$\mathcal{U}^{(3)}\in\mathbb{R}^{R_3 \times O \times R_1}$,
where $R_i$ are the manually chosen ranks\footnote{Setting $R_1=1$ recovers a Tensor Train \cite{Oseledets2011TensorTrainD} $\mu$MoE.}.
The weight tensor's elements in TR format are given by: $w_{nio} = \text{tr}\big( {\mathbf{U}^{(1)}_{:n:}} {\mathbf{U}^{(2)}_{:i:}} {\mathbf{U}^{(3)}_{:o:}} \big)$ \cite{Zhao2016TensorRD}.
TR$\mu$MoE's forward passes can be computed efficiently by contracting the first two factor tensors with the input/expert coefficients vectors and then combining the results:
\begin{align}
    \mathbf{y}
        = \sum_{n=1}^N\sum_{i=1}^I \mathbf{w}_{ni:} z_i a_{n}
        = \sum_{r_1=1}^{R_1}\sum_{r_3=1}^{R_3}
        \big(\underbrace{
        (\mathcal{U}^{(1)}\times_2 \mathbf{a})
        (\mathcal{U}^{(2)}\times_2 \mathbf{z})
        }_{[R_1\times R_3]}\big)_{r_1r_3}
        \mathbf{u}^{(3)}_{r_3:r_1}
        \in\mathbb{R}^{O},
        \label{eq:fast-TR}
\end{align}
yielding a modified FLOP count of
$(R_1NR_2 + R_2IR_3 + R_1R_2R_3 + R_1OR_3)$ with just $(R_1NR_2 + R_2IR_3 + R_3OR_1)$ parameters.
With large $N$ contributing to the computational cost only through $R_1NR_2$, the TR$\mu$MoE can prove even more resource-efficient than CP$\mu$MoEs by choosing small values of $R_1,R_2$.
We refer readers to \cref{sec:app:decomps} for a further discussion of decomposition choice, derivations of how tensor rank translates to expert matrix rank, and FLOPs comparisons.

\subsubsection{$\boldsymbol\mu$MoEs recover dense MoEs as a special case}
\label{sec:bilinear-moe}

Finally, we note how unfactorized $\mu$MoE layers with a single level of expert hierarchy recover dense MoE layers \cite{shazeer2017,chen2020dynamic} as a special case.
When computing \cref{eq:mmoe-moden} over the full materialized weight tensor, one can alternatively write the output element-wise as
$y_o=\mathbf{a}^\top\mathbf{W}_{::o}\mathbf{z}$.
This highlights an interesting technical connection between neural network layers: dense MoE layers in this tensor formulation can be seen to share a similar functional form to bilinear layers, which have also found applications in interpretability \cite{sharkey2023technical,pearce2024weightbased}.

\section{Experiments}
\label{sec:experiments}

We start in \cref{sec:exp:prune} by presenting both qualitative and quantitative experiments validating that the experts learn to specialize in processing different semantic clusters of the input data.
In \cref{sec:exp:intervene} we demonstrate one practical benefit of the learned specialism--showing how expert-conditional re-writing can correct for specific demographic bias in CelebA attribute classification.
Finally, in \cref{sec:exp:performance} we train both large language and large vision models with $\mu$MoE layers throughout--providing qualitative evidence of expert specialism and model performance competitive with networks using MLP blocks.
Please see \cref{sec:app:ablation} for detailed ablation studies, and \cref{sec:app:additional-performance} for experiments with hierarchical $\mu$MoEs.

\paragraph{Implementation details}
Before applying the activation function to the expert coefficients we apply batch- and layer-normalization to $\mu$MoE layers in vision and language models respectively (see \cref{app:sec:bn-ablation} for an ablation).
Interestingly, we do not find the need for any load-balancing losses.
We fix the TR$\mu$MoE ranks to be $R_1=R_2=4$ throughout (see \cref{sec:app:tr-rank}).

\subsection{Expert specialism: visualization \& intervention}
\label{sec:exp:prune}

Our first objective is to show that \textbf{scaling $\boldsymbol\mu$MoE's expert count leads to more specialized experts}. We provide evidence of this effect both qualitatively (through \textit{visualization}) and quantitatively (through \textit{intervention}).

To isolate the impact of $\mu$MoE layers and varying expert counts, we first explore the controlled setting of fine-tuning large foundation models CLIP \cite{radford2021learning} \texttt{ViT-B-32} and DINO \cite{caron2021dino} on ImageNET1k (following the fine-tuning protocol in \citet{ilharco2022patching,ilharco2023editing}).
Whilst fine-tuning large foundation models is an important application of $\mu$MoE layers in its own right (e.g. as explored later in \cref{sec:exp:intervene} for fairer models),
the ability to cheaply train many models with different $\mu$MoE layer configurations forms an ideal setting in which to study their properties.

\subsubsection{Qualitative results}
\label{sec:prune-qualitative}

\begin{figure}[]
    \centering
    \includegraphics[width=\linewidth]{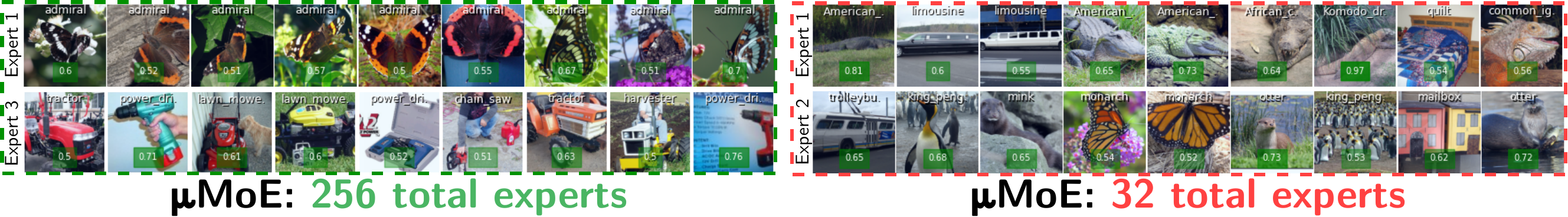}
    \caption{
    Specialization in \textcolor{ForestGreen}{$256$} vs \textcolor{red}{$32$} total expert 
    CP$\mu$MoE layers (fine-tuned on CLIP \texttt{ViT-B-32}).
    Each row displays \textit{randomly} selected images processed (with coefficient $\geq0.5$) by the first few experts for the two models.
    The more we scale the expert count, the greater the apparent expert specialism (to single visual themes or image categories).
    }
    \label{fig:expert-slices}
\end{figure}

We first show \textit{random} examples in \cref{fig:expert-slices} of images processed (with expert coefficient $\geq0.5$) by the experts by each CP$\mu$MoE layer (the class labels and expert coefficients are overlaid in white and green text respectively).
Using only a modest number of experts (e.g. 32) appears to lead to some `polysemanticity' \cite{elhage2022toy} in experts--with some processing unrelated classes of images (e.g. `gators', `limos', and a `quilt' for Expert 1 on the right).
On the other hand, using a much larger number of total experts appears to yield more specialization, with many experts contributing their computation to only images of the same single class label or broader semantic category. %
Please see \cref{fig:image-grids} in the Appendix for many more random images for the first $10$ experts per model to observe this same trend more generally, and \cref{fig:image-grids-specific} for even finer-grained specialism with $2048$-expert $\mu$MoE layers.

\subsubsection{Quantitative results: expert monosemanticity}
\label{sec:prune-quantitative}

The qualitative evidence above hints at the potential of a prominent benefit to scaling up the number of experts with $\mu$MoEs.
Such subjective interpretations alone about expect specialism are \textit{hypotheses}, rather than conclusions however \cite{rauker2023toward}.
Similarities in images processed by the same expert give us an intuitive explanation of its function but do not show the expert's computation contributes \textit{causally} \cite{elazar2021amnesic,ravfogel2021counterfactual,meng2022locating} to the subtask of processing
specific human-understandable patterns of input features \cite{rudin2019stop,casper2023critique}.
However, the absence of ground-truth labels for interpretable features of the input one may be interested in (e.g. specific types of textures in images, or words related to `Harry Potter') makes this difficult to quantify in any objective or systematic manner.

Despite the absence of fine-grained labels, we \textit{can} quantify and compare the class-level specialism a $\mu$MoE expert exhibits on the ImageNET1k dataset as an (imperfect) proxy \cite{hod2021quantifying}.
\begin{wrapfigure}[19]{r}{0.5\textwidth}
    \centering
    \includegraphics[width=0.48\textwidth]{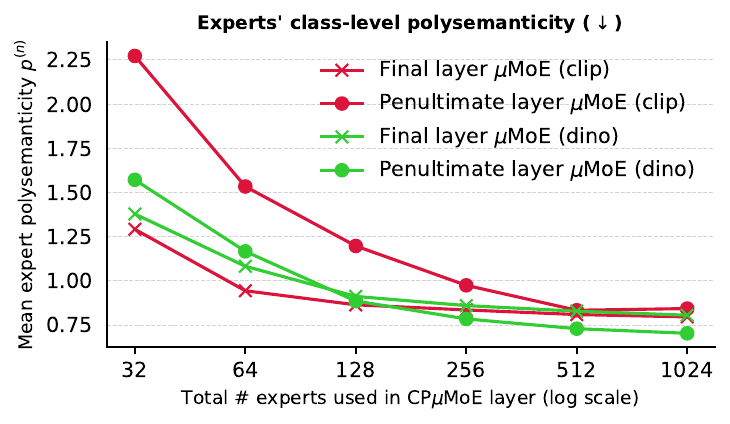}
    \caption{\textbf{Higher expert counts lead to more monosemantic experts}: mean expert class-level polysemanticity of \cref{eq:polysemanticity} ($\downarrow$) as a function of the total number of experts. Results are shown for both CLIP ViT-B-32 and DINO models fine-tuned on ImageNET1k with CP$\mu$MoE layers.}
    \label{fig:polysemanticity}
\end{wrapfigure}
Following the causal intervention protocol of \citet{elazar2021amnesic}, we ask the specific counterfactual question about solely each expert $n$ in a $\mu$MoE layer in turn:
\textit{``had expert $n$'s weight matrix $\mathbf{W}_{n}$ not contributed its computation, would the network's test-set accuracy for class $c$ have dropped?''}
Practically speaking, given a network fine-tuned with an $\mu$MoE layer, we achieve this by intervening in the forward pass by zeroing the $n^\text{th}$ expert's weight matrix $\mathbf{W}_{n}:=\mathbf{0}$, leaving every other aspect of the forward pass completely untouched.
Let the elements of $\mathbf{y},\hat{\mathbf{y}}^{(n)}\in\mathbb{R}^{C}$ denote the test set accuracy for the $C=1000$ ImageNET1k classes, pre- and post-intervention of expert $n$ respectively.
We collect the normalized difference to per-class accuracy in the vector $\mathbf{d}^{(n)}$, whose elements are given by $d^{(n)}_c = (y_c - \hat{y}^{(n)}_c)/y_c$.
At the two extremes, when the full network's accuracy for class $c$ drops completely from $y_c$ to $0$ upon manually excluding expert $n$'s computation
we get $d^{(n)}_c=1$, whilst $d^{(n)}_c=0$ means the absence of the subcomputation did not change class $c$'s test set accuracy at all.
We thus estimate the `class-level polysemanticity' of expert $n$ as the distance between the difference vector and the one-hot vector:
\begin{equation}
    \label{eq:polysemanticity}
    p^{(n)} = \vert\vert \mathbf{d}^{(n)}-\mathds{1}^{(n)}\vert\vert_2,
\end{equation}
where index $\mathrm{argmax}_c(d_c^{(n)})$ of $\mathds{1}^{(n)}$ has a value of $1$ (and values of $0$ everywhere else). This encodes the signature of a perfectly class-level monosemantic expert, for which \textit{all} accuracy for a single class alone is lost in the counterfactual scenario in which the expert $n$ did not contribute.
We plot in \cref{fig:polysemanticity} the average expert polysemanticity $p^{(n)}$ for all experts with non-zero difference vectors\footnote{I.e. we include only experts that, when ablated in isolation, alter the class accuracy; please see the Appendix for discussion on expert load.}, observing a steady drop in its value as $N$ increases from $32$ to $1024$ total experts.
In other words, \textbf{increasing $N$ leads to individual experts increasingly responsible for a single subtask}: classifying all inputs of just one class.
As shown in \cref{fig:polysemanticity} we observe this trend both when $\mu$MoEs are used as final classification layers and as penultimate layers (followed by a ReLU activation and linear classification layer), and for multiple pre-trained foundation models.
We further refer readers to the bar plots of the values of $\mathbf{d}^{(n)}$ (the per-class accuracy changes) in \cref{fig:sup:class-ablate-penultimate,fig:sup:class-ablate-final}, where this trend is observable through mass concentrated on increasingly fewer class labels as the number of experts increases.

\subsection{Expert re-writing: conditional bias correction}
\label{sec:exp:intervene}

We further validate the modular expert hypothesis of $\mu$MoEs and simultaneously provide a concrete example of its usefulness by correcting demographic bias in attribute classification.
Classifiers trained to minimize the standard binary cross-entropy loss often exhibit poor performance for demographic subpopulations with low support \cite{buolamwini2018gender,gebru2021datasheets}.
By identifying which combination of experts is responsible for processing target subpopulations, we show how one can straightforwardly manually correct mispredictions in a targeted way--without \textit{any} re-training.

We focus on mitigating bias towards two low-support subpopulations in models with $\mu$MoE final layers fine-tuned on CelebA \cite{liu2015faceattributes}: (a) bias towards images labeled as `old females' for age prediction \cite{jain2023distilling}, and (b) bias towards images labeled as `blond males' for blond hair prediction \cite{mao2023lastlayer}.
Concretely, we train $N=128$ multi-label $\mu$MoE final layer models for the $40$ binary attributes in CelebA, jointly optimizing a pre-trained CLIP ViT-B-32 model \cite{radford2021learning} backbone, again following the fine-tuning setup in \citet{ilharco2022patching,ilharco2023editing}.
All results presented in this section are the average of 10 runs with different random seeds.
\begin{table*}[t]
\caption{Fairness metrics for baseline models and after applying standard fairness techniques, for the two experiments on CelebA. A \texttt{CP$\mu$MoE-r512-e128} model is used as the final layer.}
\vspace{0.25em}
\label{tab:intervention-fairness}
\centering
\resizebox{\textwidth}{!}{
\begin{tabular}{lcccccclccccccc}
\toprule
& \multicolumn{5}{c}{\bignumber{(a) Bias towards `Old females' for `Age' prediction head}} & &\multicolumn{6}{c}{\bignumber{(b) Bias towards `Blond males' for `Blond Hair' prediction head}} \\
\cmidrule{2-14}
{} & \bignumber{Target}  & \bignumber{Equality of} & \bignumber{STD} & \bignumber{Subpop.} & \bignumber{Test set} & {} & & \bignumber{Target}  & \bignumber{Equality of} & \bignumber{STD} & \bignumber{Subpop.} & \bignumber{Test set} \\
{} &\bignumber{subpop. acc. ($\uparrow$)} & \bignumber{opp. \cite{hardt2016equality} ($\downarrow$)} & \bignumber{bias \cite{wang2020mitigating} ($\downarrow$)} & \bignumber{Max-Min \cite{lahoti2020fairness} ($\uparrow$)} & \bignumber{acc. ($\uparrow$)} & {} & & \bignumber{subpop. acc. ($\uparrow$)} & \bignumber{opp. \cite{hardt2016equality} ($\downarrow$)} & \bignumber{bias \cite{wang2020mitigating} ($\downarrow$)} & \bignumber{Max-Min \cite{lahoti2020fairness} ($\uparrow$)} & \bignumber{acc. ($\uparrow$)} & \bignumber{\# Params} \\
\midrule
\bignumber{Linear}               & \bignumber{0.516} & \bignumber{0.226} & \bignumber{0.185} & \bignumber{0.516} & \bignumber{88.944} & & &  \bignumber{0.346} & \bignumber{0.534} & \bignumber{0.263} & \bignumber{0.346} & \bignumber{95.833} & \bignumber{30.7K} \\
\bignumber{HighRankLinear}     & \bignumber{0.513} & \bignumber{0.228} & \bignumber{0.186} & \bignumber{0.513} & \bignumber{88.920}  & & & \bignumber{0.353} & \bignumber{0.529} & \bignumber{0.260} & \bignumber{0.353} & \bignumber{95.831} & \bignumber{827K} \\
\bignumber{\textbf{CP$\bm{\mu}$MoE}}               & \bignumber{\textbf{0.555}} & \bignumber{\textbf{0.197}} & \bignumber{\textbf{0.167}} & \bignumber{\textbf{0.555}} & \bignumber{\textbf{89.048}} & &  & \bignumber{\textbf{0.409}} & \bignumber{\textbf{0.476}} & \bignumber{\textbf{0.236}} & \bignumber{\textbf{0.409}} & \bignumber{\textbf{95.893}} & \bignumber{578K} \\
\midrule
\bignumber{+ oversample}  & \bignumber{0.669} & \bignumber{0.086} & \bignumber{0.120} & \bignumber{0.669} & \bignumber{\textbf{89.009}} & & & \bignumber{0.655} & \bignumber{0.226} & \bignumber{0.131} & \bignumber{0.655} & \bignumber{\textbf{95.750}} & \bignumber{578K} \\
\bignumber{+ adv. debias \cite{alvi2018turning}} & \bignumber{0.424} & \bignumber{0.274} & \bignumber{0.226} & \bignumber{0.424} & \bignumber{87.785} & & & \bignumber{0.193} & \bignumber{0.630} & \bignumber{0.325} & \bignumber{0.193} & \bignumber{95.031} & \bignumber{579K} \\
\bignumber{+ blind thresh. \cite{hardt2016equality}}   & \bignumber{0.843} & \bignumber{\textbf{0.082}} & \bignumber{0.084} & \bignumber{0.700} & \bignumber{83.369} & & & \bignumber{0.843} & \bignumber{0.139} & \bignumber{0.063} & \bignumber{0.841} & \bignumber{92.447} & \bignumber{578K} \\
\bignumber{+ expert thresh. \textbf{(ours)}}     & \bignumber{\textbf{0.866}} & \bignumber{0.097} & \bignumber{\textbf{0.066}} & \bignumber{\textbf{0.756}} & \bignumber{84.650} &&&  \bignumber{\textbf{0.847}} & \bignumber{\textbf{0.051}} & \bignumber{\textbf{0.048}} & \bignumber{\textbf{0.846}} & \bignumber{94.895} & \bignumber{578K} \\
\bottomrule
\end{tabular}
}
\end{table*}

\paragraph{Experimental setup}
Let $C$ be a set collecting the expert coefficients $\mathbf{a}\in\mathbb{R}^N$ from forward passes of the training images belonging to the target subpopulation. We evaluate the subpopulation's mean expert coefficients $\mathbf{\bar{a}}=1/|C| \sum_{\mathbf{a}\in C} \mathbf{a} \in \mathbb{R}^N$,
proposing to manually re-write the output of this expert combination.
We modify the layer's forward pass for the $o^\text{th}$ output head for attribute of interest (e.g. `blond hair') as:
\begin{equation}
\label{eq:mmoe-intervention}
    y_o=\mathbf{a}^\top\mathbf{W}_{::o}\mathbf{z} 
        + \lambda\mathbf{\bar{a}}^\top\mathbf{a}.
\end{equation}
Here, the term $\lambda\mathbf{\bar{a}}\in\mathbb{R}^N$ specifies, for each expert, how much to increase/decrease the logits for attribute $o$, with $\lambda$ being a scaling hyperparameter\footnote{We set $\lambda:=N$ for all experiments for simplicity, but we note that its value could require tuning in different experimental setups. The sign of $\lambda$ is chosen to correct the bias in the target direction (whether to move the logits positively/negatively towards CelebA's e.g. young/old binary age labels respectively).}. Taking the dot product with an input image's expert coefficients $\mathbf{a}$ applies the relevant experts' correction terms (in the same way it selects a subset of the most relevant experts' weight matrices).
We report a range of standard fairness metrics for both the model rewriting and networks trained with existing techniques (that aim to mitigate demographic bias without requiring images' sensitive attribute value at test time). These are shown in \cref{tab:intervention-fairness} for the two different experiments on CelebA, where the proposed intervention outperforms baseline alternative methods in the majority of settings.
Please see \cref{sec:app:fairness-details} for details about the baseline methods and fairness metrics used, and further discussion of results.

\subsection{Large language/vision $\boldsymbol\mu$MoE networks}
\label{sec:large-models}

\begin{figure}[]
    \centering
    \includegraphics[width=\linewidth]{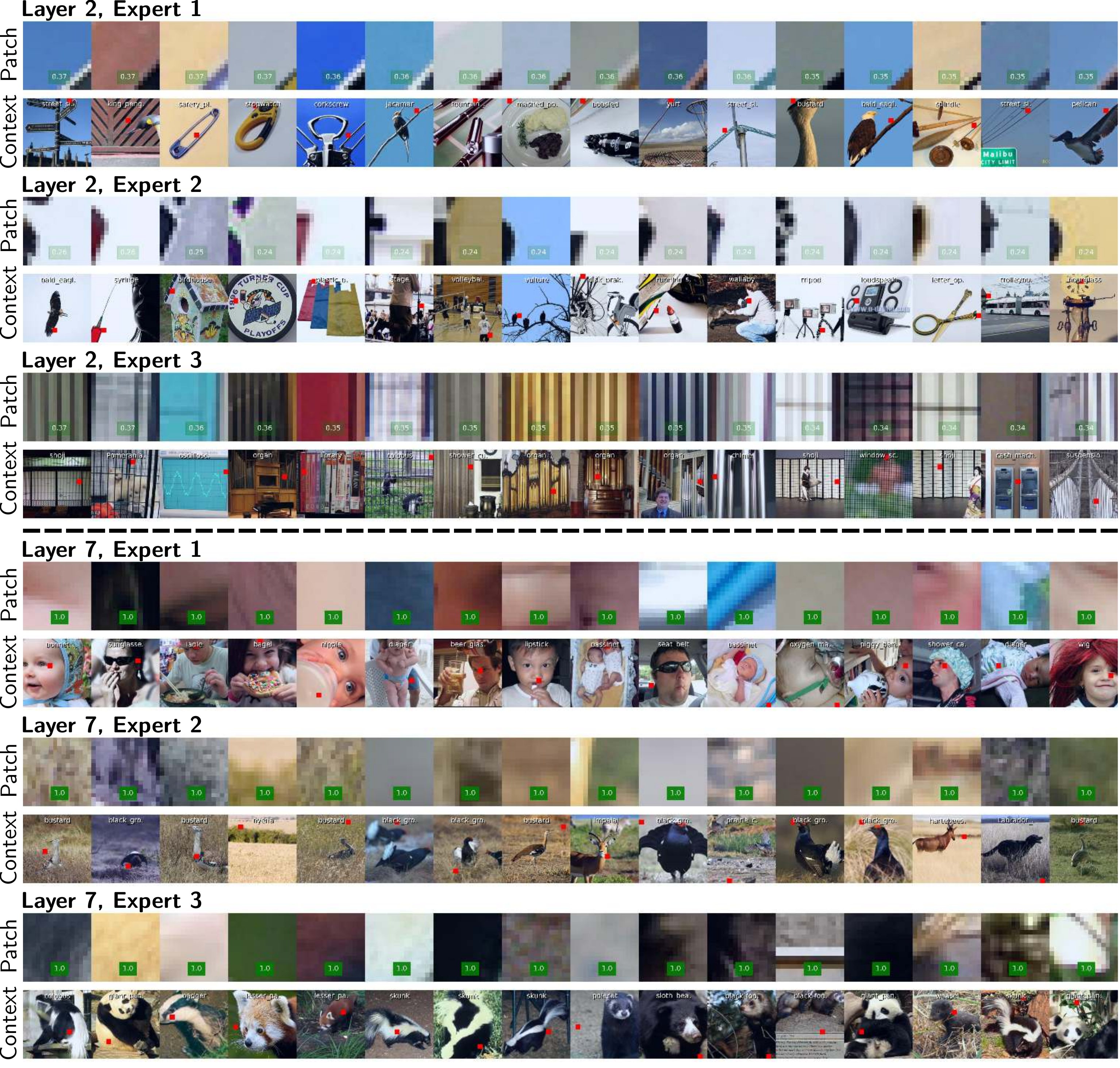}
    \caption{
    Top-activating patches (top rows) and their full images (second rows) for the first 3 experts across 2 \texttt{CP$\mu$MoE-e64} layers in $\mu$MoE MLP-mixer \cite{tolstikhin2021mlp} models--$\mu$MoE blocks exhibit coarse-grained specialism (e.g. texture) earlier and more fine-grained specialism (e.g. objects) deeper in the network.}
    \label{fig:expert-slices-mlpmixer}
\end{figure}

Finally, we train from scratch $12$ layer $124$M-parameter GPT-2 \cite{radford2019language} LLMs on OpenWebText \cite{Gokaslan2019OpenWeb} for the language domain and $8$ layer \texttt{S-16} variant\footnote{The \texttt{S-16} model is the largest configuration that fits into 4x80GB A100 GPUs using the original paper's batch size of 4096.} MLP-Mixers \cite{tolstikhin2021mlp} on ImageNET1k \cite{deng2009imagenet} for vision.
We replace \textit{every} MLP block's 2 linear layers with 2 $\mu$MoE layers. Each token $t$'s input vector $\mathbf{z}_t\in\mathbb{R}^I$ is therefore transformed with $\mu$MoE blocks of the form:
\begin{equation*}
    \mathbf{y}_t =
    \sum_{n_2=1}^N\sum_{h=1}^{H} \mathbf{w}^{(2)}_{{n_2}h:}
    \texttt{GELU}
    \bigg(
        \sum_{{n_1}=1}^N\sum_{i=1}^I \mathbf{w}^{(1)}_{{n_1}i:} z_{ti} a_{t{n_1}}
    \bigg)_{h}
    a_{t{n_2}},
    \quad \mathbf{a}_t= \phi(\mathbf{G}^\top\mathbf{z}_t),
\end{equation*}
where $\mathbf{a}_t\in\mathbb{R}^N$ are the expert coefficients for each specific token and block, $H$ is the dimension of the block's hidden layer,
and $\mathcal{W}^{(1)}\in\mathbb{R}^{N\times I \times H}, \mathcal{W}^{(2)}\in\mathbb{R}^{N\times H \times O}$ are the (implicit) $\mu$MoE weight tensors for each of the two layers.
We manually set the $\mu$MoE ranks to parameter-match each original network and set the number of experts (per block) to $N=64$ for vision models and $N=256$ for LLMs.
Consequently, with this configuration, \textbf{each layer's $\boldsymbol\mu$MoE block performs computations with $N$ experts yet has the same parameter counts and FLOPs as a single, dense MLP block}.

\begin{figure}[]
    \centering
    \includegraphics[width=\linewidth]{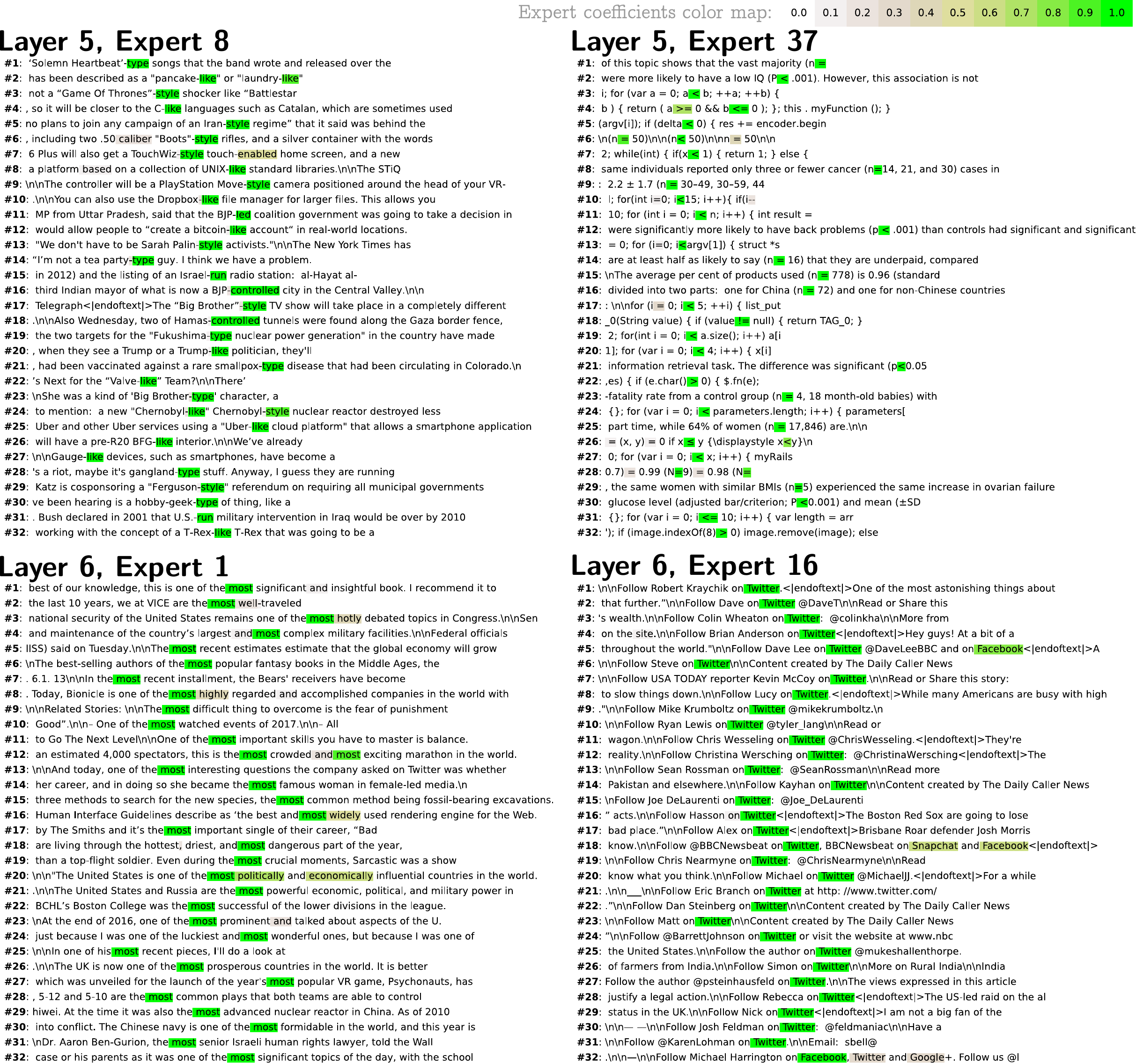}
    \caption{Top-activating generated tokens for 4 manually selected experts for GPT-2 trained with CP$\mu$MoE blocks at every layer (each token is highlighted by the coefficient of the expert in question), exhibiting specializations to concepts including compound adjectives and equality operators.
    }
    \label{fig:expert-slices-gpt}
\end{figure}

\paragraph{$\boldsymbol\mu$MoE-Mixer} For vision, our key findings are that earlier $\mu$MoE channel-mixing blocks' experts appear (qualitatively) to exhibit specialisms to colors, shapes, and textures, whilst later layers exhibit more object-specific specialization.
We plot the patches from the training set for which each expert most contributes its computation in \cref{fig:expert-slices-mlpmixer} for both a shallow and deep layer to illustrate this--earlier layers' experts contribute strongly to the processing of similar \textit{patches} (top rows, e.g. specific edges) whilst later layers' experts process tokens based more on the similarity of their surrounding semantic context (bottom rows, e.g. images of animals).
We further show in \cref{fig:mlp-mixer-subfigures} results for the first 2 experts across all 8 blocks where such scale-specific specialism is apparent across the entire network.

\paragraph{$\boldsymbol\mu$MoE-GPT2} For LLMs, we see promising qualitative evidence of experts specializing throughout a corpus of $1$M generated 100-token sequences.
At layer 5, for example, the generated tokens that use expert 8 with the highest coefficient are compound adjectives (\cref{fig:expert-slices-gpt}), whilst expert 37 most highly activates for equality and comparison operators in code and scientific text (please see examples of many unfiltered experts in \cref{fig:gpt-layer5-specialism,fig:gpt-layer5-specialism2}).
Whilst monosemanticity is not always attained, $\mu$MoE layers nonetheless facilitate a level of specialism not facilitated by dense MLP layers. %

One important result here is that $\mu$MoE networks in this setup are significantly more parameter-efficient than both dense and sparse MoEs with the same expert count, as shown in \cref{tab:n64-param-compare}. For example, GPT-2 models with $256$ sparse/dense MoE experts require a prohibitive $14.5$B MLP parameters alone, relative to just $57$M MLP parameters with $\mu$MoEs of the same expert counts.

\begin{table}[t]
    \centering
    \caption{Comparison of $\mu$MoEs and dense MLPs across different models and tasks. We use $N=64$ $\mu$MoE experts for the two vision tasks and $N=256$ for GPT2.
    MLP mixers and GPT2s are pre-trained for 300 epochs and 100k iterations respectively, whilst CLIP is fine-tuned for 10 epochs.
    }
    \label{tab:mumoe-acc}
    \resizebox{\textwidth}{!}{%
        \begin{tabular}{@{}lccccccc@{}}
        \toprule
        & \multicolumn{2}{c}{\textbf{MLP-mixer \texttt{S-16}} (ImageNET1k)} & \multicolumn{2}{c}{\textbf{GPT-2 NanoGPT} (OWT)} & \multicolumn{2}{c}{\textbf{CLIP \texttt{B-32}} (ImageNET1k)} \\ 
        & Val. acc. ($\uparrow$) & \#params & Val. loss ($\downarrow$) & \#params & Val. acc. ($\uparrow$) & \#params \\ \midrule
        MLPs & 70.31 & 18.5M & \textbf{2.876} & 124M & 77.99 & 769K \\
        \textbf{TR$\boldsymbol\mu$MoEs} & 71.26 & 18.3M & 2.886 & 124M & \textbf{78.71} & 771K \\
        \textbf{CP$\boldsymbol\mu$MoEs} & \textbf{71.29} & 18.6M & 2.893 & 124M & 78.07 & 769K \\ \bottomrule
        \end{tabular}
    }
\end{table}

\paragraph{$\boldsymbol\mu$MoE performance}
\label{sec:exp:performance}
\begin{wrapfigure}[9]{r}{0.43\textwidth}
    \centering
    \captionof{table}{MLP parameters required for networks with the same expert counts.}
    \label{tab:n64-param-compare}
    \resizebox{0.42\textwidth}{!}{%
        \begin{tabular}{@{}lcc@{}}
        \toprule
        & NanoGPT (\texttt{gpt2}) & MLP-Mixer (\texttt{S-16}) \\
        Model & $N=256$ & $N=64$ \\
        \midrule
        Dense/Sparse MoE & \mednumber{$14.5$B} & \mednumber{$1.13$B} \\
        \textbf{CP$\mu$MoE} & \mednumber{\textbf{$\textbf{57.0}$M}} & \mednumber{\textbf{$\textbf{17.7}$M}} \\
        \textbf{TR$\mu$MoE} & \mednumber{\textbf{$\textbf{57.4}$M}} & \mednumber{\textbf{$\textbf{17.4}$M}} \\ \bottomrule
        \end{tabular}
    }
\end{wrapfigure}

Finally, we substantiate our claim that networks pre-trained and fine-tuned with parameter-matched $\mu$MoE layers are competitive with their existing linear layer alternatives across multiple domains/machine learning tasks.
We present in \cref{tab:mumoe-acc} the performance results for MLP-Mixer \texttt{S-16} \cite{tolstikhin2021mlp}, NanoGPT \texttt{GPT-2} \cite{radford2019language}, and (fine-tuned) CLIP \texttt{ViT-B-32} \cite{radford2021learning} models on the OWT and ImageNET1k datasets.
Following \cref{sec:prune-qualitative}, we replace all linear layers with $\mu$MoE blocks (and a single $\mu$MoE final layer for fine-tuning CLIP). We initialize all linear layers following the default PyTorch $U[-k,k]$ initialization for a fair comparison.
Please see \cref{sec:app:config} for experimental details and learning curves, and \cref{sec:app:additional-performance} for experiments with varying expert count and hierarchical $\mu$MoEs.
Crucially, whilst $\mu$MoE layers provide additional interpretability benefits through scalable expert specialization, they do not sacrifice accuracy when parameter-matched to MLP blocks, as seen from the comparable performance.

\section{Conclusion}
\label{sec:conclusion}

In this paper, we introduced the Multilinear Mixture of Experts layer ($\mu$MoE).
We demonstrated that larger expert counts lead to increased specialization, and how $\mu$MoE layers make this computationally tractable through factorized forward passes.
$\mu$MoEs scale to large expert counts much more gracefully than existing MoEs, yet avoid the issues from popular gating mechanisms.
As a further practical example of $\mu$MoE's task decomposition, we illustrated how manual guided edits can be made to correct bias towards demographic subpopulations in fine-tuned foundation models.
Having also shown matching performance in addition to expert specialism in both large vision and language models, we believe $\mu$MoE layers constitute an important step towards facilitating increasingly performant models that do not trade off fairness/interpretability for accuracy.

\paragraph{Limitations}
Firstly, it is important to state again that our quantitative evaluation only captures expert behavior on the test set, not out-of-distribution data \cite{casper2023critique,bolukbasi2021interpretabilityillusion}.
Furthermore, expert specialism in large models is only demonstrated qualitatively (through the expert coefficients) due to the absence of fine-grained labels. Developing ways of quantifying fine-grained expert specialism is an important direction for future research.
Finally, our experimental results demonstrated comparable accuracies of $\mu$MoE networks only for models with parameter counts on the order of 100 million. Where resources permit, future work should explore the scalability of expert specialization and performance of $\mu$MoEs in even larger-scale LLMs.

\bibliography{bib}

\begin{thebibliography}{90}
\providecommand{\natexlab}[1]{#1}
\providecommand{\url}[1]{\texttt{#1}}
\expandafter\ifx\csname urlstyle\endcsname\relax
  \providecommand{\doi}[1]{doi: #1}\else
  \providecommand{\doi}{doi: \begingroup \urlstyle{rm}\Url}\fi

\bibitem[Jacobs et~al.(1991{\natexlab{a}})Jacobs, Jordan, Nowlan, and Hinton]{jacobs1991adaptive}
Robert~A Jacobs, Michael~I Jordan, Steven~J Nowlan, and Geoffrey~E Hinton.
\newblock Adaptive mixtures of local experts.
\newblock \emph{Neural computation}, 3\penalty0 (1):\penalty0 79--87, 1991{\natexlab{a}}.

\bibitem[Jiang et~al.(2024)Jiang, Sablayrolles, Roux, Mensch, Savary, Bamford, Chaplot, de~las Casas, Hanna, Bressand, Lengyel, Bour, Lample, Lavaud, Saulnier, Lachaux, Stock, Subramanian, Yang, Antoniak, Scao, Gervet, Lavril, Wang, Lacroix, and Sayed]{jiang2024mixtral}
Albert~Q. Jiang, Alexandre Sablayrolles, Antoine Roux, Arthur Mensch, Blanche Savary, Chris Bamford, Devendra~Singh Chaplot, Diego de~las Casas, Emma~Bou Hanna, Florian Bressand, Gianna Lengyel, Guillaume Bour, Guillaume Lample, Lélio~Renard Lavaud, Lucile Saulnier, Marie-Anne Lachaux, Pierre Stock, Sandeep Subramanian, Sophia Yang, Szymon Antoniak, Teven~Le Scao, Théophile Gervet, Thibaut Lavril, Thomas Wang, Timothée Lacroix, and William~El Sayed.
\newblock Mixtral of experts, 2024.

\bibitem[Lepikhin et~al.(2021)Lepikhin, Lee, Xu, Chen, Firat, Huang, Krikun, Shazeer, and Chen]{lepikhin2021gshard}
Dmitry Lepikhin, HyoukJoong Lee, Yuanzhong Xu, Dehao Chen, Orhan Firat, Yanping Huang, Maxim Krikun, Noam Shazeer, and Zhifeng Chen.
\newblock {GS}hard: Scaling giant models with conditional computation and automatic sharding.
\newblock In \emph{Int. Conf. Learn. Represent. (ICLR)}, 2021.

\bibitem[Fedus et~al.(2022)Fedus, Zoph, and Shazeer]{fedus2022switch}
William Fedus, Barret Zoph, and Noam Shazeer.
\newblock Switch transformers: Scaling to trillion parameter models with simple and efficient sparsity.
\newblock \emph{The Journal of Machine Learning Research}, 23\penalty0 (1):\penalty0 5232--5270, 2022.

\bibitem[Gale et~al.(2023)Gale, Narayanan, Young, and Zaharia]{gale2023megablocks}
Trevor Gale, Deepak Narayanan, Cliff Young, and Matei Zaharia.
\newblock Megablocks: Efficient sparse training with mixture-of-experts.
\newblock \emph{Proceedings of Machine Learning and Systems}, 5, 2023.

\bibitem[Riquelme et~al.(2021)Riquelme, Puigcerver, Mustafa, Neumann, Jenatton, Susano~Pinto, Keysers, and Houlsby]{riquelme2021scaling}
Carlos Riquelme, Joan Puigcerver, Basil Mustafa, Maxim Neumann, Rodolphe Jenatton, Andr{\'e} Susano~Pinto, Daniel Keysers, and Neil Houlsby.
\newblock Scaling vision with sparse mixture of experts.
\newblock \emph{Adv. Neural Inform. Process. Syst. (NeurIPS)}, 34:\penalty0 8583--8595, 2021.

\bibitem[Mustafa et~al.(2022)Mustafa, Ruiz, Puigcerver, Jenatton, and Houlsby]{mustafa2022multimodal}
Basil Mustafa, Carlos~Riquelme Ruiz, Joan Puigcerver, Rodolphe Jenatton, and Neil Houlsby.
\newblock Multimodal contrastive learning with {LIM}oe: the language-image mixture of experts.
\newblock In Alice~H. Oh, Alekh Agarwal, Danielle Belgrave, and Kyunghyun Cho, editors, \emph{Adv. Neural Inform. Process. Syst. (NeurIPS)}, 2022.

\bibitem[Ha et~al.(2017)Ha, Dai, and Le]{ha2017hypernetworks}
David Ha, Andrew~M. Dai, and Quoc~V. Le.
\newblock Hypernetworks.
\newblock In \emph{Int. Conf. Learn. Represent. (ICLR)}, 2017.

\bibitem[Vaswani et~al.(2017)Vaswani, Shazeer, Parmar, Uszkoreit, Jones, Gomez, Kaiser, and Polosukhin]{vaswani2017attention}
Ashish Vaswani, Noam Shazeer, Niki Parmar, Jakob Uszkoreit, Llion Jones, Aidan~N Gomez, {\L}ukasz Kaiser, and Illia Polosukhin.
\newblock Attention is all you need.
\newblock \emph{Adv. Neural Inform. Process. Syst. (NeurIPS)}, 30, 2017.

\bibitem[Han et~al.(2021)Han, Huang, Song, Yang, Wang, and Wang]{han2021dynamic}
Yizeng Han, Gao Huang, Shiji Song, Le~Yang, Honghui Wang, and Yulin Wang.
\newblock Dynamic neural networks: A survey.
\newblock \emph{IEEE Trans. Pattern Anal. Mach. Intell. (TPAMI)}, 44\penalty0 (11):\penalty0 7436--7456, 2021.

\bibitem[Chen et~al.(2020)Chen, Dai, Liu, Chen, Yuan, and Liu]{chen2020dynamic}
Yinpeng Chen, Xiyang Dai, Mengchen Liu, Dongdong Chen, Lu~Yuan, and Zicheng Liu.
\newblock Dynamic convolution: Attention over convolution kernels.
\newblock In \emph{IEEE Conf. Comput. Vis. Pattern Recog. (CVPR)}, pages 11030--11039, 2020.

\bibitem[Jacobs et~al.(1991{\natexlab{b}})Jacobs, Jordan, and Barto]{jacobs1991task}
Robert~A Jacobs, Michael~I Jordan, and Andrew~G Barto.
\newblock Task decomposition through competition in a modular connectionist architecture: The what and where vision tasks.
\newblock \emph{Cognitive science}, 15\penalty0 (2):\penalty0 219--250, 1991{\natexlab{b}}.

\bibitem[Lipton(2018)]{lipton2018mythos}
Zachary~C. Lipton.
\newblock The mythos of model interpretability.
\newblock \emph{Communications of the ACM}, 61\penalty0 (10):\penalty0 36–43, September 2018.
\newblock ISSN 1557-7317.

\bibitem[Ribeiro et~al.(2016)Ribeiro, Singh, and Guestrin]{ribeiro2016should}
Marco~Tulio Ribeiro, Sameer Singh, and Carlos Guestrin.
\newblock " why should i trust you?" explaining the predictions of any classifier.
\newblock In \emph{Proceedings of the 22nd ACM SIGKDD international conference on knowledge discovery and data mining}, pages 1135--1144, 2016.

\bibitem[Mao et~al.(2023)Mao, Deng, Yao, Ye, Kawaguchi, and Zou]{mao2023lastlayer}
Yuzhen Mao, Zhun Deng, Huaxiu Yao, Ting Ye, Kenji Kawaguchi, and James Zou.
\newblock Last-layer fairness fine-tuning is simple and effective for neural networks.
\newblock In \emph{Proceedings of the 2nd Workshop on Spurious Correlations, Invariance and Stability at the International Conference on Machine Learning (ICML 2023)}, 2023.

\bibitem[Cherepanova et~al.(2021)Cherepanova, Nanda, Goldblum, Dickerson, and Goldstein]{cherepanova2021technical}
Valeriia Cherepanova, Vedant Nanda, Micah Goldblum, John~P Dickerson, and Tom Goldstein.
\newblock Technical challenges for training fair neural networks.
\newblock \emph{arXiv preprint arXiv:2102.06764}, 2021.

\bibitem[Shazeer et~al.(2017)Shazeer, Mirhoseini, Maziarz, Davis, Le, Hinton, and Dean]{shazeer2017}
Noam Shazeer, *Azalia Mirhoseini, *Krzysztof Maziarz, Andy Davis, Quoc Le, Geoffrey Hinton, and Jeff Dean.
\newblock Outrageously large neural networks: The sparsely-gated mixture-of-experts layer.
\newblock In \emph{Int. Conf. Learn. Represent. (ICLR)}, 2017.

\bibitem[Mohammed et~al.(2022)Mohammed, Liu, and Raffel]{mohammed2022models}
Muqeeth Mohammed, Haokun Liu, and Colin Raffel.
\newblock Models with conditional computation learn suboptimal solutions.
\newblock In \emph{I Can't Believe It's Not Better Workshop: Understanding Deep Learning Through Empirical Falsification}, 2022.

\bibitem[Puigcerver et~al.(2024)Puigcerver, Riquelme, Mustafa, and Houlsby]{puigcerver2024sparse}
Joan Puigcerver, Carlos Riquelme, Basil Mustafa, and Neil Houlsby.
\newblock From sparse to soft mixtures of experts.
\newblock In \emph{Int. Conf. Learn. Represent. (ICLR)}, 2024.

\bibitem[Elazar et~al.(2021)Elazar, Ravfogel, Jacovi, and Goldberg]{elazar2021amnesic}
Yanai Elazar, Shauli Ravfogel, Alon Jacovi, and Yoav Goldberg.
\newblock Amnesic probing: Behavioral explanation with amnesic counterfactuals.
\newblock \emph{Transactions of the Association for Computational Linguistics}, 9:\penalty0 160--175, 2021.

\bibitem[Bengio et~al.(2015)Bengio, Bacon, Pineau, and Precup]{bengio2015conditional}
Emmanuel Bengio, Pierre-Luc Bacon, Joelle Pineau, and Doina Precup.
\newblock Conditional computation in neural networks for faster models.
\newblock In \emph{Int. Conf. Mach. Learn. Worksh. (ICMLW)}, 2015.

\bibitem[Jordan and Jacobs(1993)]{jordan1993hierarchical}
M.I. Jordan and R.A. Jacobs.
\newblock Hierarchical mixtures of experts and the em algorithm.
\newblock In \emph{Proceedings of 1993 International Conference on Neural Networks (IJCNN-93-Nagoya, Japan)}, volume~2, pages 1339--1344 vol.2, 1993.
\newblock \doi{10.1109/IJCNN.1993.716791}.

\bibitem[Eigen et~al.(2013)Eigen, Ranzato, and Sutskever]{Eigen2013LearningFR}
David Eigen, Marc'Aurelio Ranzato, and Ilya Sutskever.
\newblock Learning factored representations in a deep mixture of experts.
\newblock In \emph{Int. Conf. Mach. Learn. Worksh. (ICMLW)}, volume abs/1312.4314, 2013.

\bibitem[Yang et~al.(2019)Yang, Bender, Le, and Ngiam]{yang2019condconv}
Brandon Yang, Gabriel Bender, Quoc~V Le, and Jiquan Ngiam.
\newblock Condconv: Conditionally parameterized convolutions for efficient inference.
\newblock \emph{Adv. Neural Inform. Process. Syst. (NeurIPS)}, 32, 2019.

\bibitem[Du et~al.(2022)Du, Huang, Dai, Tong, Lepikhin, Xu, Krikun, Zhou, Yu, Firat, et~al.]{du2022glam}
Nan Du, Yanping Huang, Andrew~M Dai, Simon Tong, Dmitry Lepikhin, Yuanzhong Xu, Maxim Krikun, Yanqi Zhou, Adams~Wei Yu, Orhan Firat, et~al.
\newblock Glam: Efficient scaling of language models with mixture-of-experts.
\newblock In \emph{Int. Conf. Mach. Learn. (ICML)}, pages 5547--5569. PMLR, 2022.

\bibitem[Gupta et~al.(2022)Gupta, Mukherjee, Subudhi, Gonzalez, Jose, Awadallah, and Gao]{gupta2022multitasklearners}
Shashank Gupta, Subhabrata Mukherjee, Krishan Subudhi, Eduardo Gonzalez, Damien Jose, Ahmed~H Awadallah, and Jianfeng Gao.
\newblock Sparsely activated mixture-of-experts are robust multi-task learners.
\newblock \emph{arXiv preprint arXiv:2204.07689}, 2022.

\bibitem[Gururangan et~al.(2022)Gururangan, Lewis, Holtzman, Smith, and Zettlemoyer]{Gururangan2022domainlm}
Suchin Gururangan, Mike Lewis, Ari Holtzman, Noah Smith, and Luke Zettlemoyer.
\newblock Demix layers: Disentangling domains for modular language modeling.
\newblock In \emph{Proceedings of the 2022 Conference of the North American Chapter of the Association for Computational Linguistics: Human Language Technologies}. Association for Computational Linguistics, 2022.
\newblock \doi{10.18653/v1/2022.naacl-main.407}.

\bibitem[Ismail et~al.(2023)Ismail, Arik, Yoon, Taly, Feizi, and Pfister]{ismail2023interpretablemoe}
Aya~Abdelsalam Ismail, Sercan~O Arik, Jinsung Yoon, Ankur Taly, Soheil Feizi, and Tomas Pfister.
\newblock Interpretable mixture of experts.
\newblock \emph{Transactions on Machine Learning Research}, 2023.
\newblock ISSN 2835-8856.

\bibitem[Lewis et~al.(2021)Lewis, Bhosale, Dettmers, Goyal, and Zettlemoyer]{Lewis2021BASELS}
Mike Lewis, Shruti Bhosale, Tim Dettmers, Naman Goyal, and Luke Zettlemoyer.
\newblock Base layers: Simplifying training of large, sparse models.
\newblock In \emph{Int. Conf. Mach. Learn. (ICML)}, 2021.

\bibitem[Wu et~al.(2022)Wu, Liu, Chen, Chen, Dai, and Yuan]{wu2022residual}
Lemeng Wu, Mengchen Liu, Yinpeng Chen, Dongdong Chen, Xiyang Dai, and Lu~Yuan.
\newblock Residual mixture of experts, 2022.

\bibitem[Xue et~al.(2022)Xue, Shi, Wei, Lou, Liu, and You]{xue2022go}
Fuzhao Xue, Ziji Shi, Futao Wei, Yuxuan Lou, Yong Liu, and Yang You.
\newblock Go wider instead of deeper.
\newblock In \emph{Conf. on Artifi. Intel. (AAAI)}, volume~36, pages 8779--8787, 2022.

\bibitem[Davis and Arel(2013)]{davis2013low}
Andrew Davis and Itamar Arel.
\newblock Low-rank approximations for conditional feedforward computation in deep neural networks.
\newblock \emph{arXiv preprint arXiv:1312.4461}, 2013.

\bibitem[Li et~al.(2021)Li, Chen, Dai, mengchen liu, Chen, Yu, Yuan, Liu, Chen, and Vasconcelos]{li2021revisiting}
Yunsheng Li, Yinpeng Chen, Xiyang Dai, mengchen liu, Dongdong Chen, Ye~Yu, Lu~Yuan, Zicheng Liu, Mei Chen, and Nuno Vasconcelos.
\newblock Revisiting dynamic convolution via matrix decomposition.
\newblock In \emph{Int. Conf. Learn. Represent. (ICLR)}, 2021.

\bibitem[Gao et~al.(2022)Gao, Liu, Zhao, Lu, and Wen]{gao2022ttparameter}
Ze-Feng Gao, Peiyu Liu, Wayne~Xin Zhao, Zhong-Yi Lu, and Ji-Rong Wen.
\newblock Parameter-efficient mixture-of-experts architecture for pre-trained language models.
\newblock In \emph{Proceedings of the 29th International Conference on Computational Linguistics}, Gyeongju, Republic of Korea, October 2022. International Committee on Computational Linguistics.

\bibitem[Oseledets(2011)]{Oseledets2011TensorTrainD}
I.~Oseledets.
\newblock Tensor-train decomposition.
\newblock \emph{SIAM J. Sci. Comput.}, 33:\penalty0 2295--2317, 2011.

\bibitem[Hu et~al.(2021)Hu, Shen, Wallis, Allen-Zhu, Li, Wang, Wang, and Chen]{hu2021lora}
Edward~J Hu, Yelong Shen, Phillip Wallis, Zeyuan Allen-Zhu, Yuanzhi Li, Shean Wang, Lu~Wang, and Weizhu Chen.
\newblock Lora: Low-rank adaptation of large language models.
\newblock \emph{arXiv preprint arXiv:2106.09685}, 2021.

\bibitem[Novikov et~al.(2015)Novikov, Podoprikhin, Osokin, and Vetrov]{novikov2015tensorizing}
Alexander Novikov, Dmitrii Podoprikhin, Anton Osokin, and Dmitry~P Vetrov.
\newblock Tensorizing neural networks.
\newblock \emph{Adv. Neural Inform. Process. Syst. (NeurIPS)}, 28, 2015.

\bibitem[Garipov et~al.(2016)Garipov, Podoprikhin, Novikov, and Vetrov]{garipov2016ultimate}
Timur Garipov, Dmitry Podoprikhin, Alexander Novikov, and Dmitry Vetrov.
\newblock Ultimate tensorization: compressing convolutional and fc layers alike.
\newblock \emph{arXiv preprint arXiv:1611.03214}, 2016.

\bibitem[Novikov et~al.(2017)Novikov, Trofimov, and Oseledets]{alex2016exponential}
Alexander Novikov, Mikhail Trofimov, and Ivan Oseledets.
\newblock Exponential machines.
\newblock In \emph{Int. Conf. Learn. Represent. Worksh.}, 2017.

\bibitem[Georgopoulos et~al.(2021)Georgopoulos, Oldfield, Nicolaou, Panagakis, and Pantic]{georgopoulos2021mitigating}
Markos Georgopoulos, James Oldfield, Mihalis~A Nicolaou, Yannis Panagakis, and Maja Pantic.
\newblock Mitigating demographic bias in facial datasets with style-based multi-attribute transfer.
\newblock \emph{Int. J. Comput. Vis. (IJCV)}, 129\penalty0 (7):\penalty0 2288--2307, 2021.

\bibitem[Babiloni et~al.(2020)Babiloni, Marras, Slabaugh, and Zafeiriou]{babiloni2020tesa}
Francesca Babiloni, Ioannis Marras, Gregory Slabaugh, and Stefanos Zafeiriou.
\newblock Tesa: Tensor element self-attention via matricization.
\newblock In \emph{IEEE Conf. Comput. Vis. Pattern Recog. (CVPR)}, pages 13945--13954, 2020.

\bibitem[Georgopoulos et~al.(2020)Georgopoulos, Chrysos, Pantic, and Panagakis]{georgopoulos2020multilinear}
Markos Georgopoulos, Grigorios Chrysos, Maja Pantic, and Yannis Panagakis.
\newblock Multilinear latent conditioning for generating unseen attribute combinations.
\newblock In \emph{Int. Conf. Mach. Learn. (ICML)}, 2020.

\bibitem[Cheng et~al.(2024)Cheng, Chrysos, Georgopoulos, and Cevher]{cheng2024multilinear}
Yixin Cheng, Grigorios~G. Chrysos, Markos Georgopoulos, and Volkan Cevher.
\newblock Multilinear operator networks, 2024.

\bibitem[Kossaifi et~al.(2020)Kossaifi, Toisoul, Bulat, Panagakis, Hospedales, and Pantic]{kossaifi2020factconv}
Jean Kossaifi, Antoine Toisoul, Adrian Bulat, Yannis Panagakis, Timothy~M. Hospedales, and Maja Pantic.
\newblock Factorized higher-order cnns with an application to spatio-temporal emotion estimation.
\newblock In \emph{IEEE Conf. Comput. Vis. Pattern Recog. (CVPR)}. IEEE, June 2020.

\bibitem[Bulat et~al.(2020)Bulat, Kossaifi, Tzimiropoulos, and Pantic]{bulat2020incremental}
Adrian Bulat, Jean Kossaifi, Georgios Tzimiropoulos, and Maja Pantic.
\newblock Incremental multi-domain learning with network latent tensor factorization.
\newblock In \emph{Conf. on Artifi. Intel. (AAAI)}, volume~34, pages 10470--10477, 2020.

\bibitem[Yang and Hospedales(2017)]{yang2017multitask}
Yongxin Yang and Timothy~M. Hospedales.
\newblock Deep multi-task representation learning: A tensor factorisation approach.
\newblock In \emph{Int. Conf. Learn. Represent. (ICLR)}, 2017.

\bibitem[Chrysos et~al.(2020)Chrysos, Moschoglou, Bouritsas, Panagakis, Deng, and Zafeiriou]{chrysos2020p}
Grigorios~G Chrysos, Stylianos Moschoglou, Giorgos Bouritsas, Yannis Panagakis, Jiankang Deng, and Stefanos Zafeiriou.
\newblock P-nets: Deep polynomial neural networks.
\newblock In \emph{IEEE Conf. Comput. Vis. Pattern Recog. (CVPR)}, pages 7325--7335, 2020.

\bibitem[Chrysos et~al.(2021)Chrysos, Moschoglou, Bouritsas, Deng, Panagakis, and Zafeiriou]{chrysos2021pami}
Grigorios~G. Chrysos, Stylianos Moschoglou, Giorgos Bouritsas, Jiankang Deng, Yannis Panagakis, and Stefanos~P Zafeiriou.
\newblock Deep polynomial neural networks.
\newblock \emph{IEEE Trans. Pattern Anal. Mach. Intell. (TPAMI)}, page 1–1, 2021.
\newblock ISSN 1939-3539.

\bibitem[Babiloni et~al.(2021)Babiloni, Marras, Kokkinos, Deng, Chrysos, and Zafeiriou]{babiloni2021poly}
Francesca Babiloni, Ioannis Marras, Filippos Kokkinos, Jiankang Deng, Grigorios Chrysos, and Stefanos Zafeiriou.
\newblock Poly-nl: Linear complexity non-local layers with 3rd order polynomials.
\newblock In \emph{Int. Conf. Comput. Vis. (ICCV)}, pages 10518--10528, 2021.

\bibitem[Kossaifi et~al.(2017)Kossaifi, Khanna, Lipton, Furlanello, and Anandkumar]{kossaifi2017tensor}
Jean Kossaifi, Aran Khanna, Zachary Lipton, Tommaso Furlanello, and Anima Anandkumar.
\newblock Tensor contraction layers for parsimonious deep nets.
\newblock In \emph{IEEE Conf. Comput. Vis. Pattern Recog. Worksh. (CVPRW)}, pages 26--32, 2017.

\bibitem[Babiloni et~al.(2023)Babiloni, Tanay, Deng, Maggioni, and Zafeiriou]{babiloni2023factorised}
Francesca Babiloni, Thomas Tanay, Jiankang Deng, Matteo Maggioni, and Stefanos Zafeiriou.
\newblock Factorized dynamic fully-connected layers for neural networks.
\newblock In \emph{Int. Conf. Comput. Vis. Worksh. (ICCVW)}, pages 1374--1383, October 2023.

\bibitem[Hitchcock(1927)]{Hitchcock1927TheEO}
Frank~Lauren Hitchcock.
\newblock The expression of a tensor or a polyadic as a sum of products.
\newblock \emph{Journal of Mathematics and Physics}, 6:\penalty0 164--189, 1927.

\bibitem[Kolda and Bader(2009)]{kolda2009tensorapplications}
Tamara~G. Kolda and Brett~W. Bader.
\newblock Tensor decompositions and applications.
\newblock \emph{SIAM Review}, 51\penalty0 (3):\penalty0 455--500, 2009.
\newblock \doi{10.1137/07070111X}.

\bibitem[Peters et~al.(2019)Peters, Niculae, and Martins]{peters2019entmax}
Ben Peters, Vlad Niculae, and Andr{\'e} F.~T. Martins.
\newblock Sparse sequence-to-sequence models.
\newblock In Anna Korhonen, David Traum, and Llu{\'\i}s M{\`a}rquez, editors, \emph{Proceedings of the 57th Annual Meeting of the Association for Computational Linguistics}, pages 1504--1519, Florence, Italy, July 2019. Association for Computational Linguistics.
\newblock \doi{10.18653/v1/P19-1146}.

\bibitem[Correia et~al.(2019)Correia, Niculae, and Martins]{correia2019entmax}
Gon{\c{c}}alo~M. Correia, Vlad Niculae, and Andr{\'e} F.~T. Martins.
\newblock Adaptively sparse transformers.
\newblock In Kentaro Inui, Jing Jiang, Vincent Ng, and Xiaojun Wan, editors, \emph{Proceedings of the 2019 Conference on Empirical Methods in Natural Language Processing and the 9th International Joint Conference on Natural Language Processing (EMNLP-IJCNLP)}, pages 2174--2184, Hong Kong, China, November 2019. Association for Computational Linguistics.
\newblock \doi{10.18653/v1/D19-1223}.

\bibitem[Carroll and Chang(1970)]{Carroll1970AnalysisOI}
J.~Douglas Carroll and Jih~Jie Chang.
\newblock Analysis of individual differences in multidimensional scaling via an n-way generalization of “eckart-young” decomposition.
\newblock \emph{Psychometrika}, 35:\penalty0 283--319, 1970.

\bibitem[fvc()]{fvcore}
fvcore: Flop counter for pytorch models.
\newblock \url{https://github.com/facebookresearch/fvcore}.
\newblock Accessed: 2024-05-16.

\bibitem[Zhao et~al.(2016)Zhao, Zhou, Xie, Zhang, and Cichocki]{Zhao2016TensorRD}
Qibin Zhao, Guoxu Zhou, Shengli Xie, Liqing Zhang, and Andrzej Cichocki.
\newblock Tensor ring decomposition.
\newblock \emph{ArXiv}, abs/1606.05535, 2016.

\bibitem[Sharkey(2023)]{sharkey2023technical}
Lee Sharkey.
\newblock A technical note on bilinear layers for interpretability.
\newblock \emph{arXiv preprint arXiv:2305.03452}, 2023.

\bibitem[Pearce et~al.(2024)Pearce, Dooms, and Rigg]{pearce2024weightbased}
Michael~T. Pearce, Thomas Dooms, and Alice Rigg.
\newblock Weight-based decomposition: A case for bilinear {MLP}s, 2024.

\bibitem[Radford et~al.(2021)Radford, Kim, Hallacy, Ramesh, Goh, Agarwal, Sastry, Askell, Mishkin, Clark, et~al.]{radford2021learning}
Alec Radford, Jong~Wook Kim, Chris Hallacy, Aditya Ramesh, Gabriel Goh, Sandhini Agarwal, Girish Sastry, Amanda Askell, Pamela Mishkin, Jack Clark, et~al.
\newblock Learning transferable visual models from natural language supervision.
\newblock In \emph{Int. Conf. Mach. Learn. (ICML)}, 2021.

\bibitem[Caron et~al.(2021)Caron, Touvron, Misra, J\'egou, Mairal, Bojanowski, and Joulin]{caron2021dino}
Mathilde Caron, Hugo Touvron, Ishan Misra, Herv\'e J\'egou, Julien Mairal, Piotr Bojanowski, and Armand Joulin.
\newblock Emerging properties in self-supervised vision transformers.
\newblock In \emph{Int. Conf. Comput. Vis. (ICCV)}, 2021.

\bibitem[Ilharco et~al.(2022)Ilharco, Wortsman, Gadre, Song, Hajishirzi, Kornblith, Farhadi, and Schmidt]{ilharco2022patching}
Gabriel Ilharco, Mitchell Wortsman, Samir~Yitzhak Gadre, Shuran Song, Hannaneh Hajishirzi, Simon Kornblith, Ali Farhadi, and Ludwig Schmidt.
\newblock Patching open-vocabulary models by interpolating weights.
\newblock \emph{Adv. Neural Inform. Process. Syst. (NeurIPS)}, 35:\penalty0 29262--29277, 2022.

\bibitem[Ilharco et~al.(2023)Ilharco, Ribeiro, Wortsman, Schmidt, Hajishirzi, and Farhadi]{ilharco2023editing}
Gabriel Ilharco, Marco~Tulio Ribeiro, Mitchell Wortsman, Ludwig Schmidt, Hannaneh Hajishirzi, and Ali Farhadi.
\newblock Editing models with task arithmetic.
\newblock In \emph{Int. Conf. Learn. Represent. (ICLR)}, 2023.

\bibitem[Elhage et~al.(2022)Elhage, Hume, Olsson, Schiefer, Henighan, Kravec, Hatfield-Dodds, Lasenby, Drain, Chen, et~al.]{elhage2022toy}
Nelson Elhage, Tristan Hume, Catherine Olsson, Nicholas Schiefer, Tom Henighan, Shauna Kravec, Zac Hatfield-Dodds, Robert Lasenby, Dawn Drain, Carol Chen, et~al.
\newblock Toy models of superposition.
\newblock \emph{arXiv preprint arXiv:2209.10652}, 2022.

\bibitem[R{\"a}uker et~al.(2023)R{\"a}uker, Ho, Casper, and Hadfield-Menell]{rauker2023toward}
Tilman R{\"a}uker, Anson Ho, Stephen Casper, and Dylan Hadfield-Menell.
\newblock Toward transparent ai: A survey on interpreting the inner structures of deep neural networks.
\newblock In \emph{2023 IEEE Conference on Secure and Trustworthy Machine Learning (SaTML)}, pages 464--483. IEEE, 2023.

\bibitem[Ravfogel et~al.(2021)Ravfogel, Prasad, Linzen, and Goldberg]{ravfogel2021counterfactual}
Shauli Ravfogel, Grusha Prasad, Tal Linzen, and Yoav Goldberg.
\newblock Counterfactual interventions reveal the causal effect of relative clause representations on agreement prediction.
\newblock In Arianna Bisazza and Omri Abend, editors, \emph{Proceedings of the 25th Conference on Computational Natural Language Learning}, pages 194--209, Online, November 2021. Association for Computational Linguistics.

\bibitem[Meng et~al.(2022)Meng, Bau, Andonian, and Belinkov]{meng2022locating}
Kevin Meng, David Bau, Alex Andonian, and Yonatan Belinkov.
\newblock Locating and editing factual associations in gpt.
\newblock \emph{Adv. Neural Inform. Process. Syst. (NeurIPS)}, 35:\penalty0 17359--17372, 2022.

\bibitem[Rudin(2019)]{rudin2019stop}
Cynthia Rudin.
\newblock Stop explaining black box machine learning models for high stakes decisions and use interpretable models instead.
\newblock \emph{Nature machine intelligence}, 1\penalty0 (5):\penalty0 206--215, 2019.

\bibitem[Casper(2023)]{casper2023critique}
Stephen Casper.
\newblock Broad critiques of interpretability research.
\newblock 2023.
\newblock URL \url{https://www.alignmentforum.org/s/a6ne2ve5uturEEQK7/p/gwG9uqw255gafjYN4}.

\bibitem[Hod et~al.(2021)Hod, Filan, Casper, Critch, and Russell]{hod2021quantifying}
Shlomi Hod, Daniel Filan, Stephen Casper, Andrew Critch, and Stuart Russell.
\newblock Quantifying local specialization in deep neural networks.
\newblock \emph{arXiv preprint arXiv:2110.08058}, 2021.

\bibitem[Buolamwini and Gebru(2018)]{buolamwini2018gender}
Joy Buolamwini and Timnit Gebru.
\newblock Gender shades: Intersectional accuracy disparities in commercial gender classification.
\newblock In \emph{Conference on fairness, accountability and transparency}, pages 77--91. PMLR, 2018.

\bibitem[Gebru et~al.(2021)Gebru, Morgenstern, Vecchione, Vaughan, Wallach, Iii, and Crawford]{gebru2021datasheets}
Timnit Gebru, Jamie Morgenstern, Briana Vecchione, Jennifer~Wortman Vaughan, Hanna Wallach, Hal~Daum{\'e} Iii, and Kate Crawford.
\newblock Datasheets for datasets.
\newblock \emph{Communications of the ACM}, 64\penalty0 (12):\penalty0 86--92, 2021.

\bibitem[Liu et~al.(2015)Liu, Luo, Wang, and Tang]{liu2015faceattributes}
Ziwei Liu, Ping Luo, Xiaogang Wang, and Xiaoou Tang.
\newblock Deep learning face attributes in the wild.
\newblock In \emph{Int. Conf. Comput. Vis. (ICCV)}, December 2015.

\bibitem[Jain et~al.(2023)Jain, Lawrence, Moitra, and Madry]{jain2023distilling}
Saachi Jain, Hannah Lawrence, Ankur Moitra, and Aleksander Madry.
\newblock Distilling model failures as directions in latent space.
\newblock In \emph{Int. Conf. Learn. Represent. (ICLR)}, 2023.

\bibitem[Hardt et~al.(2016)Hardt, Price, and Srebro]{hardt2016equality}
Moritz Hardt, Eric Price, and Nati Srebro.
\newblock Equality of opportunity in supervised learning.
\newblock In \emph{Adv. Neural Inform. Process. Syst. (NeurIPS)}, 2016.

\bibitem[Wang and Deng(2020)]{wang2020mitigating}
Mei Wang and Weihong Deng.
\newblock Mitigating bias in face recognition using skewness-aware reinforcement learning.
\newblock In \emph{IEEE Conf. Comput. Vis. Pattern Recog. (CVPR)}, pages 9322--9331, 2020.

\bibitem[Lahoti et~al.(2020)Lahoti, Beutel, Chen, Lee, Prost, Thain, Wang, and Chi]{lahoti2020fairness}
Preethi Lahoti, Alex Beutel, Jilin Chen, Kang Lee, Flavien Prost, Nithum Thain, Xuezhi Wang, and Ed~Chi.
\newblock Fairness without demographics through adversarially reweighted learning.
\newblock \emph{Adv. Neural Inform. Process. Syst. (NeurIPS)}, 33:\penalty0 728--740, 2020.

\bibitem[Alvi et~al.(2018)Alvi, Zisserman, and Nell{\aa}ker]{alvi2018turning}
Mohsan Alvi, Andrew Zisserman, and Christoffer Nell{\aa}ker.
\newblock Turning a blind eye: Explicit removal of biases and variation from deep neural network embeddings.
\newblock In \emph{Proceedings of the European Conference on Computer Vision (ECCV) Workshops}, 2018.

\bibitem[Tolstikhin et~al.(2021)Tolstikhin, Houlsby, Kolesnikov, Beyer, Zhai, Unterthiner, Yung, Steiner, Keysers, Uszkoreit, et~al.]{tolstikhin2021mlp}
Ilya~O Tolstikhin, Neil Houlsby, Alexander Kolesnikov, Lucas Beyer, Xiaohua Zhai, Thomas Unterthiner, Jessica Yung, Andreas Steiner, Daniel Keysers, Jakob Uszkoreit, et~al.
\newblock {MLP}-mixer: An all-{MLP} architecture for vision.
\newblock \emph{Adv. Neural Inform. Process. Syst. (NeurIPS)}, 34:\penalty0 24261--24272, 2021.

\bibitem[Radford et~al.(2019)Radford, Wu, Child, Luan, Amodei, and Sutskever]{radford2019language}
Alec Radford, Jeffrey Wu, Rewon Child, David Luan, Dario Amodei, and Ilya Sutskever.
\newblock Language models are unsupervised multitask learners.
\newblock \emph{OpenAI Blog}, 2019.
\newblock URL \url{https://cdn.openai.com/better-language-models/language_models_are_unsupervised_multitask_learners.pdf}.

\bibitem[Gokaslan and Cohen(2019)]{Gokaslan2019OpenWeb}
Aaron Gokaslan and Vanya Cohen.
\newblock Openwebtext corpus.
\newblock \url{http://Skylion007.github.io/OpenWebTextCorpus}, 2019.

\bibitem[Deng et~al.(2009)Deng, Dong, Socher, Li, Li, and Fei-Fei]{deng2009imagenet}
Jia Deng, Wei Dong, Richard Socher, Li-Jia Li, Kai Li, and Li~Fei-Fei.
\newblock Imagenet: A large-scale hierarchical image database.
\newblock In \emph{IEEE Conf. Comput. Vis. Pattern Recog. (CVPR)}, pages 248--255, 2009.

\bibitem[Bolukbasi et~al.(2021)Bolukbasi, Pearce, Yuan, Coenen, Reif, Vi{\'e}gas, and Wattenberg]{bolukbasi2021interpretabilityillusion}
Tolga Bolukbasi, Adam Pearce, Ann Yuan, Andy Coenen, Emily Reif, Fernanda Vi{\'e}gas, and Martin Wattenberg.
\newblock An interpretability illusion for bert.
\newblock \emph{arXiv preprint arXiv:2104.07143}, 2021.

\bibitem[Rogozhnikov(2022)]{rogozhnikov2022einops}
Alex Rogozhnikov.
\newblock Einops: Clear and reliable tensor manipulations with einstein-like notation.
\newblock In \emph{Int. Conf. Learn. Represent. (ICLR)}, 2022.

\bibitem[Tucker(1966)]{Tucker1966SomeMN}
Ledyard~R. Tucker.
\newblock Some mathematical notes on three-mode factor analysis.
\newblock \emph{Psychometrika}, 31:\penalty0 279--311, 1966.

\bibitem[Eckart and Young(1936)]{eckart1936approximation}
Carl Eckart and Gale Young.
\newblock The approximation of one matrix by another of lower rank.
\newblock \emph{Psychometrika}, 1\penalty0 (3):\penalty0 211--218, 1936.

\bibitem[Sharma et~al.(2024)Sharma, Ash, and Misra]{sharma2024the}
Pratyusha Sharma, Jordan~T. Ash, and Dipendra Misra.
\newblock The truth is in there: Improving reasoning in language models with layer-selective rank reduction.
\newblock In \emph{Int. Conf. Learn. Represent. (ICLR)}, 2024.

\bibitem[Wortsman et~al.(2022)Wortsman, Ilharco, Kim, Li, Kornblith, Roelofs, Lopes, Hajishirzi, Farhadi, Namkoong, et~al.]{wortsman2022robust}
Mitchell Wortsman, Gabriel Ilharco, Jong~Wook Kim, Mike Li, Simon Kornblith, Rebecca Roelofs, Raphael~Gontijo Lopes, Hannaneh Hajishirzi, Ali Farhadi, Hongseok Namkoong, et~al.
\newblock Robust fine-tuning of zero-shot models.
\newblock In \emph{IEEE Conf. Comput. Vis. Pattern Recog. (CVPR)}, pages 7959--7971, 2022.

\bibitem[Wang et~al.(2020)Wang, Qinami, Karakozis, Genova, Nair, Hata, and Russakovsky]{wang2020towards}
Zeyu Wang, Klint Qinami, Ioannis~Christos Karakozis, Kyle Genova, Prem Nair, Kenji Hata, and Olga Russakovsky.
\newblock Towards fairness in visual recognition: Effective strategies for bias mitigation.
\newblock In \emph{IEEE Conf. Comput. Vis. Pattern Recog. (CVPR)}, pages 8919--8928, 2020.

\end{thebibliography}

\newpage
\appendix

\addcontentsline{toc}{section}{Appendix} %
\part{Appendix} %
\parttoc %

\section{Broader impact}
\label{sec:app:broader-impact}
This paper presents work whose goal is to advance the field of \textit{interpretable} machine learning. Our goal is not to improve model capabilities but rather an orthogonal one of designing architectures more interpretable and controllable. As with many work with an interpretability focus, however, the $\mu$MoE layer could nonetheless facilitate the further development of SOTA models through its more expressive computation. We thus encourage the development of further guardrails against potentially harmful dual-uses of such technology. We release our code upon acceptance to facilitate further research along such lines.

\section{Fast $\boldsymbol\mu$MoE implementations}
\label{sec:app:implementations}

We here detail how to implement the fast forward passes of the $\mu$MoE models in a batch-wise manner, where each mini-batch element is a 2D matrix of shape $\mathbf{Z}\in\mathbb{R}^{T \times C}$ (with `token' and `channel' dimensions) with PyTorch and einops' \cite{rogozhnikov2022einops} einsum:

\subsection{CP$\boldsymbol\mu$MoE einsum implementation}
The CP$\mu$MoE forward pass can be implemented with:
\begin{python}
# CPmuMoE (r=CP rank, b=batch_dim, t=tokens,
# i=input_dim, o=output_dim, a[e]=expert_coefs, n*=expert_dims)
y = einsum(G3, a[0]@G1.T, z@G2.T, 'r o, b t r, b t r -> b t o')
\end{python}
And a two-level hierarchical CP$\mu$MoE with an additional factor matrix as:
\begin{python}
# CPmuMoE (r=CP rank, b=batch_dim, t=tokens,
# i=input_dim, o=output_dim, a[e]=expert_coefs, n*=expert_dims)
#################
# A 2-level hierarchical CPmuMoE, assuming Gi's of appropriate shape
y = einsum(G4, a[0]@G1.T, a[1]@G2.T, z@G3.T,
           'r o, b t r, b t r, b t r -> b t o')
\end{python}

\subsection{TR$\boldsymbol\mu$MoE einsum implementation}
TR$\mu$MoEs can be implemented with:
\begin{python}
# TRmuMoE (r*=TR ranks, b=batch_dim, t=tokens,
# i=input_dim, o=output_dim, a[e]=expert_coefs, n*=expert_dims)

# batched mode-2 tensor-vector products
f1 = einsum(a[0], G1, 'b t n1, r1 n1 r2 -> b t r1 r2')
f2 = einsum(z, G2, 'b t i, r2 i r3 -> b t r2 r3')

# batch-multiply f1@f2
fout = einsum(f1, f2, 'b t r1 r2, b t r2 r3 -> b t r1 r3')

# contract with final TR core
y = einsum(G3, fout, 'r3 o r1, b t r1 r3  -> b t o')
\end{python}
And a two-level hierarchical version with an additional TR-core as:
\begin{python}
# TRmuMoE (r*=TR ranks, b=batch_dim, t=tokens,
# i=input_dim, o=output_dim, a[e]=expert_coefs, n*=expert_dims)
#################
# A 2-level hierarchical TRmuMoE, assuming additional TR cores Gi
f1 = einsum(a[0], G1, 'b t n1, r1 n1 r2 -> b t r1 r2')
f2 = einsum(a[1], G2, 'b t n2, r2 n2 r3 -> b t r2 r3')
f3 = einsum(z, G3, 'b t i, r3 i r4 -> b t r3 r4')

# batch-multiply f1@f2@f3
fout = einsum(f1, f2, 'b t r1 r2, b t r2 r3 -> b t r1 r3')
fout = einsum(fout, f3, 'b t r1 r3, b t r3 r4 -> b t r1 r4')

# contract with final TR core
y = einsum(G4, fout, 'r4 o r1, b t r1 r4  -> b t o')
\end{python}

\section{$\boldsymbol\mu$MoE forward pass visualization}
\begin{figure}[h]
    \centering
    \includegraphics[width=1.0\linewidth]{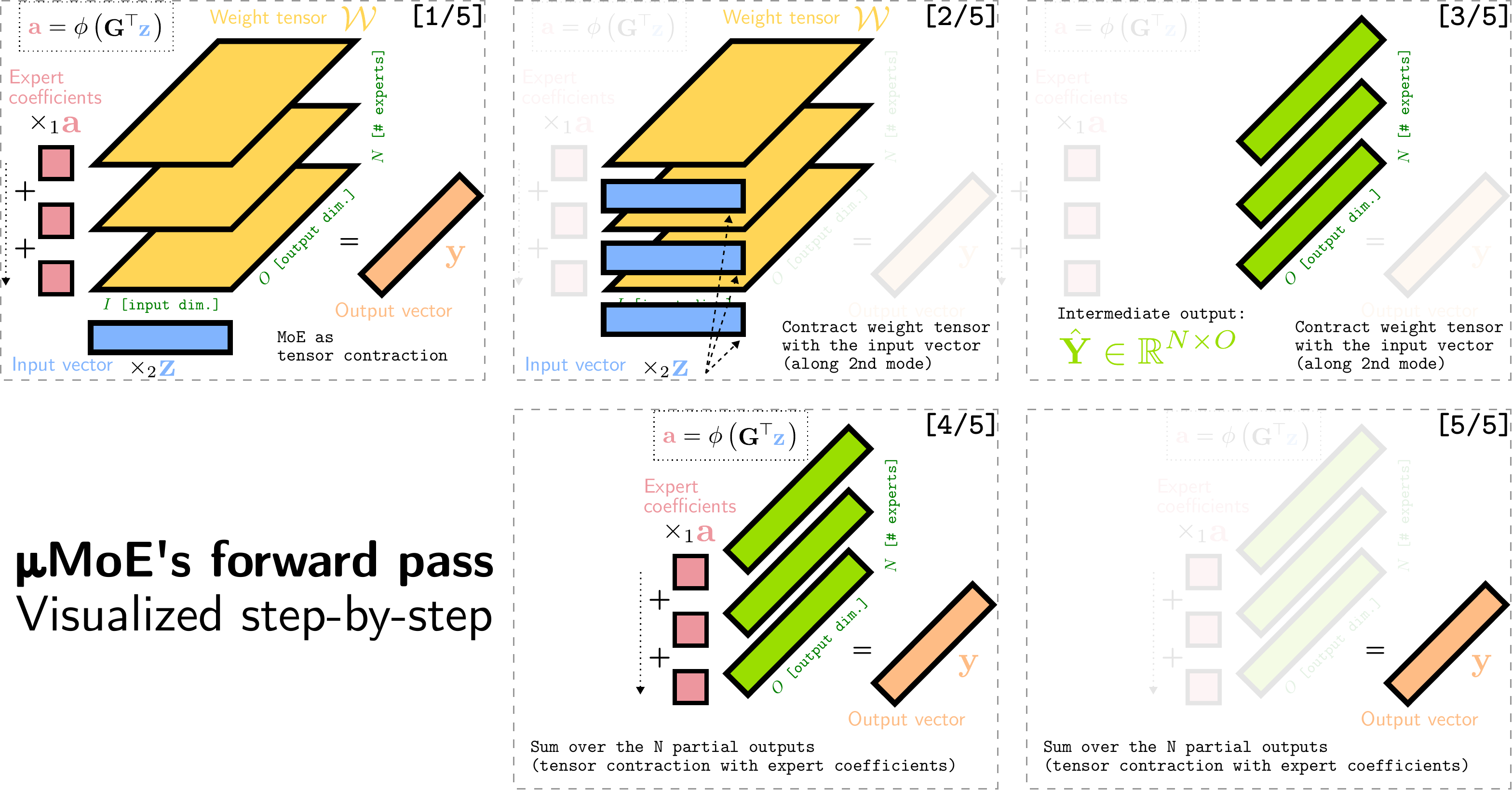}
    \caption{An intuitive visualization of the $\mu$MoE (unfactorized) forward pass, as visualized (as a series of tensor contractions) in 5 steps. Each step contributes to producing the output vector $\mathbf{y}\in\mathbb{R}^O$ either by contracting with the expert coefficients $\mathbf{a}\in\mathbb{R}^N$, or with the input vector $\mathbf{z}\in\mathbb{R}^I$, along the appropriate mode of the collective weight tensor $\mathcal{W}\in\mathbb{R}^{N\times I \times O}$.}
    \label{fig:step-by-step}
\end{figure}

For intuition, we provide a visualization in \cref{fig:step-by-step} of the step-by-step series of tensor contractions $\mathcal{W}\times_1\mathbf{a}\times_2\mathbf{z}\in\mathbb{R}^O$ that the $\mu$MoE computes (in non-factorized form).

\section{Decomposition choice, matrix rank, and computational cost}
\label{sec:app:decomps}

In this section we present a further detailed discussion of decomposition choice, validating our choices and comparing alternative options.
The computational costs of each fast $\mu$MoE forward pass and tensor-matrix rank relationships implications derived in this section are summarized in \cref{tab:decomp-tradeoff}.

\begin{table}[h]
\caption{A computational comparison of decomposition choice for $\mu$MoE layers and existing MoEs.}
\label{tab:decomp-tradeoff}
\centering
\resizebox{\linewidth}{!}{
\centering
  \begin{tabular}{lccccc}
    \toprule
    &\textbf{Param-efficient} & \textbf{Param-efficient} & & & \\
    &\textbf{(medium $N$)} & \textbf{(large $N$)} & \textbf{\# Parameters} & \textbf{Estimated \# FLOPs} & \textbf{Max. expert matrix rank} \\
    \midrule
    Dense MoE & \nomark & \nomark & $NIO$ & $NIO$ & $\min\{I,O\}$ \\
    Sparse MoE & \nomark & \nomark & $NIO$ & $KIO$ & $\min\{I,O\}$ \\
    \textbf{CP$\boldsymbol\mu$MoE} & \yesmark & \sosomark & $R(N+I+O)$ & $R(N+I+O)$ & $\min\{I,O,R\}$ \\
    \textbf{TR$\boldsymbol\mu$MoE} & \yesmark & \yesmark & $R_1NR_2 + R_2IR_3 + R_3OR_1$ & $R_2IR_3 + R_1NR_2 + R_1R_2R_3 + R_1OR_3$ & $\min\big\{ R_3\cdot \min\{R_1,R_2\}, I, O \big\}$ \\
    \hline
  \end{tabular}
}
\end{table}

\subsection{Tensor ranks to matrix rank}

One important consideration is how the chosen tensor ranks bound the resulting experts' matrix rank in $\mu$MoE layers.
Here, we derive the matrix ranks as a function of tensor ranks for each model in turn.

\subsubsection{CP$\boldsymbol\mu$MoEs: rank analysis}
CP$\mu$MoEs are parameterized by factor matrices $\mathbf{U}^{(1)}\in\mathbb{R}^{R \times N}, \mathbf{U}^{(2)}\in\mathbb{R}^{R \times I}, \mathbf{U}^{(3)}\in\mathbb{R}^{R \times O}$ for chosen CP-rank $R$.
Following \textit{Section 3} of \citet{kolda2009tensorapplications} which provides the matricization/unfolding of CP tensors,
we can write expert $n$'s weight matrix as
\begin{align}
    \mathbf{W}_n = {\mathbf{U}^{(2)}}^\top \left( {\mathbf{U}_{:n}^{(1)}}^\top \odot {\mathbf{U}^{(3)}}^\top \right)^\top\in\mathbb{R}^{I \times O},
\end{align}
where $\odot$ is the Khatri-Rao product \cite{kolda2009tensorapplications}, and
$\mathbf{U}_{:n}^{(1)}\in\mathbb{R}^{R\times 1}$ is the column of the factor matrix associated with expert $n$ (including a singleton dimension for the Khatri-Rao product to be well-defined).
Through the linear algebra rank inequality for matrix products, we have
\begin{align}
    \text{rank}(\mathbf{W}_{n})=\text{rank}\left({\mathbf{U}^{(2)}}^\top \left( {\mathbf{U}_{:n}^{(1)}}^\top \odot {\mathbf{U}^{(3)}}^\top \right)^\top\right) \leq \min \bigg\{
    \text{rank}(\underbrace{{\mathbf{U}^{(2)}}}_{R\times I}),
    \text{rank}(\underbrace{{\mathbf{U}_{:n}^{(1)}}^\top \odot {\mathbf{U}^{(3)}}^\top}_{O \times R})
    \bigg\}.
\end{align}
Therefore a single CP$\mu$MoE's $n$th expert's matrix rank is bounded by $\min\{I,O,R\}$.

\subsubsection{TR$\boldsymbol\mu$MoEs: rank analysis}
\label{sec:app:tr-rank}

We now turn our attention to TR$\mu$MoEs, where we will see that the TR ranks $R_1,R_2,R_3$ translate very favorably into matrix rank at smaller computational cost than with CP$\mu$MoEs. 
First recall that TR$\mu$MoEs are parameterized instead by core tensors
$\mathcal{U}^{(1)}\in\mathbb{R}^{R_1 \times N \times R_2}$,
$\mathcal{U}^{(2)}\in\mathbb{R}^{R_2 \times I \times R_3}$,
$\mathcal{U}^{(3)}\in\mathbb{R}^{R_3 \times O \times R_1}$, with chosen ranks $R_1,R_2,R_3$.
We can derive an expression to materialize expert $n$'s matrix through the sum of matrix products of the TR cores as:
\begin{align}
    \mathbf{W}_{n} = \sum_{{r_3}=1}^{R_3}
    \bigg(
        \underbrace{\mathbf{U}^{(3)}_{{r_3}::}}_{O\times R_1}
        \underbrace{\mathbf{U}^{(1)}_{:n:}}_{R_1\times R_2}
        \underbrace{\mathbf{U}^{(2)}_{::{r_3}}}_{R_2\times I}
    \bigg)^\top\in\mathbb{R}^{I\times O}.\label{eq:tr-matrix}
\end{align}
The matrix product rank inequality applies to each $I\times O$ matrix summand, whilst the matrix sum rank inequality applies to the outer matrix sum:
\begin{align}
    \text{rank}(\mathbf{W}_{n}) &= \text{rank}\bigg(
    \sum_{{r_3}=1}^{R_3}
    \big(
        \mathbf{U}^{(3)}_{{r_3}::}
        \mathbf{U}^{(1)}_{:n:}
        \mathbf{U}^{(2)}_{::{r_3}}
    \big)^\top
    \bigg) \\
    &\leq
    \sum_{{r_3}=1}^{R_3}
    \text{rank}\big(
    \big(
        \mathbf{U}^{(3)}_{{r_3}::}
        \mathbf{U}^{(1)}_{:n:}
        \mathbf{U}^{(2)}_{::{r_3}}
    \big)^\top
    \big) \\
    &\leq \sum_{{r_3}=1}^{R_3}
    \min \bigg\{
        \text{rank}\big( \mathbf{U}^{(3)}_{{r_3}::} \big),
        \text{rank}\big( \mathbf{U}^{(1)}_{:n:} \big),
        \text{rank}\big( \mathbf{U}^{(2)}_{::{r_3}} \big),
    \bigg\}.
\end{align}
Consequently, expert $n$'s materialized weight matrix in TR$\mu$MoEs has a more generous upper bound of $\min\big\{R_3 \cdot \min\{ R_1, R_2 \}, I, O \big\}$\footnote{Regardless of how large $R_3$ is, the rank of the matrix cannot exceed $\min\{I,O\}$.}.

Through this analysis, we observe that one can choose large values of $R_3$ yet small $R_1,R_2$ to yield a high expert matrix rank with few parameters, justifying the choice of $R_1=R_2=4$ in the main paper.

\subsubsection{Tucker$\boldsymbol\mu$MoEs: rank analysis}

One popular alternative decomposition is the Tucker decomposition \cite{Tucker1966SomeMN}. Here we derive the resulting matrix rank of this alternative $\mu$MoE variant and detail why it's not as desirable as the proposed $\mu$MoE variants.

A Tucker$\mu$MoE composes an $\mu$MoE weight tensor through the series of mode-$n$ products \cite{kolda2009tensorapplications}: $\mathcal{W}=\mathcal{Z} \times_1 \mathbf{U}^{(1)} \times_2 \mathbf{U}^{(2)} \times_3 \mathbf{U}^{(3)}$, where $\mathcal{Z}\in\mathbb{R}^{R_N\times R_I \times R_O}$ is the so-called `core tensor'
and $\mathbf{U}_1\in\mathbb{R}^{N\times R_N}, \mathbf{U}_2\in\mathbb{R}^{I\times R_I}, \mathbf{U}_3\in\mathbb{R}^{O\times R_O}$ are the `factor matrices' for the tensor's three modes.

Again following \citet{kolda2009tensorapplications} a single expert $n$’s weight matrix can be rewritten through the matricization involving the Kronecker product $\otimes$ as:
\begin{align}
    \mathbf{W}_{n} = {\mathbf{U}^{(2)}} \mathbf{Z}_{(2)} \left( {\mathbf{U}^{(1)}_{n}} \otimes {\mathbf{U}^{(3)}} \right)^\top\in\mathbb{R}^{I \times O},
\end{align}
where $\mathbf{Z}_{(2)}\in\mathbb{R}^{R_I \times (R_O \cdot R_N)}$ is the so-called mode-$2$ (matrix) unfolding of the core tensor \cite{kolda2009tensorapplications}. Consequently, the same rank inequality applies:
\begin{align}
    \text{rank}(\mathbf{W}_{n})&=
        \text{rank}\left({\mathbf{U}^{(2)}} \mathbf{Z}_{(2)} \left( {\mathbf{U}^{(1)}_{n}} \otimes {\mathbf{U}^{(3)}} \right)^\top\right) \\
        &\leq
        \min \bigg\{
            \text{rank}(\underbrace{\mathbf{U}^{(2)}}_{I\times R_I}),
            \text{rank}(\underbrace{\mathbf{Z}_{(2)}}_{\mathclap{R_I \times (R_O\cdot R_N)}}),
            \text{rank}(\underbrace{\mathbf{U}^{(1)}_n \otimes \mathbf{U}^{(3)}}_{O\times (R_O\cdot R_N)}) \bigg\},
\end{align}
Where we see the much more restrictive matrix rank upper bound applies: $\min\left\{ \min(I, R_I), \min(R_I,R_O\cdot R_N), \min(O, R_O) \right\}$. Thus in practice, \textit{both} $R_I,R_O$ need to be large to yield a large matrix rank, which is in conflict with the goal of maintaining a moderate number of parameters.

\begin{wrapfigure}[21]{r}{0.4\textwidth}
    \centering
    \includegraphics[width=0.38\textwidth]{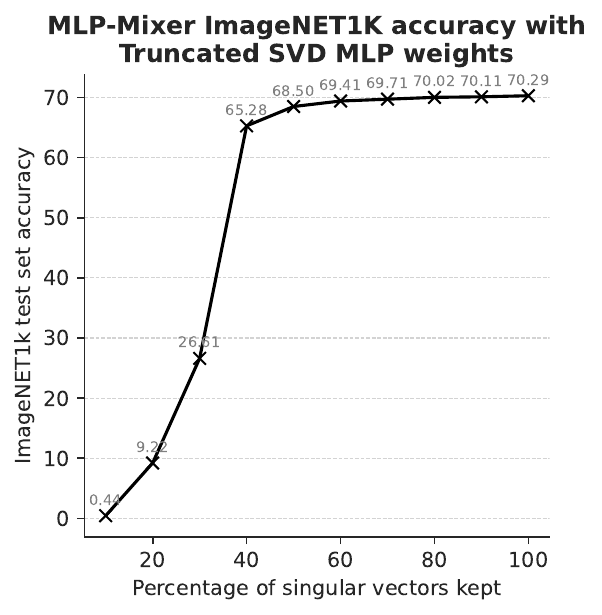}
    \caption{Val. accuracy for an \texttt{S-16} MLP-mixer when performing truncated SVD on all MLP's linear layers' weight; model accuracy is closely retained even with half the singular vectors.}
    \label{fig:mixer-truncatedsvd}
\end{wrapfigure}

\subsection{Why is low-rankness a reasonable assumption?}
Given we've seen that parameter-efficient $\mu$MoE layers lead to low-rank expert weight matrices, a natural question is whether or not low-rankness in MLP linear layers' weight matrices is a reasonable assumption or constraint.

Our strongest piece of evidence supporting the claim is experimental in nature:
we've seen from the results in \cref{sec:exp:performance} 
that using all parameter-matched $\mu$MoE layers for both MLP mixers and GPT-2 models leads to no significant drop in accuracy from their linear layer counterparts (see also \cref{sec:app:additional-performance} for many more results).

To investigate this further we perform a rank ablation on our trained MLP-Mixer model with the original linear layers' weights.
Concretely, we compute the truncated SVD of each MLP block's 2 linear layer weight matrices. We explore the impact on the model's ImageNET1k validation set accuracy when using only the top-$k$ singular vectors/values (the best rank-$k$ approximation \cite{eckart1936approximation}).
The validation set accuracy using truncated SVD weights in every mixer block is plotted in \cref{fig:mixer-truncatedsvd}--we see here that discarding as many as \textit{half} the total number of (bottom) singular vectors/values to approximate the original weights leads to negligible difference to the validation set accuracy. In other words, low-rank approximations of MLP Mixers' weights retain their representational power sufficiently well to produce nearly the same validation set accuracy as the original model.
Such findings are consistent with results in recent work in the language domain \cite{sharma2024the}, where low-rank approximations of MLP layers can even sometimes boost original performance.
The accuracy retained by MLP Mixers here even after such aggressive rank reduction constitutes further evidence that full-rank weights are not always necessary.

\subsection{MoE/$\boldsymbol\mu$MoE parameter count comparisons}
\begin{figure}
    \centering
    \includegraphics[width=0.5\textwidth]{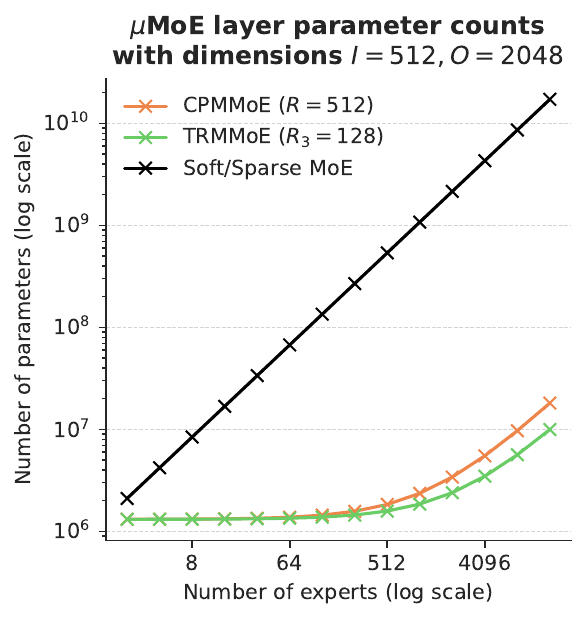}
    \caption{$\mu$MoE layer parameter count as a function of expert count.}
    \label{fig:param-ablation}
\end{figure}

We plot in \cref{fig:param-ablation} the parameter counts for $\mu$MoE layers as a function of the expert counts (sweeping from $N=2$ experts through to $N=16,384$), relative to dense/sparse MoEs (with rank $R_1=R_2=4$ TR$\mu$MoEs), for the first layer in a MLP-mixer channel-mixing block \cite{tolstikhin2021mlp}. As can be seen, both $\mu$MoE variants are vastly more parameter-efficient than dense/sparse MoEs.

Given TR$\mu$MoEs offer even better parameter efficiency for larger numbers of experts, we suggest opting for CP$\mu$MoEs when using expert counts less than $\sim128$, and considering TR$\mu$MoEs for higher values.

\paragraph{Latency and memory usage} comparisons between the $\mu$MoE, linear layers, and alternative MoEs are shown in \cref{tab:layer_comparison}, where the $\mu$MoEs perform favorably.

\begin{table}[]
    \centering
    \caption{Comparison of different layers' peak memory usage and latency (per single input). We use 128 experts in each MoE layer, and set the rank of the $\mu$MoEs to parameter-match that of the linear layer.}
    \resizebox{0.75\linewidth}{!}{ %
    \begin{tabular}{lcc}
        \toprule
        Layer type & Peak memory usage (MB) & Latency per single input (ms) \\
        \midrule
        Linear layer & 12.07 & 0.01 \\
        Dense MoE ($N=128$) & 390.17 & 1.17 \\
        Sparse MoE ($N=128$) & 765.19 & 0.80 \\
        TR$\mu$MoE ($N=128$) & 15.87 & 0.94 \\
        CP$\mu$MoE ($N=128$) & 14.02 & 1.05 \\
        \bottomrule
    \end{tabular}
    }
    \label{tab:layer_comparison}
\end{table}

\section{Hierarchical $\mu$MoE model derivations}
\label{sec:app:full-model}

\begin{figure}[]
    \centering
    \includegraphics[width=1.0\linewidth]{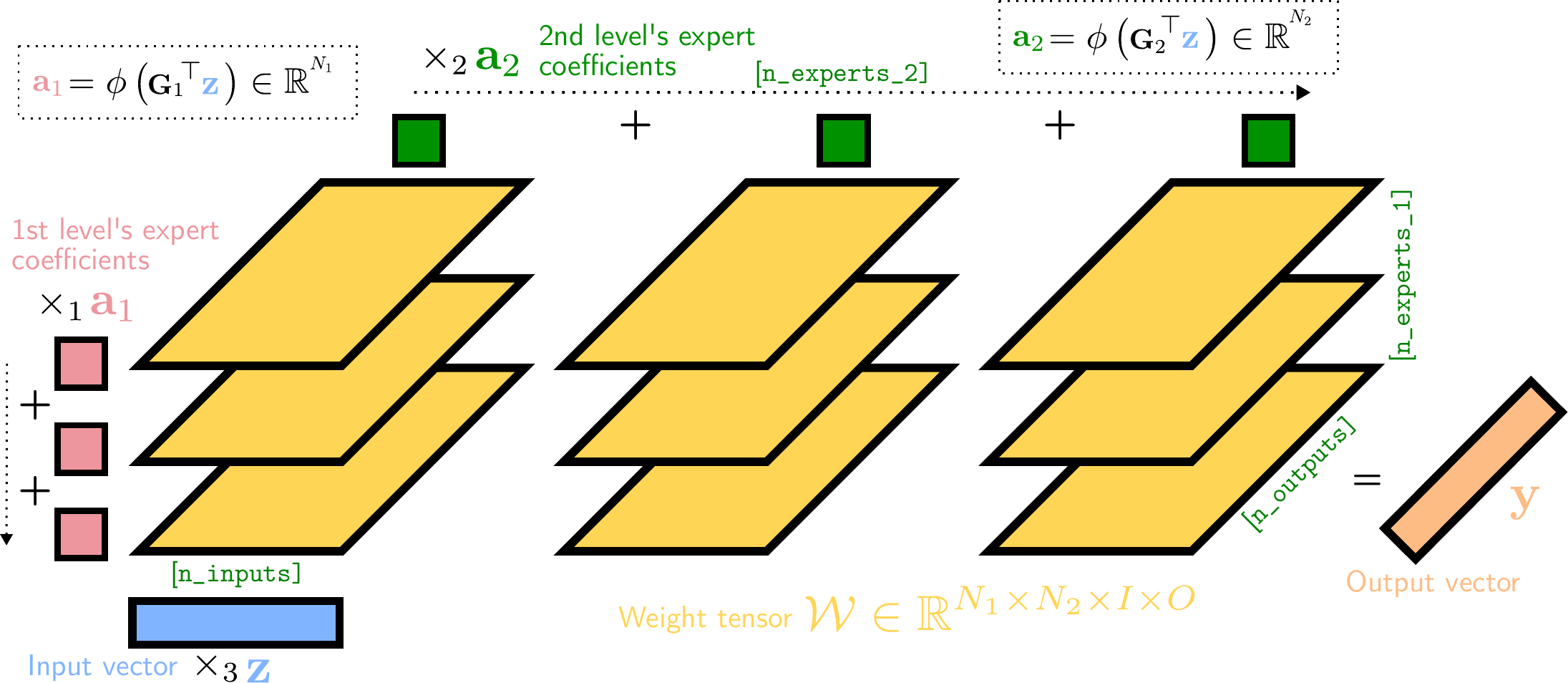}
    \caption{Illustration of a \textbf{two-hierarchy} $\mu$MoE layer's (unfactorized) forward pass as a series of tensor contractions. The $N_1\cdot N_2$ many experts' weight matrices are visualized as $2$D horizontal slices in yellow, which are (1) matrix-multiplied with the input vector, (2) summed over the first expert mode (weighted by the first expert coefficients $\mathbf{a}_1$ in red), and (3) summed over the second expert mode (weighted by the second expert mode's coefficients $\mathbf{a}_2$ in dark green).}
    \label{fig:hierarchy-fig}
\end{figure}

In the main paper, the fast forward passes are derived for a single level of expert hierarchy.
One additional attractive property of $\mu$MoEs is their straightforward extension to multiple levels of expert hierarchy--one simply increments the number of modes of the weight tensor and includes another tensor contraction with new expert coefficients.
Hierarchical $\mu$MoEs intuitively implement ``and'' operators in expert selection at each level, and further provide a mechanism through which to increase the total expert count at a small parameter cost.
Here, we derive the fast forward passes for $\mu$MoE layers in their most general form with $E$ levels of expert hierarchy.
For intuition, we first further visualize $\mu$MoE layers with 2 levels of hierarchy in \cref{fig:hierarchy-fig}--note how we have an extra mode to the weight tensor, and an extra contraction over the new expert mode to combine its outputs.

Given that hierarchical $\mu$MoEs involve very high-order tensors, we adopt the popular mode-$n$ product \cite{kolda2009tensorapplications} to express the forward passes in as readable a way as possible.
The \textbf{mode-$n$ (vector) product} of a tensor $\mathcal{X}\in\mathbb{R}^{I_1\times I_2 \times\ldots \times I_N}$ and vector $\mathbf{u}\in\mathbb{R}^{I_n}$ is denoted by $\mathcal{X}\times_n\mathbf{u}$ \cite{kolda2009tensorapplications}, with its elements given by:
$$(\mathcal{X}\times_n\mathbf{u})_{i_1\ldots i_{n-1} i_{n+1}\ldots i_N}= \sum_{i_n=1}^{I_n}x_{i_1i_2\ldots i_N} u_{i_n}.$$
We first introduce the formulation of an $E$-level hierarchical $\mu$MoE layer from \cref{eq:mmoe-moden} in the main paper: given input $\mathbf{z}\in\mathbb{R}^I$, the most general form of $\mu$MoE layer is parameterized by weight tensor $\mathcal{W}\in\mathbb{R}^{N_1 \times \ldots \times N_E \times I \times O}$ and $E$ many expert gating parameters $\{\mathbf{G}_e\in\mathbb{R}^{I\times N_e}\}_{e=1}^E$.
The explicit, unfactorized forward pass is given by:
\begin{align}
    \mathbf{a}_e&=\phi(\mathbf{G}_e^\top\mathbf{z})\in\mathbb{R}^{N_e}, \quad \forall e \in \{1, \ldots, E\} \nonumber, \\
    \mathbf{y} &= \mathcal{W} \times_{1} \mathbf{a}_1 \times_2 \ldots \times_{E} \mathbf{a}_E \times_{E+1} \mathbf{z}\nonumber \\
        &= \sum_{n_1=1}^{N_1}{a_{1}}_{n_1}\ldots\sum_{n_E=1}^{N_E}{a_E}_{N_E} \big(\underbrace{\mathbf{W}_{n_1\ldots n_E::}^\top}_{O\times I}\mathbf{z}\big)
    \in\mathbb{R}^O\label{eq:hierarchy-sum},
\end{align}
where \cref{eq:hierarchy-sum} is expressed as sums over the $E$-many expert modes to make it clear that hierarchical $\mu$MoEs take convex combinations of $\prod_{e=1}^E N_e$ many experts' outputs (given there are $N_e$ experts at each level of hierarchy).
With expert coefficients $\{\mathbf{a}_e\in\mathbb{R}^{N_e}\}_{e=1}^E$, the factorized forward passes of the most general hierarchical $\mu$MoE layers are given for the two variants below.

\subsection{Hierarchical CP$\boldsymbol\mu$MoE}
The full CP$\mu$MoE model of rank $R$ has an implicit weight tensor
$\mathcal{W}=\sum_{r=1}^R \mathbf{u}^{(1)}_{r} \circ \mathbf{u}^{(2)}_{r} \circ \mathbf{u}^{(3)}_{r} \circ \cdots \circ \mathbf{u}^{(E+2)}_{r} \in \mathbb{R}^{N_1 \times \cdots \times N_E \times I \times O}$, with factor matrices $\mathbf{U}^{(1)}\in\mathbb{R}^{R \times N_1},\ldots, \mathbf{U}^{(E)}\in\mathbb{R}^{R \times N_E}, \mathbf{U}^{(E+1)}\in\mathbb{R}^{R \times I}, \mathbf{U}^{(E+2)}\in\mathbb{R}^{R \times O}$.
The implicit, factorized forward pass is given by:
\begin{align}
    \mathbf{y}
        &= \left(
        \sum_{r=1}^R \mathbf{u}^{(1)}_{r} \circ \mathbf{u}^{(2)}_{r} \circ \mathbf{u}^{(3)}_{r} \circ \cdots \circ \mathbf{u}^{(E+2)}_{r} \right)
        \times_1 \mathbf{a}_1
        \times_2 \ldots \times_{E} \mathbf{a}_{E}
        \times_{E+1} \mathbf{z}
        \nonumber  \\
        &= \sum_{r=1}^R \mathbf{u}^{(E+2)}_{r} \big(\sum_{{n_1},\ldots,{n_E},i} u_{rn_1}^{(1)} a_{1_{n_1}} \cdots u_{r{n_E}}^{(E)} a_{E_{n_E}} u_{ri}^{(E+1)}z_{i}  \big)\nonumber  \\
        &= \sum_{r=1}^R
            \mathbf{u}^{(E+2)}_{r}
            \big({\mathbf{U}^{(1)}}\mathbf{a}_1\big)_r \cdots
            \big({\mathbf{U}^{(E)}}\mathbf{a}_E\big)_r \cdot
            \big({\mathbf{U}^{(E+1)}}\mathbf{z}\big)_r
            \in\mathbb{R}^O.\label{eq:full-fast-cp}
\end{align}

\subsection{Hierarchical TR$\boldsymbol\mu$MoE}

In TR format, $\mathcal{W}\in\mathbb{R}^{N_1 \times \cdots \times N_E \times I \times O}$ has $E+2$ factor tensors:
$\mathcal{U}^{(1)}\in\mathbb{R}^{R_1 \times N_1 \times R_2}, \ldots, \mathcal{U}^{(E)}\in\mathbb{R}^{R_{E} \times N_E \times R_{E+1}}$,
$\mathcal{U}^{(E+1)}\in\mathbb{R}^{R_{E+1} \times I \times R_{E+2}}$,
$\mathcal{U}^{(E+2)}\in\mathbb{R}^{R_{E+2} \times O \times R_1}$,
where $R_i$ are the manually chosen ranks.
The weight tensor's elements are given by:
\begin{align*}
    w_{{n_1}\ldots{n_E}io} = \text{tr}\big( {\mathbf{U}^{(1)}_{:{n_1}:}}\cdots {\mathbf{U}^{(E)}_{:{n_E}:}} {\mathbf{U}^{(E+1)}_{:{i}:}} {\mathbf{U}^{(E+2)}_{:{o}:}} \big).
\end{align*}
We derive the fast factorized forward pass in terms of a series of mode-$2$ products:
\begin{align}
    \mathbf{y}
        &= \sum_{i} \sum_{{n_1},\ldots {n_E}} \mathcal{W}(n_1,\cdots,n_E,i,:) {\mathbf{a}_{1}(n_1)} \cdots {\mathbf{a}_{E}(n_E)} \mathbf{z}(i)  \\
        &= \sum_{r_1,r_{E+2}} \mathbf{u}^{(E+2)}_{r_{E+2}:r_1}  \big(\underbrace{(\mathcal{U}^{(1)}\times_2\mathbf{a}_1) \cdots (\mathcal{U}^{(E)}\times_2\mathbf{a}_E ) (\mathcal{U}^{(E+1)}\times_2\mathbf{z} ) }_{R_1\times R_{E+2}}\big)_{r_{1}r_{E+2}}
        \in\mathbb{R}^{O}.\label{eq:full-fast-TR}
\end{align}

\section{Experimental details}
\label{sec:app:config}

\subsection{Network configurations and hyperparamters}

Here we provide the full experimental details and setups to reproduce the performance results in the paper for each of the networks. We further include the per-epoch accuracy plots for additional transparency into the training processes.

\begin{figure}
    \centering
    \includegraphics[width=\linewidth]{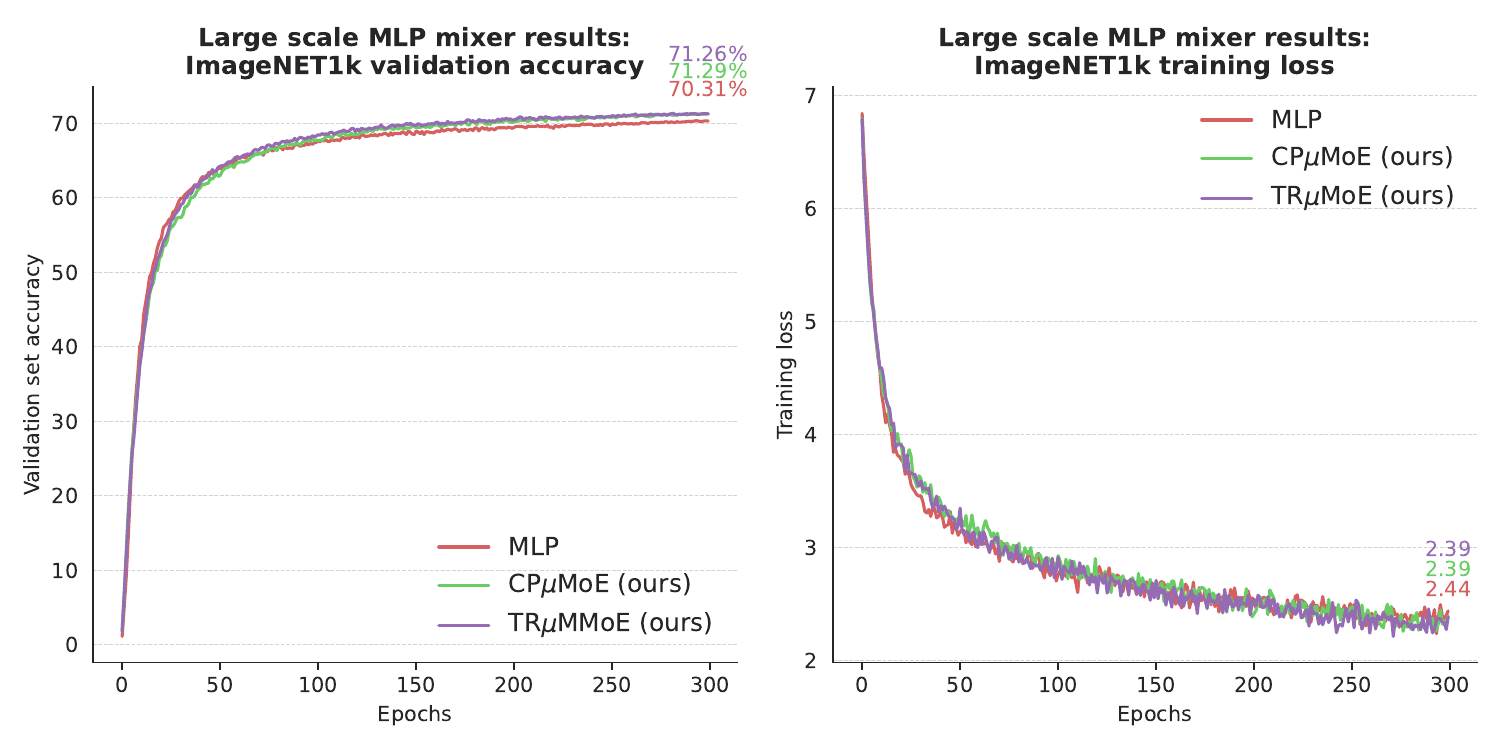}
    \caption{Training loss and validation accuracy for the MLP-mixers models for 300 epochs.}
    \label{fig:mixer-curves}
\end{figure}

\begin{figure}
    \centering
    \includegraphics[width=\linewidth]{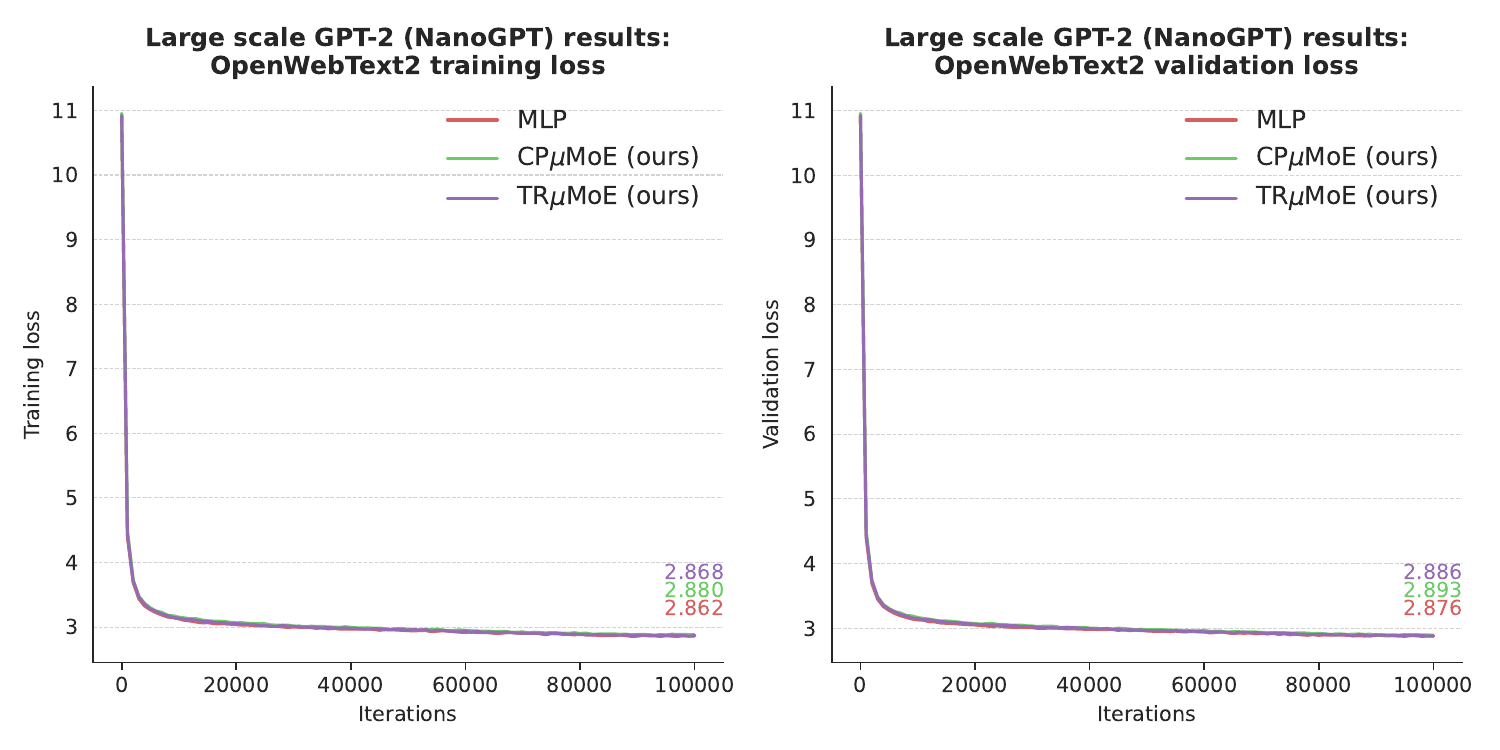}
    \caption{Training and validation loss for the GPT-2 models for 100k iterations.}
    \label{fig:gpt-curves}
\end{figure}

The experimental configurations used to reproduce the performance results in the main paper follow as closely as possible those specified in the main paper of MLP-mixer \cite{tolstikhin2021mlp} and open-source code (\url{https://github.com/lucidrains/mlp-mixer-pytorch}), the open-source code for NanoGPT (\url{https://github.com/karpathy/nanoGPT}) for GPT2 \cite{radford2019language}, and the robust fine-tuning protocol of \cite{wortsman2022robust} for CLIP \cite{radford2021learning}.
These values are summarized in \cref{tab:experimental-config}.
We plot the learning curves for the training of both models in \cref{fig:gpt-curves,fig:mixer-curves}.
\begin{table}[h]
\centering
\caption{Experimental configuration and settings for the results reported in the main paper in \cref{sec:exp:performance}.}
\label{tab:experimental-config}
\resizebox{\columnwidth}{!}{%
\begin{tabular}{@{}lcccccccccccc@{}}
\toprule
 & Learning & Batch & Weight & Warmup & Training & Stochastic & RandAugment & & Mixup & Mixed & Random & \\
 & rate & size & decay & steps & duration & depth & strength & Dropout & strength & precision & seed & Hardware \\ \midrule
MLP Mixer & 1e-3 & 4096 & 1e-4 & 10k & 300 epochs & True & 15 & 0 & 0.5 & bf16 & 0 & 4xA100 80GB \\
NanoGPT & 6e-4 & 24 & 1e-1 & 2k & 100k iter. & False & 0 & 0 & 0 & fp16 & 0 & 4xA100 80GB \\
CLIP & 3e-5 & 4096 & 1e-1 & 500 & 10 epochs & False & 0 & 0 & 0 & fp16 & 0 & 1xA100 80GB \\ \bottomrule
\end{tabular}%
}
\end{table}

\paragraph{Rank choices}
Throughout all experiments in the main paper, we fix the TR$\mu$MoE ranks for the first two modes to be $R_1=R_2=4$. This way, we can maximize the effective expert matrix ranks at a low parameter cost, as shown in \cref{sec:app:tr-rank}. The final TR rank $R_3$ is varied to parameter-match the networks in question. For CP$\mu$MoEs, we set the single CP rank $R$ to parameter-match the baselines.

\paragraph{Training times}
Each MLP mixer model takes just under 3 days to train on 4xA100 80GB GPUs. The NanoGPT models take 2-3 days to train for $100k$ iterations, with the same resources.


\subsection{Weight initialization}

We initialize each element of the factor matrices/tensors for the input and output modes from a $U[-\sqrt{k},\sqrt{k}]$ distribution (following PyTorch's linear layers' initialization strategy), for $k=1/\mathrm{in\_features}$, where $\mathrm{in\_features}$ is the dimension of the input to each factor matrix/tensor during the factorized forward passes.

Factor matrices for the expert modes are initialized to replicate the weight matrices along the expert mode (plus optional noise).
For CP$\mu$MoEs, this corresponds to sampling the factor matrices' elements from a $\mathcal{N}(1,\sigma)$ distribution.
For TR$\mu$MoEs, the weight matrices can instead be replicated along the expert mode by initializing each slice (e.g. $\mathcal{G}_1(:,i,:)$) as a diagonal matrix with its elements sampled from $\mathcal{N}(1,\sigma)$.
In all our experiments we set $\sigma:=1$ to introduce noise along the first expert mode, and $\sigma:=0$ for additional expert modes.

\section{Expert specialism: additional results}

\subsection{Large scale models}

\begin{figure}[]
    \centering
    \begin{subfigure}[b]{\textwidth}
        \includegraphics[width=\linewidth]{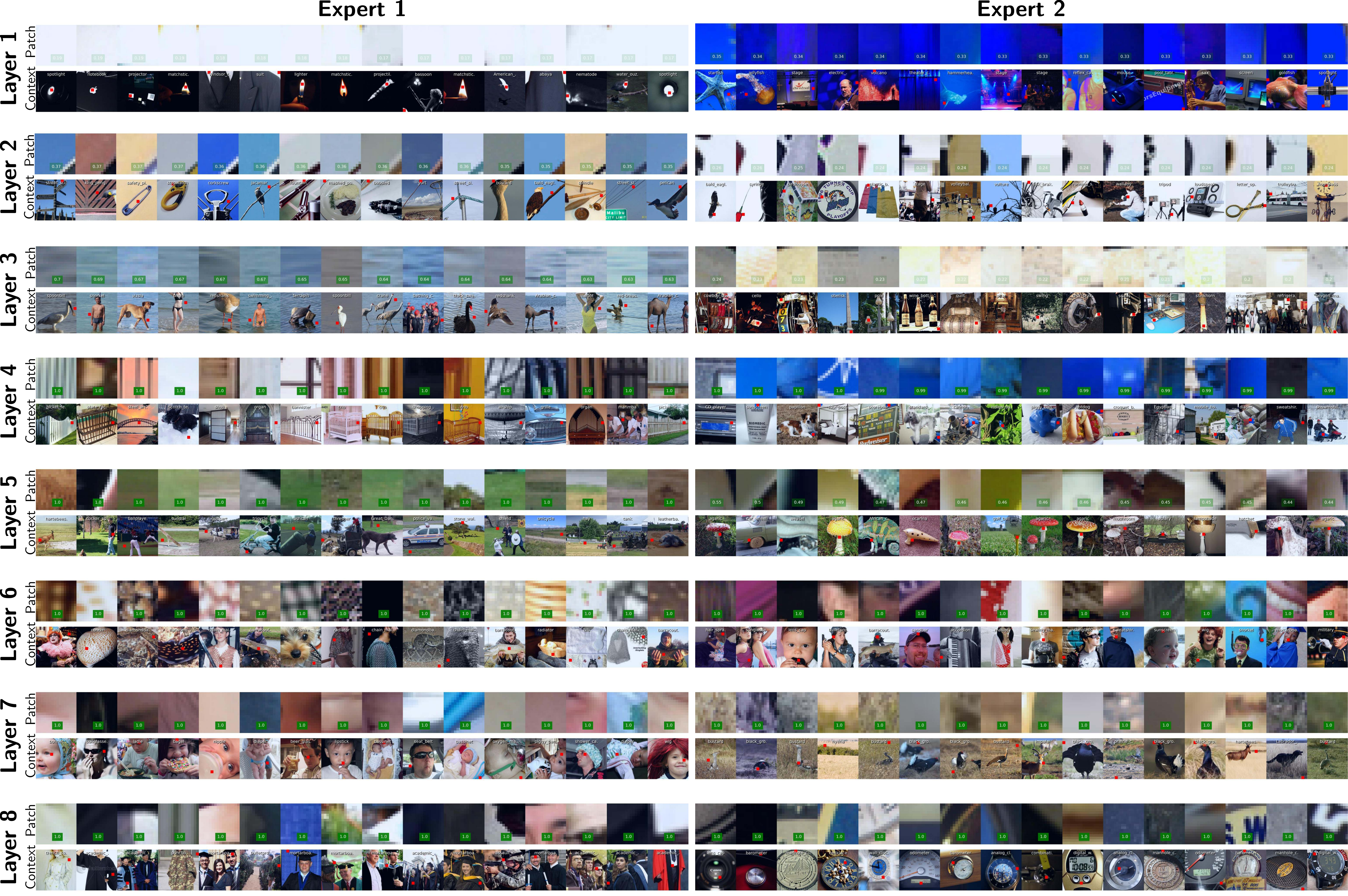}
        \caption{\textbf{CP$\mu$MoE block MLP-Mixers}: top-activating tokens.}
        \label{fig:mlp-mixer-cp-full}
    \end{subfigure}
    \begin{subfigure}[b]{\textwidth}
        \includegraphics[width=\linewidth]{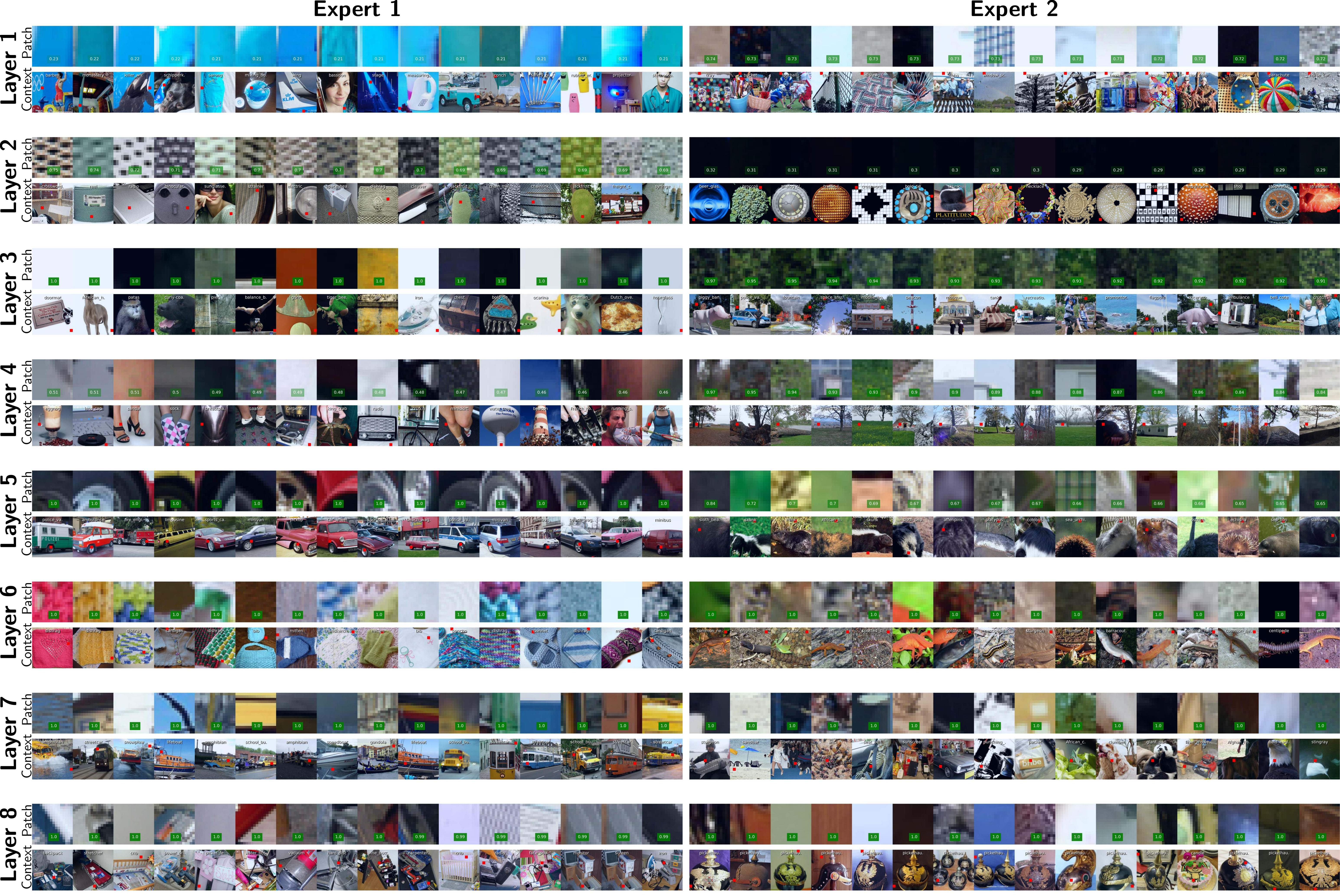}
        \caption{\textbf{TR$\mu$MoE block MLP-mixers}: top-activating tokens.}
        \label{fig:mlp-mixer-tr-full}
    \end{subfigure}
    \caption{Top-activating patches (and their surrounding image context) for the first experts at two blocks in MLP-mixer models. $\mu$MoE blocks (with $N=64$) exhibit coarse-grained specialism (e.g., texture) earlier and more fine-grained specialism (e.g., object category) deeper in the network.}
    \label{fig:mlp-mixer-subfigures}
\end{figure}

We first show in \cref{fig:mlp-mixer-subfigures} the top-activating examples for MLP-mixers trained with both CP$\mu$MoE and TR$\mu$MoE blocks. Examples are shown for the first two experts as they appear numerically for each of the $8$ layers, where we observe the same phenomenon of earlier blocks specializing to textures, and later blocks to higher-level abstract concepts/objects.

\begin{figure}
    \centering
    \includegraphics[width=\linewidth]{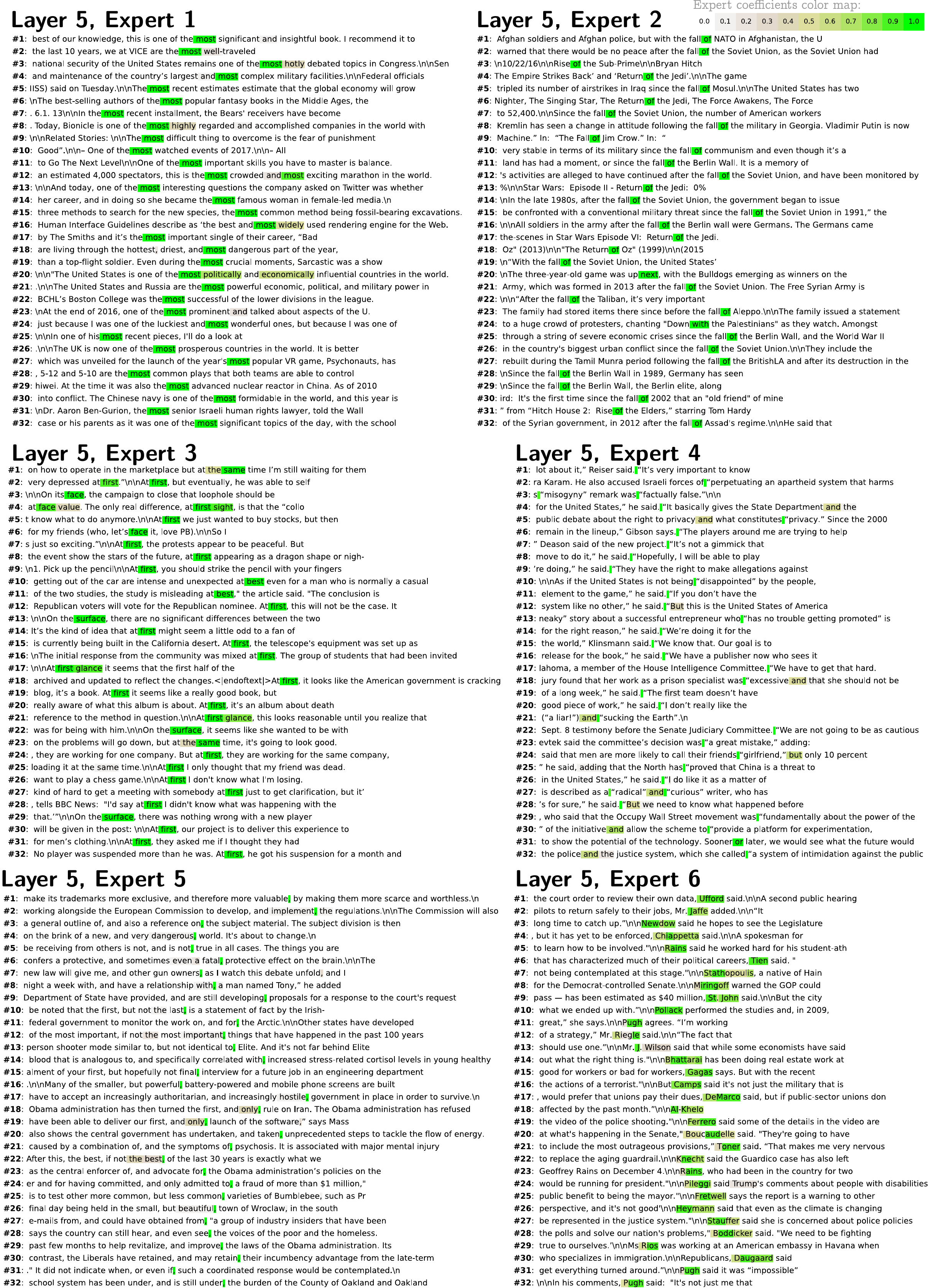}
    \caption{Top-activating $32$ tokens for the first unfiltered experts 1-6 (as ordered numerically) at layer 5 in the CP$\mu$MoE GPT2 model \textbf{(Please find the next 6 experts in \cref{fig:gpt-layer5-specialism2})}.}
    \label{fig:gpt-layer5-specialism}
\end{figure}

Secondly, in \cref{fig:gpt-layer5-specialism} we show the top $32$ activating tokens for the first $6$ experts (as they appear numerically) for layer $5$ in GPT2 models trained with CP$\mu$MoEs replacing every MLP block.
Whilst there are clear coherent themes amongst the top-activating tokens, we do see some examples of multiple themes being processed with high coefficients by the same experts (e.g. example \#20 in expert 2's top-activating examples appears unrelated to the context of the other top-activating tokens) indicating a certain degree of expert polysemanticity (as expected in the large open domain of web text).

\begin{figure}
    \centering
    \includegraphics[width=\linewidth]{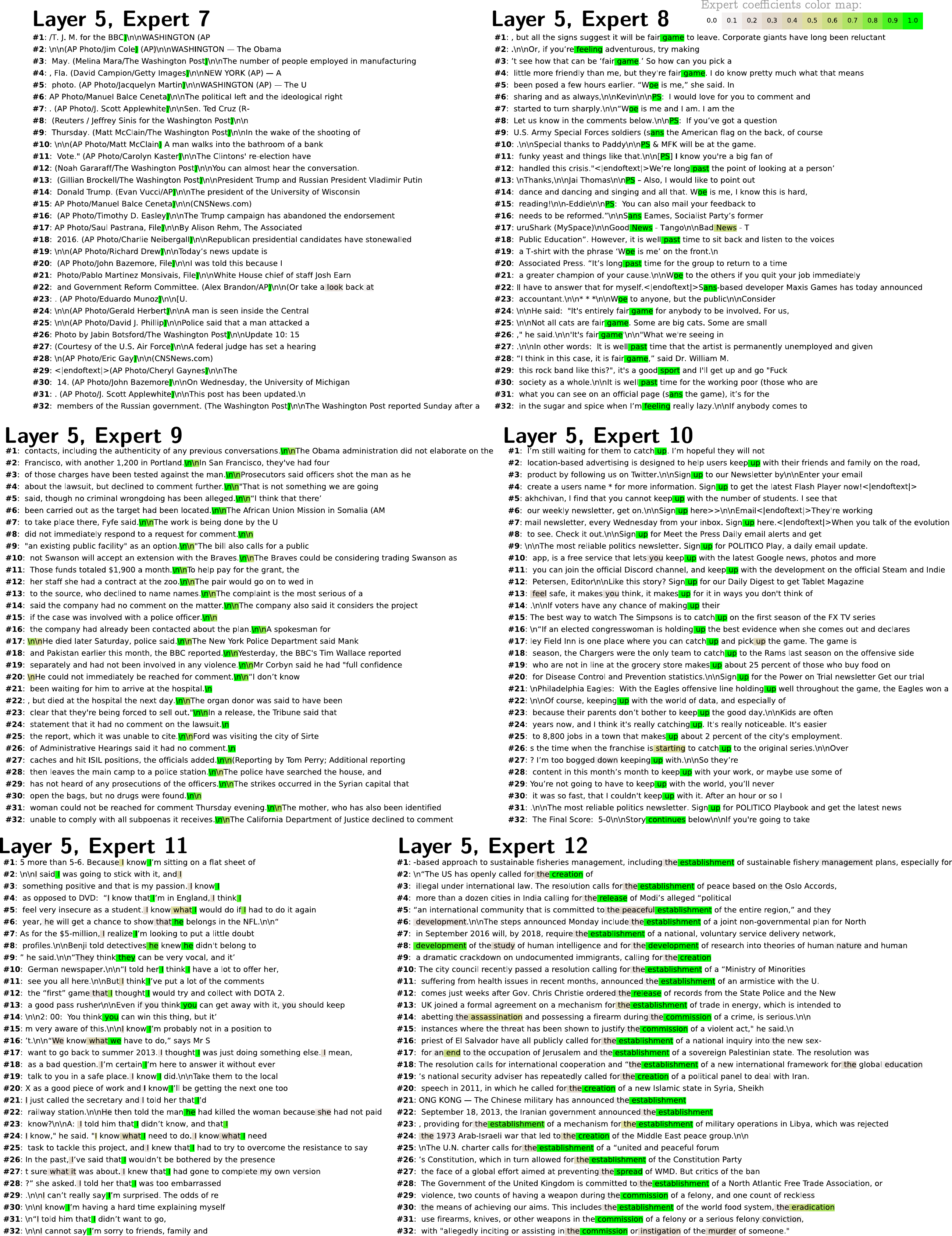}
    \caption{Top-activating $32$ tokens for the unfiltered experts 7-12 (as ordered numerically) at layer 5 in the CP$\mu$MoE GPT2 model.}
    \label{fig:gpt-layer5-specialism2}
\end{figure}

\newpage

\subsection{LLM steering}

Here we provide additional evidence that the experts' specialization is mechanistically relevant to the functionality of the network, in the sense that we use them to steer the LLM's output.

\begin{figure}
    \centering
    \includegraphics[width=1.0\textwidth]{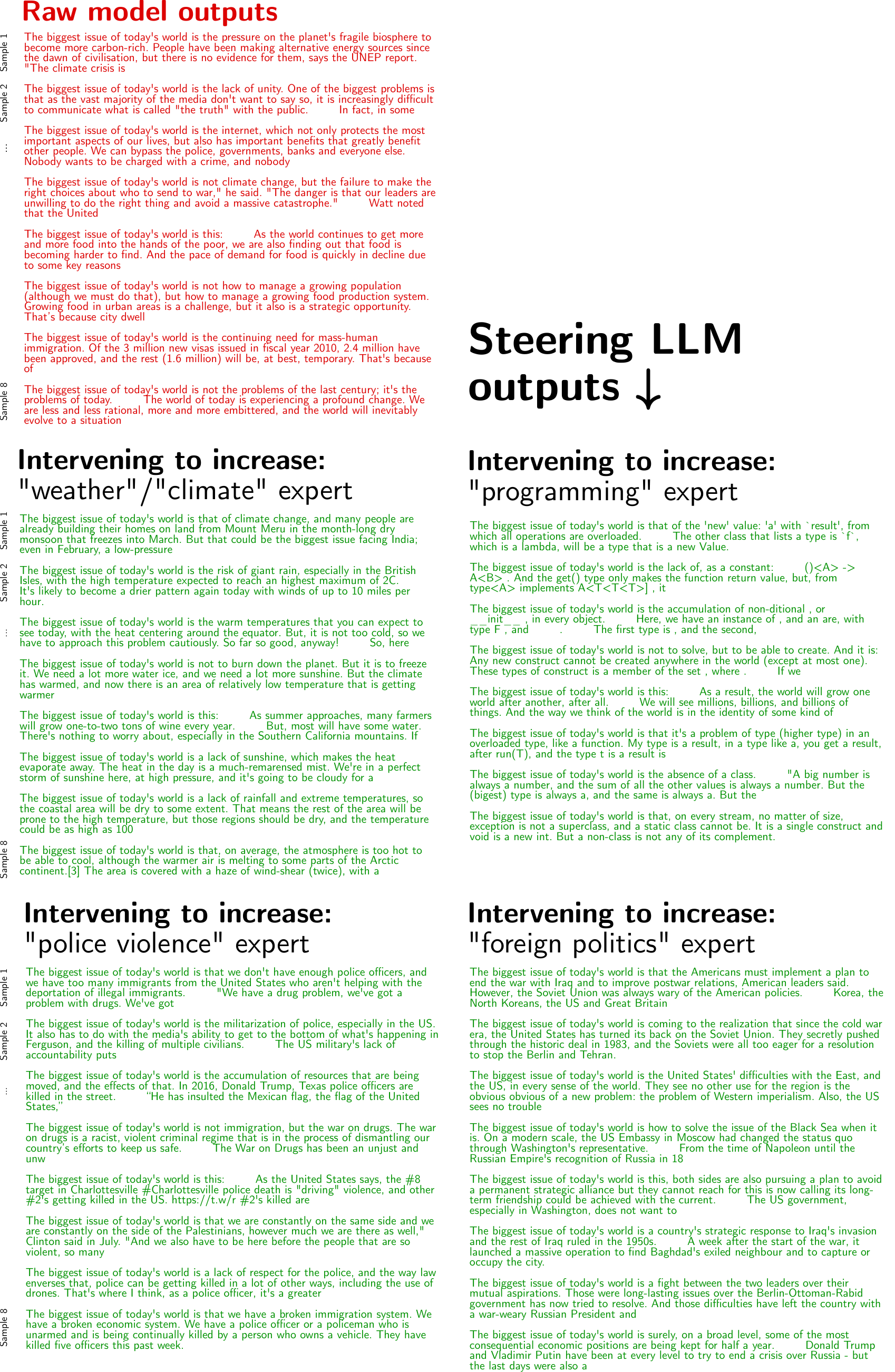}
    \caption{\textbf{Steering LLM outputs by forcefully activating experts:} adding specific manually chosen expert's cluster centers to GPT-2's activation vectors at particular layers reliably steer the LLM generations towards specific themes, based on the learned expert specialism. For example, we see an expert that steers discussion towards police violence, or about the climate. The initial prompt in every instance is the text: \texttt{``The biggest issue of today's world is''}.}
    \label{fig:steer}
\end{figure}

In particular, we use a larger GPT-2 model trained from scratch with $\mu$MoE layers at each MLP layer, \textbf{using 2048 experts at every layer}, following the setup in \cref{sec:large-models}. By modifying the forward pass of the trained model—specifically, adding selected expert cluster center vectors to each token's input latent activation vector before applying the $\mu$MoE layer—we can consistently control the model to generate outputs aligned with specific themes. Illustrations of this approach, using 4 different manually chosen experts (with their first 8 generated samples) are shown in \cref{fig:steer}. The selected experts guide the language model's outputs toward discussing topics such as climate change, police brutality, or foreign politics. We suggest that these findings further demonstrate the effectiveness of the $\mu$MoE layer in facilitating controllable generation of language model outputs.

However, we note that these initial results are hand-selected examples of some of the experts which do exhibit sensible specialization. We find many experts, when activated, do not steer the generations in such an interpretable high-level manner.

\subsection{CLIP ViT-B-32}

\begin{figure*}[]
    \centering
    \includegraphics[width=\linewidth]{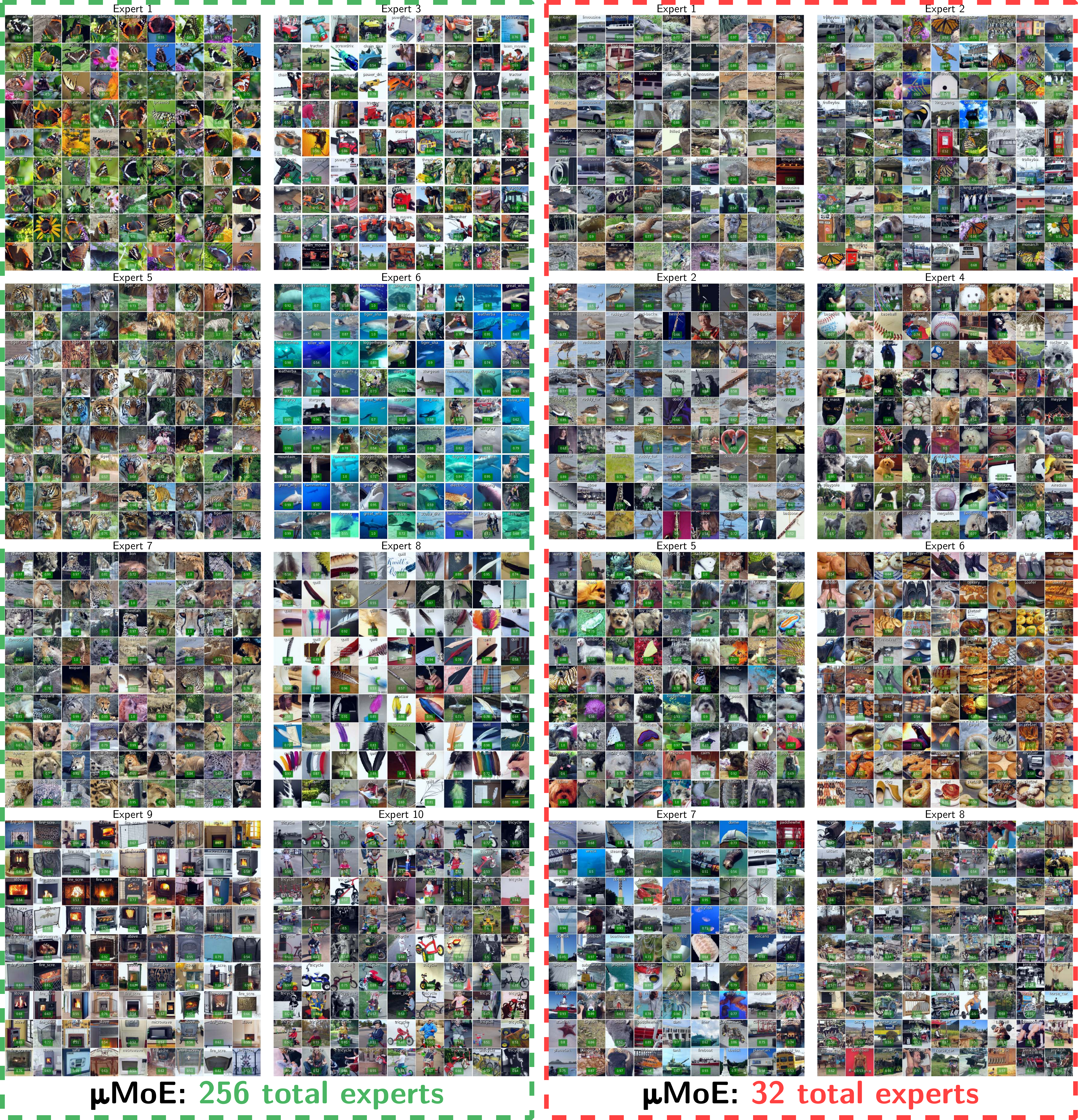}
    \caption{\textbf{\textcolor{ForestGreen}{High} vs \textcolor{red}{low} total expert count}: \textit{Randomly} selected training set images with expert coefficient $\geq0.5$ for the first $10$ numerical experts (of those processing any images with coefficient $\geq0.5$). Results are with \texttt{CP-r512} $\mu$MoE layers with \textcolor{ForestGreen}{256 (left)} and \textcolor{red}{32 (right)} total experts respectively. We highlight the apparent specialism of the experts when a higher total number is used. \textbf{(Please zoom for detail)}}
    \label{fig:image-grids}
\end{figure*}

\begin{figure*}[]
    \centering
    \includegraphics[width=1.0\linewidth]{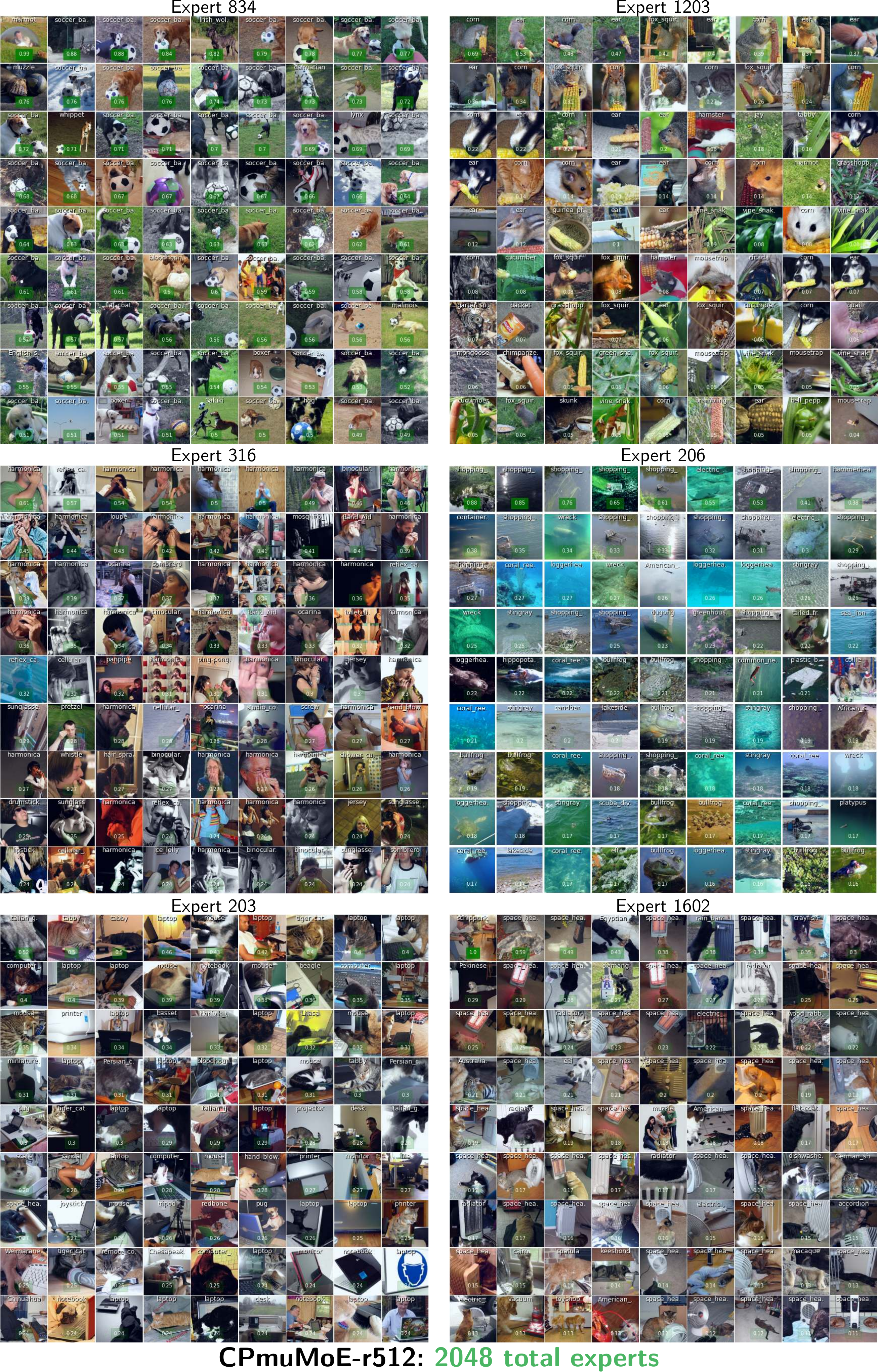}
    \caption{\textbf{Fine-grained expert specialisms}: \textit{Manually} selected experts (and images ranked by \textit{highest} expert coefficients) processing what appears to be very fine-grained categories (e.g. animals with footballs, trolleys in water, etc.). Model fine-tuned on ImageNET1k with a high number of $2048$ experts and a \texttt{CP-r512} $\mu$MoE final CLIP layer. \textbf{(Please zoom for detail)}}
    \label{fig:image-grids-specific}
\end{figure*}

\paragraph{Qualitative visualization}
Additional results to further substantiate the claims in the main paper about expert class-modularity are presented here.
Firstly in \cref{fig:image-grids} are many more random images (of those with expert coefficient $\geq 0.5$) of the first few experts as they are ordered numerically.
Furthermore, when we use an even larger number of experts (i.e. $2048$) we observe a select few experts developing what appear to be very fine-grained specialisms, as shown in \cref{fig:image-grids-specific}. For example, images with large coefficients for \#$203$ are often animals on top of laptops, whilst images with high coefficients for \#$1203$ are animals eating corn.

\begin{figure*}
    \centering
    \includegraphics[width=1.0\linewidth]{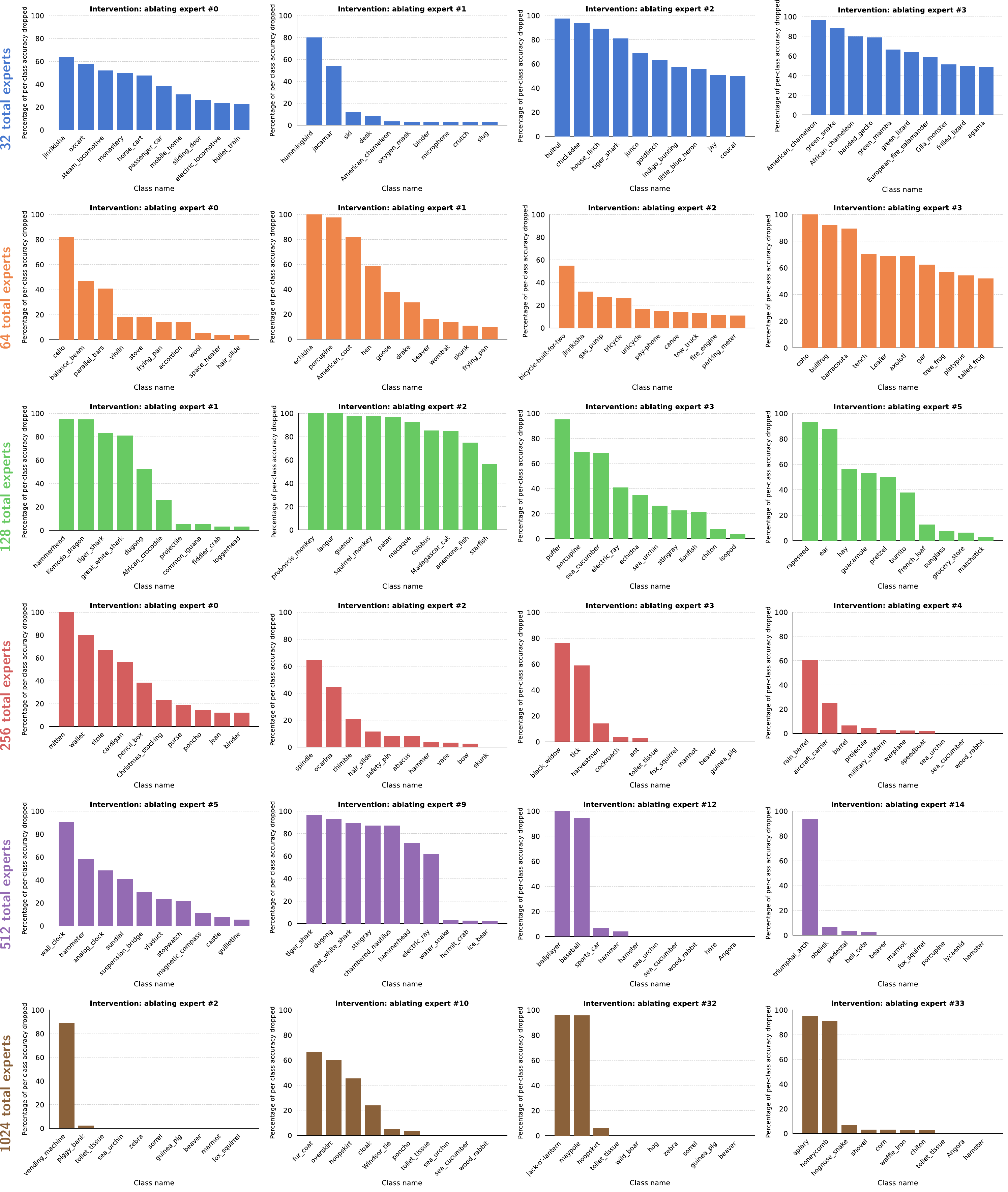}
    \caption{\textbf{Penultimate layer CP$\boldsymbol\mu$MoE}: Percentage of per-class test set accuracy lost when intervening and ablating particular experts (along the columns). In general, the more total experts (rows), the more class-level monosemantic the experts are as indicated by the mass centred on fewer classes, and with higher magnitude. Shown are the first $4$ experts in each model (row) to change $\geq0.5$ of any class' accuracy when counterfactually ablated.}
    \label{fig:sup:class-ablate-penultimate}
\end{figure*}

\begin{figure*}
    \centering
    \includegraphics[width=1.0\linewidth]{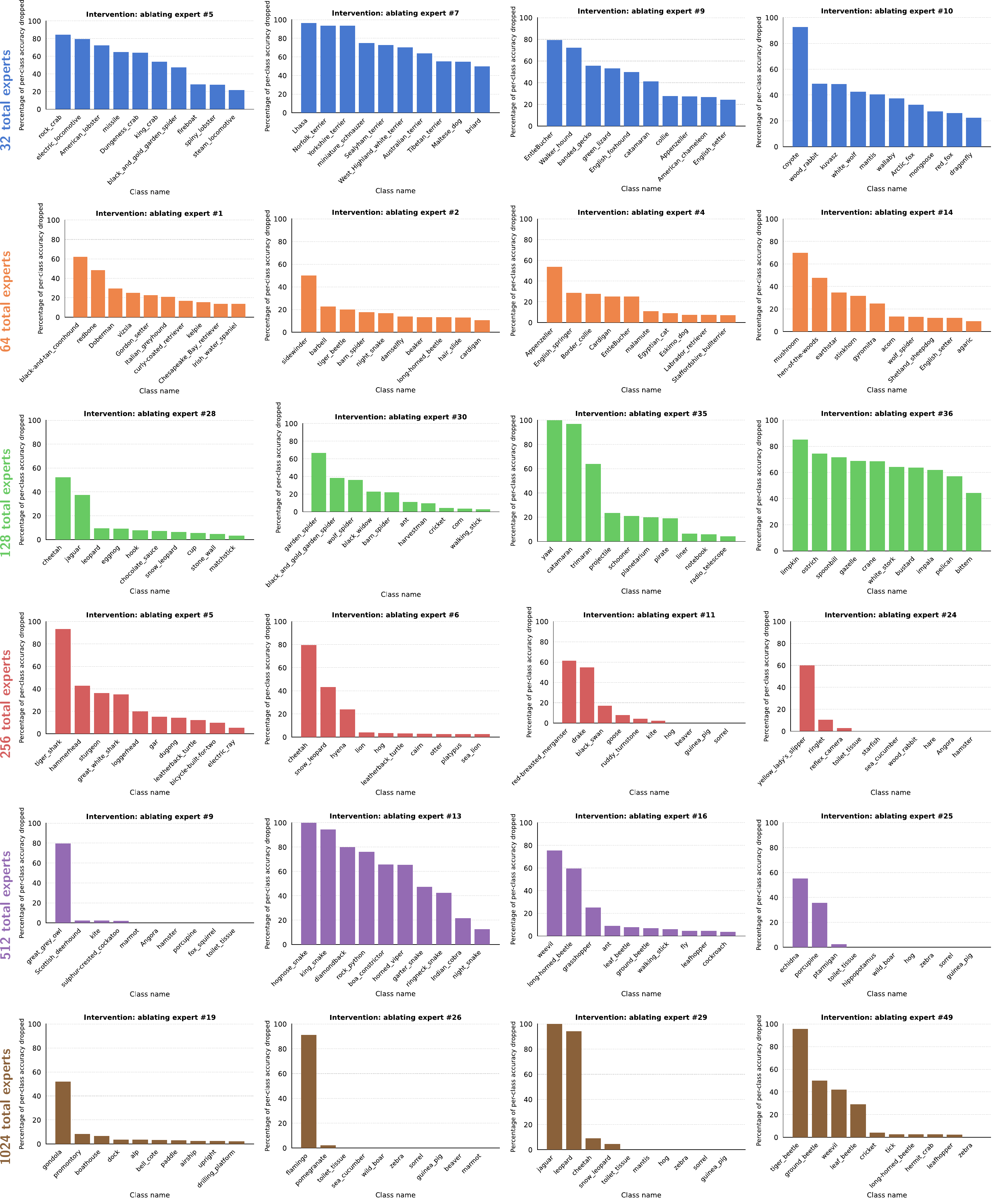}
    \caption{\textbf{Final layer CP$\boldsymbol\mu$MoE}: Percentage of per-class test set accuracy lost when intervening and ablating particular experts (along the columns). In general, the more total experts (rows), the more class-level monosemantic the experts are as indicated by the mass centred on fewer classes, and with higher magnitude. Shown are the first $4$ experts in each model (row) to change $\geq0.5$ of any class' accuracy when counterfactually ablated.}
    \label{fig:sup:class-ablate-final}
\end{figure*}

\paragraph{Counterfactual intervention barplots}
Next, we show barplots of the class labels whose test set accuracies are most changed under the counterfactual question in the main paper: ``had (expert $n$) not contributed its weight, how would the class predictions have changed?''. These are shown in \cref{fig:sup:class-ablate-penultimate} and \cref{fig:sup:class-ablate-final} when using a CP$\mu$MoE as a final and penultimate layer respectively. As can be seen, we often observe that a higher number of experts (the final rows in brown color) lead to experts that, upon ablation, cause the model to lose almost all its accuracy for fewer classes.
Experts here are chosen in numerical order and only those yielding $\geq0.5$ total accuracy change to any class upon counterfactual ablation.


\section{Ablation studies}
\label{sec:app:ablation}

\subsection{Entmax vs softmax}
\begin{figure}[H]
    \centering
    \begin{subfigure}[b]{0.5\linewidth}
        \includegraphics[width=\linewidth]{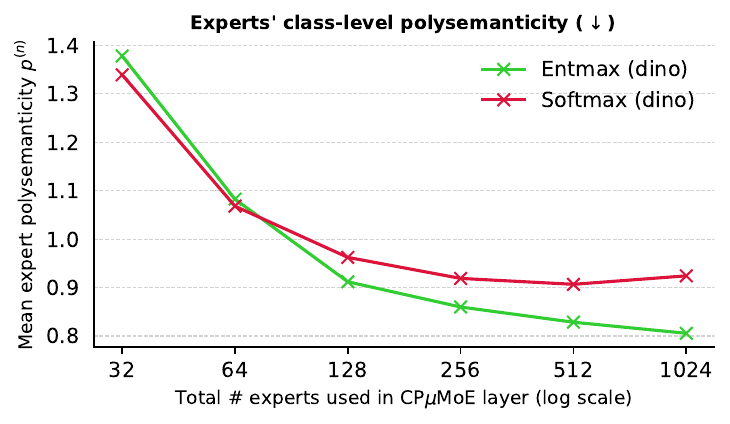}
        \label{fig:entmax-ablation-dino}
        \vspace{-2em}
        \caption{DINO backbone}
        \vspace{2em}
    \end{subfigure}%
    \begin{subfigure}[b]{0.5\linewidth}
        \includegraphics[width=\linewidth]{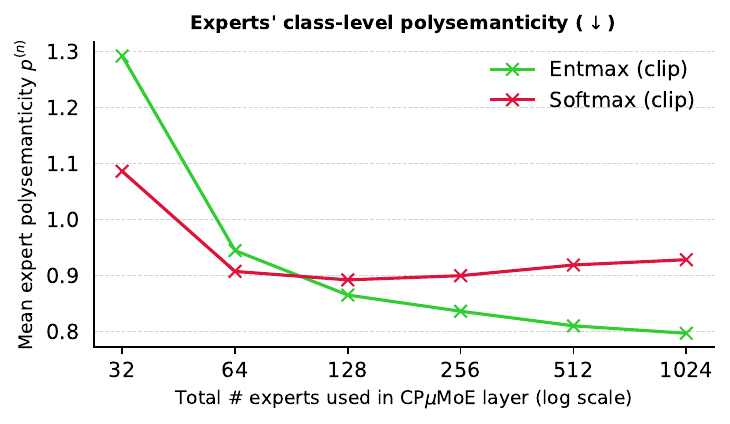}
        \label{fig:entmax-ablation-softmax}
        \vspace{-2em}
        \caption{CLIP backbone}
        \vspace{2em}
    \end{subfigure}%
    \vspace{-2em}
    \caption{\textbf{Softmax vs Entmax ablation} \texttt{CP$\mu$MoE-r512} final layers trained on ImageNET, and the resulting class-level polysemanticity. For large values of experts, the entmax activation produces more specialized experts.}
    \label{fig:entmax-ablation-polysemantic}
\end{figure}

We find the use of the entmax activation function \cite{peters2019entmax,correia2019entmax} to produce more monosemantic experts, as quantified by the measure of polysemanticity used in the main paper. We show in \cref{fig:entmax-ablation-polysemantic} the mean expert polysemanticity (of those experts that affect the class accuracy upon ablation) for \texttt{CP$\mu$MoE-r512} final layer models fine-tuned with various numbers of experts.
As can be seen, the entmax function consistently produces more monosemantic experts for larger total expert counts.
We attribute this to the sparsity in entmax's post-activation distribution (whereas the softmax function can just as readily output a uniform distribution over all expert coefficients).
\begin{wrapfigure}[9]{r}{0.5\textwidth}
\centering
\captionof{table}{Original $\mu$MoE layers' FLOPs vs the fast \texttt{einsum} forward passes in \cref{sec:app:implementations} (for $N=512$ experts with $768$-dimensional input and output dimensions).}
\label{tab:fast-flops}
\resizebox{0.9\linewidth}{!}{%
\begin{tabular}{@{}lll@{}}
\toprule
 & \textbf{CP$\boldsymbol\mu$MoE} & \textbf{TR$\boldsymbol\mu$MoE} \\ \midrule
Original FLOPs & 155.1B & 622.8B \\
\textbf{Fast model FLOPs} & \textbf{1.4M} & \textbf{3.5M} \\ \bottomrule
\end{tabular}%
}
\end{wrapfigure}

\subsection{Fast forward pass computation speedups}
We next report in \cref{tab:fast-flops} the actual number of FLOPs (as reported by \url{https://detectron2.readthedocs.io/en/latest/_modules/fvcore/nn/flop_count.html}) when executing PyTorch $\mu$MoE layers using the naive forward pass relative to the cost when using the fast \texttt{einsum} computation derived in \cref{sec:app:implementations}--the fast computation is many orders of magnitude less expensive (using one A100 GPU).

\subsection{Batch normalization}
\label{app:sec:bn-ablation}
\begin{wrapfigure}[12]{r}{0.5\textwidth}
    \centering
    \includegraphics[width=0.48\textwidth]{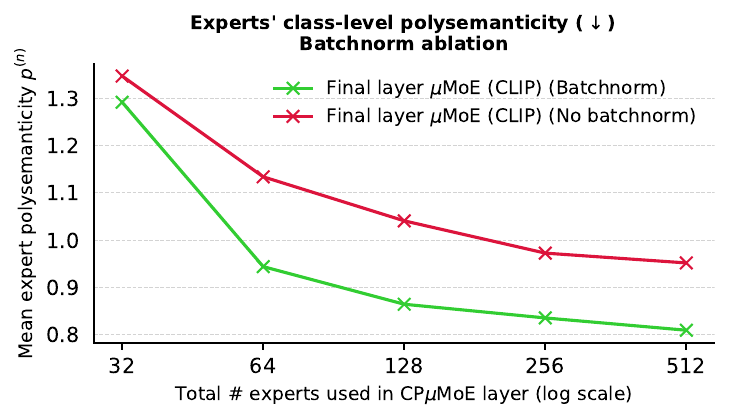}
    \caption{Ablation study: batch normalization leads to more class-level monosemantic experts.}
    \label{fig:bn-ablation}
\end{wrapfigure}

We next perform an ablation study for the use of batch normalization (BN) before the activation function for the expert coefficients.
We study CP$\mu$MoE final layer layers with CLIP ViT-B-32, quantifying BN's effect on expert class-monosemanticity as a function of the expert count.
Concretely, we perform the same class-level polysemanticity experiments as in the main paper, with and without batch normalization in \cref{fig:bn-ablation}. As can be seen clearly, the batch normalization models lead to individual experts that are increasingly class-monosemantic as desired (as a function of the total expert count).

\clearpage
\newpage
\subsection{Expert load}
\label{sec:app:expert-load}
\begin{figure*}[]
    \centering
    \begin{subfigure}[b]{1.00\linewidth}
        \includegraphics[width=\linewidth]{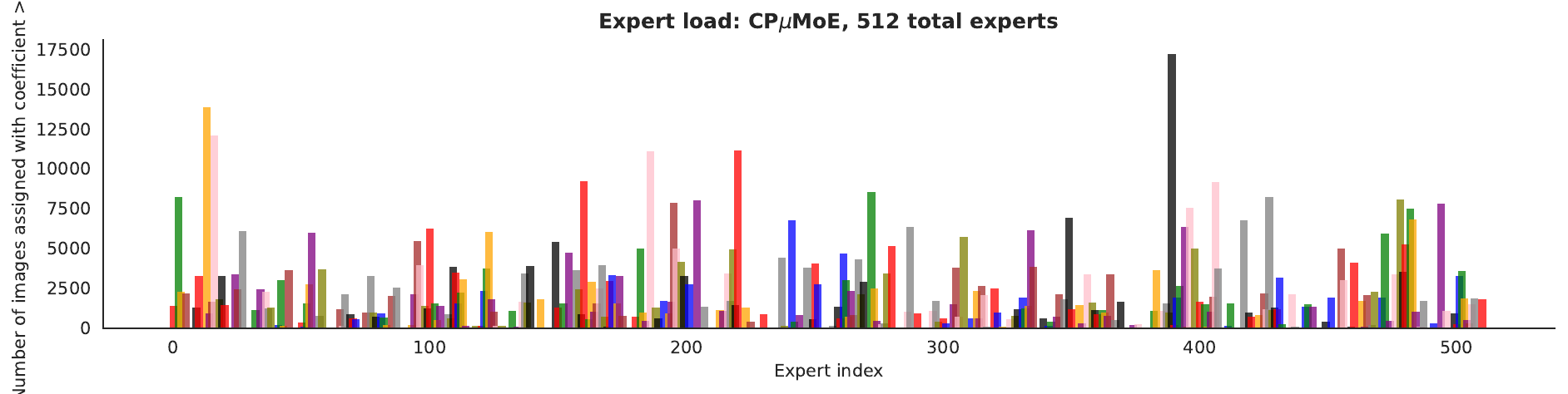}
        \label{fig:expert-load-512}
        \vspace{-2em}
        \caption{$512$ total experts}
        \vspace{2em}
    \end{subfigure}%
    
    \begin{subfigure}[b]{1.00\linewidth}
        \includegraphics[width=\linewidth]{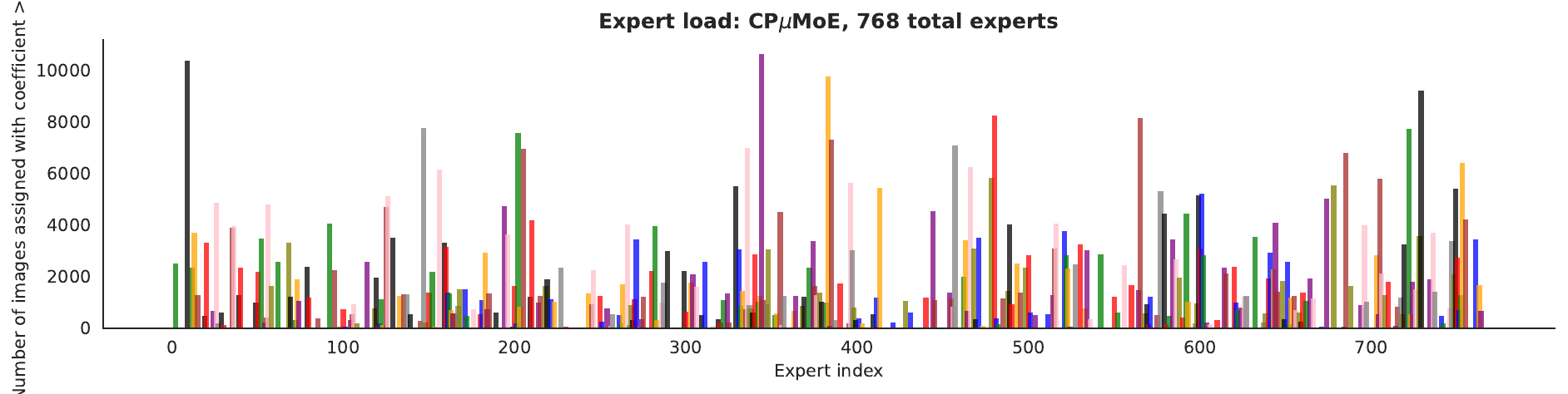}
        \label{fig:expert-load-768}
        \vspace{-2em}
        \caption{$768$ total experts}
        \vspace{2em}
    \end{subfigure}%
    \vspace{-2em}
    \caption{Expert load: Number of training set images with expert coefficient $a_n\geq0.5$ for CP$\mu$MoE models fine-tuned on ImageNET1k. Bars are drawn with 3x width and colored sequentially in a repeating order of distinct colors to help visually distinguish between neighbors.}
    \vspace{2em}
    \label{fig:expert-load}
\end{figure*}

Here, we plot the expert load in \cref{fig:expert-load} to give a visual indication of how many images are processed by each expert with $a_e\geq0.5$ for CP$\mu$MoE final layers fine-tuned on ImageNET1k with a CLIP backbone. Whilst clearly, not all experts have images with a coefficient of at least $0.5$, we see a relatively uniform spread over all experts.
Furthermore, we note the cost from `dead' experts is not particularly troublesome in an $\mu$MoE given its factorized form--speaking informally, we would rather have too many experts than too few, so long as there exist select individual experts conducting the subcomputations of interest.

\section{Additional performance results}
\label{sec:app:additional-performance}
\begin{figure}[ht]
    \centering
    \begin{subfigure}[b]{0.48\textwidth}
        \centering
        \includegraphics[width=\textwidth]{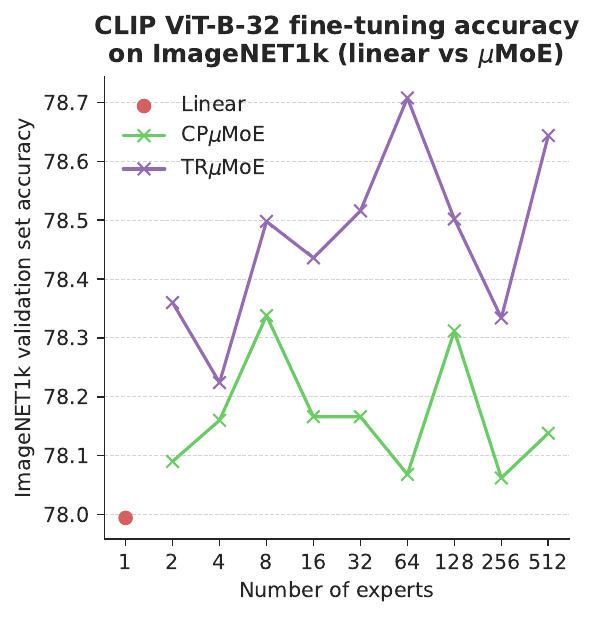}
        \caption{Accuracy comparison ($\mu$MoE vs Linear)}
        \label{fig:clip-ablation}
    \end{subfigure}
    \hfill  %
    \begin{subfigure}[b]{0.48\textwidth}
        \centering
        \includegraphics[width=\textwidth]{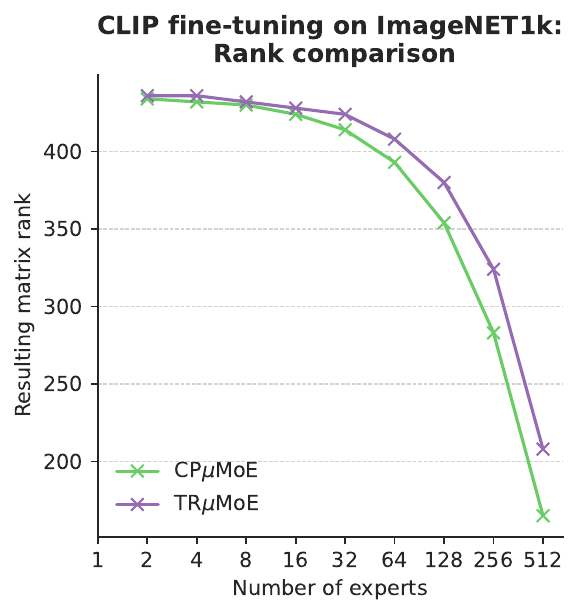}
        \caption{Rank comparison (CP$\mu$MoE vs TR$\mu$MoE)}
        \label{fig:clip-ablation-rank}
    \end{subfigure}
    \caption{Comparative analysis of fine-tuning CLIP ViT-B-32 with $\mu$MoE layers using different configurations. \textbf{All experiments have the same number of parameters}.}
    \label{fig:clip-ablation-super}
\end{figure}

\subsection{CLIP ViT-B-32 ImageNET1k ablations}
Here, we compare the performance of parameter-matched $\mu$MoE final layers (for varying expert counts $N$) to linear layers for fine-tuning large vision-language models (CLIP ViT-B-32) on ImageNET1k.
Following the robust fine-tuning protocol of \cite{wortsman2022robust}, we use the largest possible batch size (to fit on one A100 GPU) of $4096$, and the same learning rate of $3e-05$.

For $\mu$MoE layers, we reduce the layer ranks to parameter match \textit{single} linear layers for each value of total expert count. We plot in \cref{fig:clip-ablation} the ImageNET1k validation loss after 10 epochs of training, where all expert counts out-perform the linear layers initialized the same default way with elements from $U[-k,k]$.
However, to parameter-match single dense linear layers, we must decrease the $\mu$MoE layer rank upon increasing the expert count. 
This is a concrete example of where the extra parameter efficiency of TR$\mu$MoEs can come in useful (as discussed in \cref{sec:app:tr-rank}). Consequently, TR$\mu$MoEs' resulting expert matrix ranks are increasingly larger than that of CP$\mu$MoEs in the parameter-matched setting. For example, the parameter-matched layers with 512 experts in \cref{fig:clip-ablation} have a max expert matrix rank of 165 for the CP$\mu$MoE compared to a much larger 208 for the TR$\mu$MoE.

We attribute TR$\mu$MoE's even greater performance gains over CP$\mu$MoEs here to the more favorable relationship between tensor rank and expert matrix rank (a larger weight matrix rank meaning the resulting layers' activations live in a larger dimensional subspace) (see \cref{fig:clip-ablation-rank}).

\subsection{Hierarchical $\boldsymbol\mu$MoEs}

\paragraph{Hierarchical $\boldsymbol\mu$MoE Mixers}

We train from scratch two hierarchical $\mu$MoE MLP-mixer \texttt{S-16} models for $300$ epochs on ImageNET following the same configuration as in \cref{sec:exp:performance} of the main paper.
Concretely, we use a \textbf{two-level} hierarchical $\mu$MoE with $N_1=64$ experts for the first level and $N_2=2$ experts for the second layer ($128$ total effective experts).
As shown through the results in \cref{tab:hierarchy2-mixers}, the hierarchical $\mu$MoE's also perform well against the MLP alternatives, whilst providing even better parameter-efficiency.

\begin{table}[]
    \centering %
    \caption{\textbf{Hierarchical \texttt{S-16} TR$\boldsymbol\mu$MoE-mixers and CP$\boldsymbol\mu$MoE-mixers}: ImageNET1k val. accuracy at 300 epochs pre-training; $N_1=64,N_2=2$ experts).}
    \label{tab:hierarchy2-mixers}
    \resizebox{0.6\textwidth}{!}{%
        \begin{tabular}{@{}lccc@{}}
        \toprule
        Model & Val. acc. ($\uparrow$) & \# Experts per block & \# Params \\ \midrule
        MLP & 70.31 & n/a & 18.5M \\
        \textbf{CP$\boldsymbol\mu$MoE} (hierarchy=$1$) & 71.29 & $64$ & 18.6M \\
        \textbf{TR$\boldsymbol\mu$MoE} (hierarchy=$1$) & 71.26 & $64$ & 18.3M \\
        \textbf{CP$\boldsymbol\mu$MoE} (hierarchy=$2$) & 71.24 & $64\cdot2$ & 19.5M \\
        \textbf{TR$\boldsymbol\mu$MoE} (hierarchy=$2$) & \textbf{71.56} & $64\cdot2$ &18.7M \\
        \bottomrule
        \end{tabular}
    }
\end{table}

\paragraph{Hierarchical $\boldsymbol\mu$MoE fine-tuning layers}

We also perform additional experiments with hierarchical $\mu$MoEs used to fine-tune CLIP \texttt{ViT-B-32} models on ImageNET1k.
Here we use the experimental setup in \cite{ilharco2022patching,ilharco2023editing}, training each model for a single epoch with the specified learning rate of $1e-05$.
We fine-tune hierarchical $\mu$MoE CLIP models with up to $4$ levels of hierarchy as shown in \cref{tab:sup:hierarchical}, where the best-performing models (averaged over 5 runs) are found with $2$ levels of hierarchy.

\begin{table}[h]
\caption{\textbf{Hierarchical $\boldsymbol\mu$MoEs}: Mean validation-set accuracy with a CLIP ViT-B-32 fine-tuned with hierarchical $\mu$MoE final layers on ImageNET1k. Shown are the number of parameters as the number of total experts increases to $8192$ with 4 levels of hierarchy, and the corresponding number of parameters needed for each expert total using a hierarchy $1$ $\mu$MoE, and regular MoE. Results are the average over 5 runs with different seeds.
Additional expert modes for TR$\mu$MoEs have the additional ranks set equal to the corresponding number of experts at the new mode(s) (e.g. 2 and 4).
}
\label{tab:sup:hierarchical}
\centering
\vspace{0.25em}
\subcaption{Hierarchical \texttt{CP$\mu$MoE}s ($R=512$) fine-tuning CLIP \texttt{ViT-B-32} on ImageNET1k.}
\vspace{-0.25em}
\resizebox{\linewidth}{!}{
\centering
\label{tab:sup:hierarchy-cp}
\begin{tabular}{clllccc}
\toprule
    Hierarchy & Val acc & Weight tensor shape & Total \# experts & \# Params & \# Params needed (w/ 1 hierarchy $\mu$MoE) & \# Params needed (w/ regular MoE) \\
\midrule
1 & $73.78\pm0.07$ &         $\mathcal{W}\in\mathbb{R}^{128\times I\times O}$&                128 &     1,069,568 &                   1,069,568 & 98,432,000 \\
\midrule
2 & $73.84\pm0.11$ &         $\mathcal{W}\in\mathbb{R}^{128\times 2 \times I \times O}$&                256 &     1,072,128 &                   1,233,408 & 196,864,000 \\
3 & $73.80\pm0.14$ &         $\mathcal{W}\in\mathbb{R}^{128\times 2 \times 2\times I \times O}$&                512 &     1,074,688 &                   1,561,088 & 393,728,000 \\
4 & $73.82\pm0.06$ &         $\mathcal{W}\in\mathbb{R}^{128\times 2 \times 2 \times 2 \times I\times O}$&               1024 &     1,077,248 &                   2,216,448 & 787,456,000 \\
\midrule
2 & $\textbf{73.89}\pm\textbf{0.10}$ &         $\mathcal{W}\in\mathbb{R}^{128\times 4 \times I\times O}$&                512 &    1,074,688 &                   1,561,088 & 393,728,000 \\
3 & $73.85\pm0.08$ &         $\mathcal{W}\in\mathbb{R}^{128 \times 4 \times 4 \times I\times O}$&                2048 &  1,079,808 &                   3,527,168 & 1,574,912,000 \\
4 & $73.82\pm0.09$ &         $\mathcal{W}\in\mathbb{R}^{128 \times 4 \times 4 \times 4 \times I\times O}$&               8192 &    1,084,928 &                   11,391,488 & 6,299,648,000 \\
\bottomrule
\end{tabular}
}
\label{tab:sup:hierarchy-tr}
\centering
\vspace{0.25em}
\subcaption{Hierarchical \texttt{TR$\mu$MoE}s ($R_3=512$) fine-tuning CLIP \texttt{ViT-B-32} on ImageNET1k.}
\vspace{-0.25em}
\resizebox{\linewidth}{!}{
\centering
\label{tab:subtable1}
\begin{tabular}{clllccc}
\toprule
    Hierarchy & Val acc & Weight tensor shape & Total \# experts & \# Params & \# Params needed (w/ 1 hierarchy $\mu$MoE) & \# Params needed (w/ regular MoE) \\
\toprule
1 & $74.66\pm0.09$ &         $\mathcal{W}\in\mathbb{R}^{128 \times I\times O}$&                128 & 3,723,264 & 3,723,264 & 98,432,000 \\
\midrule
2 & $74.72\pm0.08$ &         $\mathcal{W}\in\mathbb{R}^{128 \times 2 \times I\times O}$&                256 & 3,724,832 &       3,823,616            & 196,864,000 \\
3 & $74.75\pm0.14$  &         $\mathcal{W}\in\mathbb{R}^{128 \times 2 \times 2 \times I\times O}$&                512 & 3,726,400 &       4,024,320         & 393,728,000 \\
4 & $74.76\pm0.11$ &         $\mathcal{W}\in\mathbb{R}^{128 \times 2 \times 2 \times 2 \times I\times O}$&               1024 & 3,727,968  &     8,851,456   & 787,456,000 \\
\midrule
2 & $\textbf{74.82}\pm\textbf{0.11}$ &         $\mathcal{W}\in\mathbb{R}^{128 \times 4 \times I\times O}$&                512 &         3,726,400 &  4,024,320    & 393,728,000 \\
3 & $74.67\pm0.12$ &        $\mathcal{W}\in\mathbb{R}^{128 \times 4 \times 4 \times I\times O}$&                2048 & 3,729,536 &   5,228,544    & 1,574,912,000 \\
4 & $74.73\pm0.11$ &         $\mathcal{W}\in\mathbb{R}^{128\times4\times4\times4\times I\times O}$&               8192 & 3,732,672 &   10,045,440   & 6,299,648,000 \\
\bottomrule
\end{tabular}
}
\end{table}

\subsection{Comparisons to dense/sparse MoEs}

\begin{figure}
    \centering
    \includegraphics[width=1.00\linewidth]{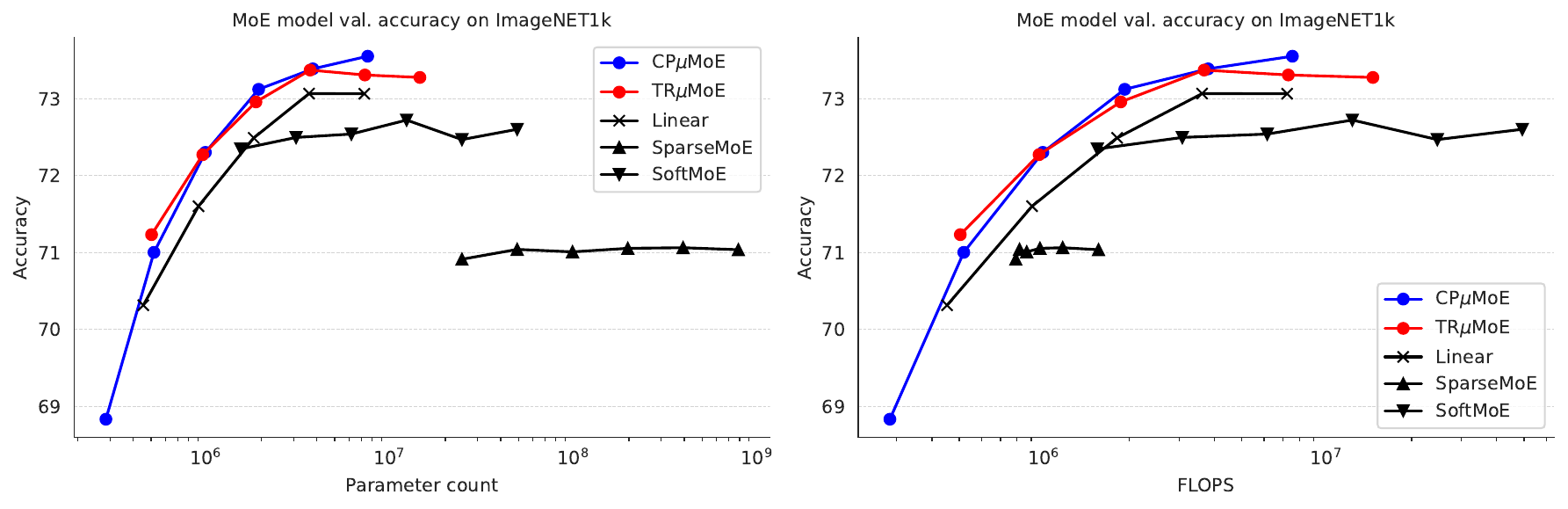}
    \caption{Results fine-tuning CLIP ViT-B-32 final layers only on ImageNET1k for 1 epoch. For $\mu$MoE layers, we increase parameter counts by varying the ranks for a fixed 64 experts. For dense (``Soft'') and sparse MoEs, we increase the parameters through increased expert counts.}
    \label{fig:compare-smoe}
\end{figure}

The goal of the $\mu$MoE layer is to facilitate more interpretable subcomputations with a similar number of parameters and FLOPs to regular dense layers.
Whilst the layer does not aim to improve on the \textit{capabilities} of existing MoE layers, we nonetheless provide an initial comparison study here in \cref{fig:compare-smoe} for completeness.
As can be seen, in addition to the scalable expert specialization provided, the $\mu$MoEs also perform very favorably against the alternative MoE models when fine-tuning CLIP on ImageNET1k.

\section{Fairness baselines \& metric details}
\label{sec:app:fairness-details}

Here we present more details about the fairness comparisons and metrics used in the main paper.

\paragraph{Metrics}
\begin{itemize}
  \setlength{\itemsep}{1pt}
  \setlength{\parskip}{0pt}
  \setlength{\parsep}{0pt}
    \item \textbf{Equality of opportunity} requires the true positive rates for the sensitive attribute subpopulations to be equal, defined in \citet{hardt2016equality} as $P(\hat{Y}=1 \vert A=0, Y=1)=P(\hat{Y}=1 \vert A=1, Y=1)$ for sensitive attribute $A$, target attribute $Y$, and predictor $\hat{Y}$. In the first of our CelebA experiments we measure the absolute difference of the true positive rates between the `blond female' and `blond male' subpopulations for the `blond hair' target attribute. For the second we measure the difference between that of the `old female' and `old male' subpopulations, taking the `old' label as the true target attribute.
    \item \textbf{Standard deviation bias} computes the standard deviation of the accuracy for the different subpopulations \cite{wang2020mitigating}. Intuitively, a small STD bias indicates similar performance across groups.
    \item \textbf{Max-Min Fairness} quantifies the worst-case performance for the different demographic subpopulations \cite{lahoti2020fairness}, with $\max\min_{y\in\mathcal{Y},a\in\mathcal{A}} P(\hat{Y}=y \vert A=a, Y=y)$. We compute this as the minimum of the test-set accuracy for the $4$ subpopulations in each experiment.
\end{itemize}

\paragraph{Baselines}
\begin{itemize}
  \setlength{\itemsep}{1pt}
  \setlength{\parskip}{0pt}
  \setlength{\parsep}{0pt}
    \item \textbf{Oversample} we oversample the low-support subpopulation to balance the number of input images that have the sensitive attribute for the value of the target attribute wherein bias occurs. For example, we oversample the `blond males' to match the number of `blond females' for the first experiment, and oversample the number of `old females' to match the number of `old males' for the second.
    \item \textbf{Blind thresholding} is implemented by unconditionally increasing/decreasing the logits in the target direction for all outputs.
    Concretely, the results in the main paper are achieved by setting $\lambda:=2.5$ and $\bar{\mathbf{a}}$ to a vector of ones in \cref{eq:mmoe-intervention} for all experiments. We find this value of $\lambda$ to give us the best results for the attribute-blind re-writing \cite{hardt2016equality}.
    \item \textbf{Adversarial debiasing} we observe in \cref{tab:intervention-fairness} the same poor performance for the adversarial debiasing technique as is reported in \citet{wang2020towards}. We hypothesize that the same issues face the technique in our experimental setup. In particular, even in the absence of discriminative information for the `gender' label in the final representation, information about correlated attributes (e.g. wearing makeup) are likely still present. This makes it fundamentally challenging to apply fairness-through-unawareness techniques in the CelebA multi-class setting.
\end{itemize}

\section{Fairness: additional results}

\subsection{Model re-writing}
\label{sec:app:supopulation-accuracy-intervention}

\begin{figure}[]
    \centering
    \begin{subfigure}[b]{0.85\linewidth}
        \includegraphics[width=\linewidth]{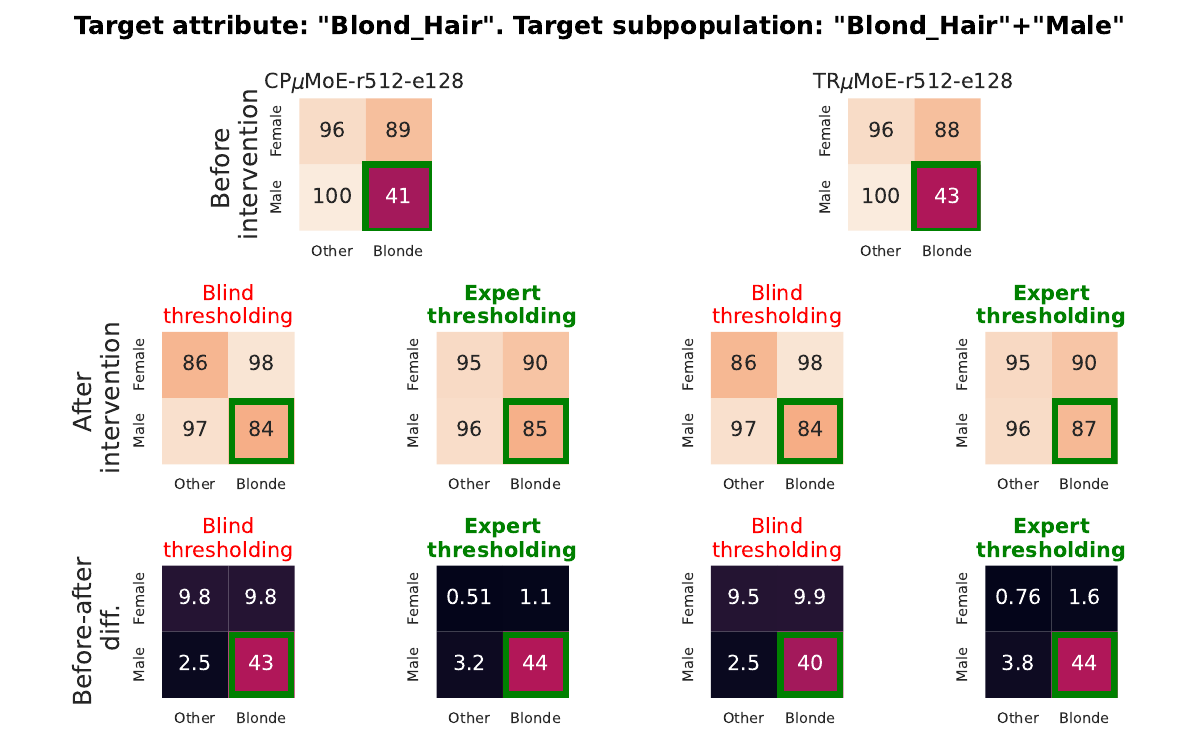}
        \label{fig:intervene-before-after-sub1}
        \caption{`Young blond' intervention for Blond hair attribute prediction head}
    \end{subfigure}%
    \hfill
    \begin{subfigure}[b]{0.85\linewidth}
        \includegraphics[width=\linewidth]{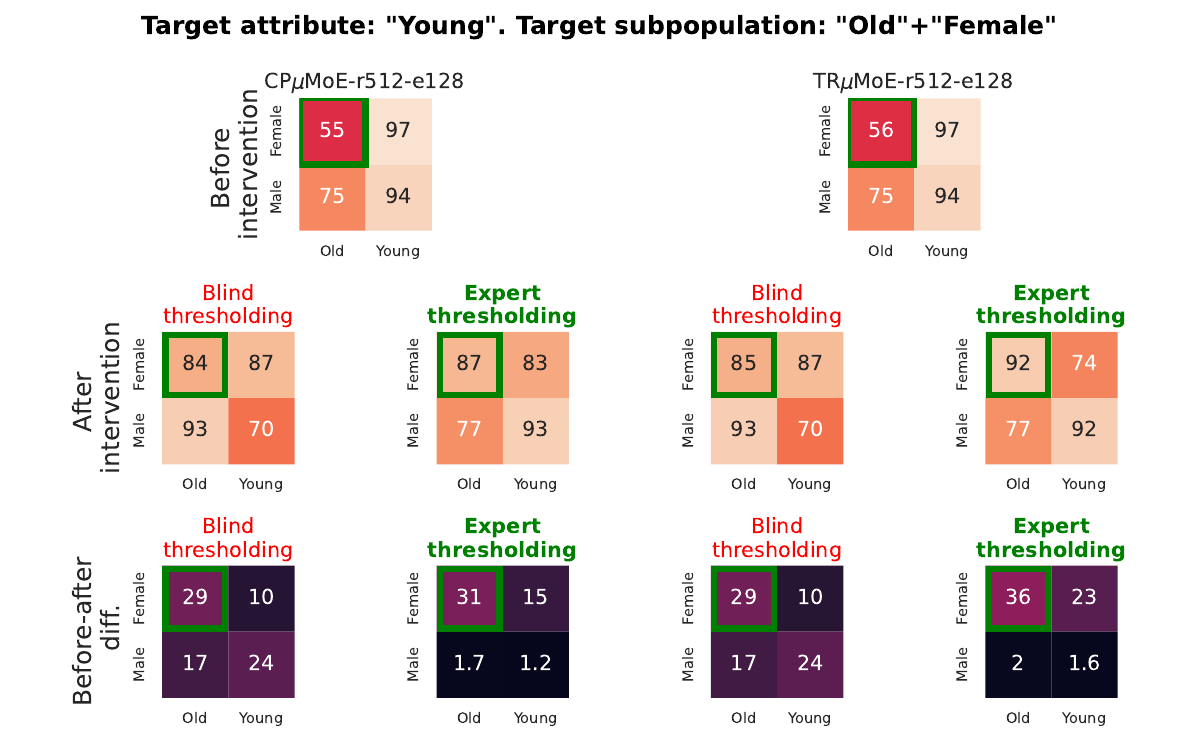}
        \label{fig:intervene-before-after-sub2}
        \caption{`Old female' intervention for age attribute prediction head}
    \end{subfigure}%
    \caption{CelebA Subpopulation accuracies before (first rows) and after intervention (second rows), followed by their absolute difference (third rows). \textbf{\textcolor{ForestGreen}{Green rectangles}} denote the target subpopulation for each experiment (subfigure).}
    \label{fig:fairness-before-after}
\end{figure}

The full per-subpopulation test set accuracies are shown in \cref{fig:fairness-before-after} for the two experiments in the main paper. The first rows show the accuracies before layer re-write, the second rows after re-write, and the third rows the absolute difference between the two. As can be seen in the `before-after difference' final rows of \cref{fig:fairness-before-after}, the proposed expert-conditional re-write provides much more precision in changing only the computation for the target populations.

\clearpage
\newpage
\section{NeurIPS Paper Checklist}

\begin{enumerate}

\item {\bf Claims}
    \item[] Question: Do the main claims made in the abstract and introduction accurately reflect the paper's contributions and scope?
    \item[] Answer: \answerYes{} %
    \item[] Justification: Claims regarding both qualitative and quantitative expert specialism for fine-tuning large foundation models are demonstrated in \cref{sec:exp:prune}, where the benefits of scaling the expert counts are also substantiated both qualitatively and quantitatively. Claims regarding bias mitigation are substantiated in \cref{sec:exp:intervene}. Qualitative expert specialism is provided for large models (along with their performance) in \cref{sec:large-models}.

\item {\bf Limitations}
    \item[] Question: Does the paper discuss the limitations of the work performed by the authors?
    \item[] Answer: \answerYes{} %
    \item[] Justification: The limitations clearly state the lack of evaluation for out-of-domain data for vision, and the difficulties in further evaluating expert specialism quantitatively in large models (given the lack of ground-truth).

\item {\bf Theory Assumptions and Proofs}
    \item[] Question: For each theoretical result, does the paper provide the full set of assumptions and a complete (and correct) proof?
    \item[] Answer: \answerNA{} %
    \item[] Justification: Technical derivations of models are made throughout (and further basic derivations of expert matrix rank), but no novel theoretical results are presented.

    \item {\bf Experimental Result Reproducibility}
    \item[] Question: Does the paper fully disclose all the information needed to reproduce the main experimental results of the paper to the extent that it affects the main claims and/or conclusions of the paper (regardless of whether the code and data are provided or not)?
    \item[] Answer: \answerYes{} %
    \item[] Justification: Full experiment settings/config/hyperparameters are provided in \cref{tab:experimental-config}, and the supporting code (\url{https://github.com/james-oldfield/muMoE}) provides even more explicit experimental instructions. Learning curves are also plotted in \cref{fig:gpt-curves,fig:mixer-curves} for additional transparency. Pseudocode implementations are also given in \cref{sec:app:implementations}.

\item {\bf Open access to data and code}
    \item[] Question: Does the paper provide open access to the data and code, with sufficient instructions to faithfully reproduce the main experimental results, as described in supplemental material?
    \item[] Answer: \answerYes{} %
    \item[] Justification: Model code for $\mu$MoEs and the experiments in the paper are found at:\url{https://github.com/james-oldfield/muMoE}.

\item {\bf Experimental Setting/Details}
    \item[] Question: Does the paper specify all the training and test details (e.g., data splits, hyperparameters, how they were chosen, type of optimizer, etc.) necessary to understand the results?
    \item[] Answer: \answerYes{} %
    \item[] Justification: As found in \cref{tab:experimental-config}, where we state we follow these choices based on the default parameters of the original papers introducing the models, or the default configurations used by the open-source maintainer for GPT2.

\item {\bf Experiment Statistical Significance}
    \item[] Question: Does the paper report error bars suitably and correctly defined or other appropriate information about the statistical significance of the experiments?
    \item[] Answer: \answerNo{} %
    \item[] Justification: We do include mean (and STD) of the results over multiple fine-tuning models, but we only have single runs over the large models due to resource constraints. For these single runs of large models, we always set all random seeds to $0$ for reproducibility.
    
\item {\bf Experiments Compute Resources}
    \item[] Question: For each experiment, does the paper provide sufficient information on the computer resources (type of compute workers, memory, time of execution) needed to reproduce the experiments?
    \item[] Answer: \answerYes{} %
    \item[] Justification: Details are provided in \cref{sec:app:config}.
    
\item {\bf Code Of Ethics}
    \item[] Question: Does the research conducted in the paper conform, in every respect, with the NeurIPS Code of Ethics \url{https://neurips.cc/public/EthicsGuidelines}?
    \item[] Answer: \answerYes{} %
    \item[] Justification: No ethical concerns to note.

\item {\bf Broader Impacts}
    \item[] Question: Does the paper discuss both potential positive societal impacts and negative societal impacts of the work performed?
    \item[] Answer: \answerYes{} %
    \item[] Justification: The paper proposed a layer that provides more transparent, explainable, and editable networks. We discuss positive social impacts throughout the paper, but also acknowledge and discuss the potential negative impacts in \cref{sec:app:broader-impact}.
    
\item {\bf Safeguards}
    \item[] Question: Does the paper describe safeguards that have been put in place for responsible release of data or models that have a high risk for misuse (e.g., pretrained language models, image generators, or scraped datasets)?
    \item[] Answer: \answerNA{} %
    \item[] Justification: No models posing a high risk of misuse are to be released.

\item {\bf Licenses for existing assets}
    \item[] Question: Are the creators or original owners of assets (e.g., code, data, models), used in the paper, properly credited and are the license and terms of use explicitly mentioned and properly respected?
    \item[] Answer: \answerYes{} %
    \item[] Justification: Yes, the open-source codebases on which we base our code are explicitly referenced.
    
\item {\bf New Assets}
    \item[] Question: Are new assets introduced in the paper well documented and is the documentation provided alongside the assets?
    \item[] Answer: \answerNA{} %
    \item[] Justification: None introduced.

\item {\bf Crowdsourcing and Research with Human Subjects}
    \item[] Question: For crowdsourcing experiments and research with human subjects, does the paper include the full text of instructions given to participants and screenshots, if applicable, as well as details about compensation (if any)? 
    \item[] Answer: \answerNA{} %
    \item[] Justification: No human subjects involved.
    
\item {\bf Institutional Review Board (IRB) Approvals or Equivalent for Research with Human Subjects}
    \item[] Question: Does the paper describe potential risks incurred by study participants, whether such risks were disclosed to the subjects, and whether Institutional Review Board (IRB) approvals (or an equivalent approval/review based on the requirements of your country or institution) were obtained?
    \item[] Answer: \answerNA{} %
    \item[] Justification: No human subjects involved.
    
\end{enumerate}

\end{document}